\documentclass[11pt]{article}
\usepackage[preprint]{acl}
\usepackage[T1]{fontenc}
\usepackage{times,amsmath,amssymb,booktabs,microtype,placeins,tikz,textcomp,graphicx}
\newsavebox{\WeightTableBox}
\author{
  Zhe Chen$^{1}$ \quad Xiaolin Zhou$^{2}$ \quad Xiyang Hu$^{2}$ \\
  $^{1}$University of Southern California \\
  $^{2}$Arizona State University \\
  \texttt{zchen116@usc.edu} \quad \texttt{xzhou226@asu.edu} \quad \texttt{xiyang.hu@asu.edu}
}
\hypersetup{
  pdftitle={Phase-Conditioned Reward Shaping for Token-Efficient LLM Reasoning},
  pdfauthor={Zhe Chen, Xiaolin Zhou, Xiyang Hu}
}
\date{}
\newsavebox{\CaseLayoutBox}
\newcommand{\CaseQuestion}[2]{%
  \par\noindent\textbf{#1}\par\smallskip
  {\setlength{\fboxsep}{7pt}\setlength{\fboxrule}{0.4pt}%
  \fcolorbox{black!25}{white}{\begin{minipage}{\dimexpr\linewidth-2\fboxsep-2\fboxrule\relax}
  \fontsize{10}{12}\selectfont\raggedright #2\end{minipage}}}\par\medskip}
\newcommand{\CaseCard}[6]{%
  \begin{minipage}[t]{#1\linewidth}\vspace{0pt}%
  \setlength{\fboxsep}{6pt}%
  \colorbox{black!5}{\begin{minipage}{\dimexpr\linewidth-2\fboxsep\relax}
  \fontsize{9}{11}\selectfont\raggedright
  {\setlength{\fboxsep}{4pt}\colorbox{black!72}{%
    \parbox{\dimexpr\linewidth-2\fboxsep\relax}{\color{white}\strut\textbf{#2}}}}\par\smallskip
  {\fontsize{8.5}{10}\selectfont Sample #3\enspace$\cdot$\enspace #4 tokens\par
  \textbf{Selected answer: #5.}}\par\medskip
  #6
  \end{minipage}}\end{minipage}}
\newcommand{\CaseLegend}{\par\smallskip{\fontsize{8}{10}\selectfont
  \textcolor{blue!65!black}{Blue: valid step/repair.}\enspace
  \textcolor{orange!70!black}{Orange: error.}\enspace
  \textcolor{violet!80!black}{Purple: repeated checking.}\enspace
  [\ldots]: omitted text.}}

\title{Phase-Conditioned Reward Shaping for Token-Efficient LLM Reasoning}
\begin{document}\maketitle\raggedbottom
\begin{abstract}

Extended reasoning can help language models solve complex problems, but long responses often contain both useful computations and redundant checks. Conventional length penalties encourage shorter responses without distinguishing the functions of intermediate steps. We propose \textbf{Phase-Conditioned Reward Shaping (PCRS)}, a framework that combines phase identification with training-time reward allocation. We investigate two classification routes: PCRS-H with a hidden-state classifier and PCRS-T with a frozen text classifier. Our main text classifier implementation assigns overlapping reasoning, execution, and revision labels to native chunks, then uses these labels and exact repetition to adjust existing token penalties. We compare the completed configurations on six mathematical and scientific reasoning benchmarks and examine four reward-weight settings with the text classifier. Compared with a baseline using token-level length penalties, the default text classifier configuration reduces mean response length by 3.13\%; sampled Pass@1 is approximately 62.07\% for both methods, while oracle Pass@16 rises from 86.08\% to 87.67\%. These results show how phase-informed penalty allocation can reduce output volume and improve multi-attempt coverage without inference-time classification.\looseness=-1

\end{abstract}
\section{Introduction}
A reasoning model can spend tokens verifying a useful constraint or repeating an unhelpful calculation. Although both increase response length, they serve different purposes in the solution process. Existing approaches improve reasoning efficiency through length-sensitive reinforcement learning \citep{arora2025training}, adaptive penalties based on problem-level solve rates \citep{xiang2025alp}, and token-level rewards that distinguish necessary reasoning prefixes from later computation \citep{jiang2025decs}. However, these methods primarily determine how much reasoning to penalize or where to apply the penalty, without explicitly considering the computational roles of intermediate steps. This raises a central question: can length penalties be allocated according to the type of computation being performed and its redundancy, rather than relying mainly \mbox{on response length or position?}

\begin{figure*}[t]
\centering
\includegraphics[width=\textwidth, trim={0 1.5cm 0 2.5cm}, clip]{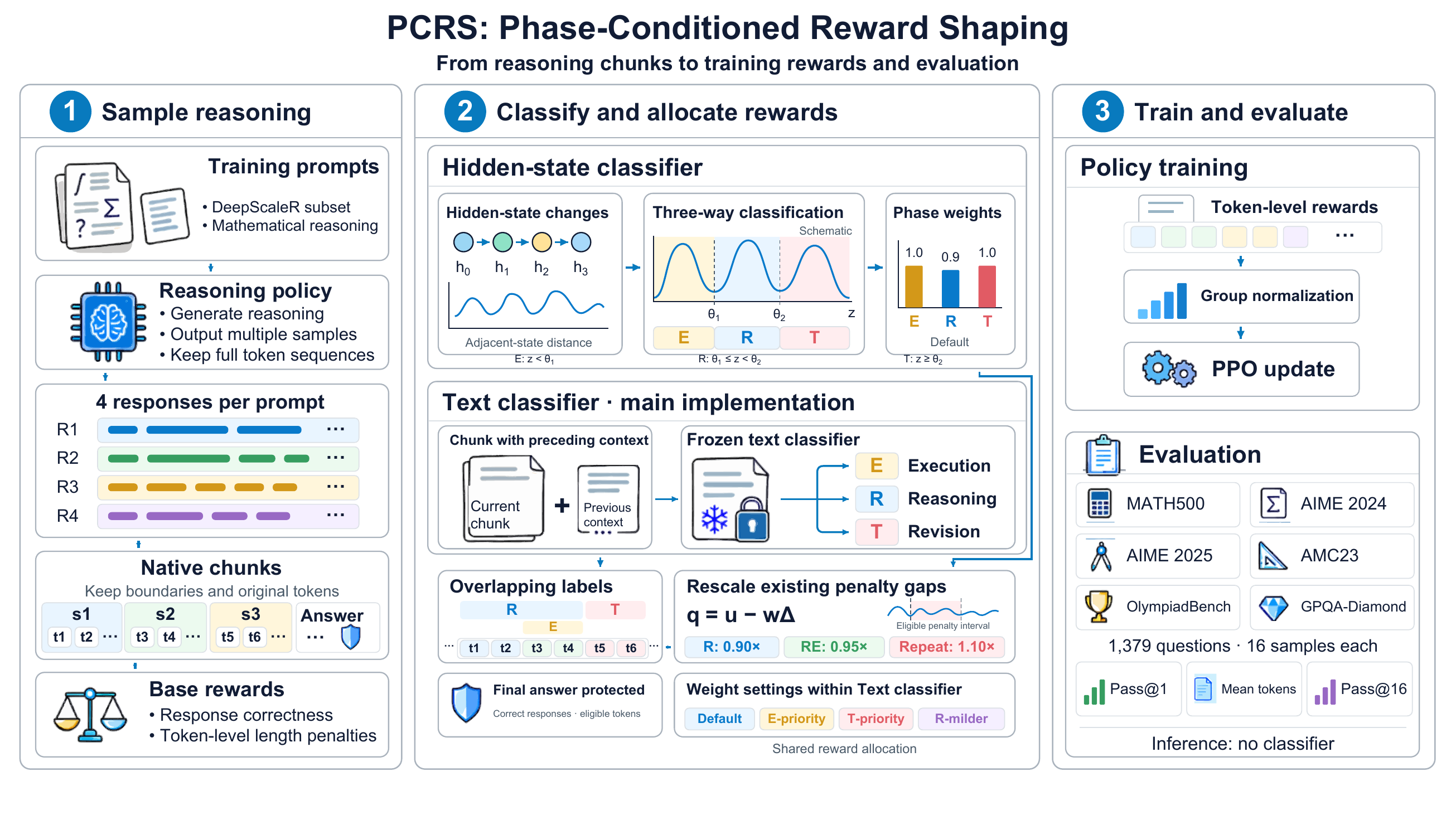}
\caption{PCRS overview. PCRS-H assigns exclusive phase labels from hidden states; PCRS-T predicts overlapping labels from text. Both routes use these labels and exact repetition to adjust length penalties before policy optimization. R denotes reasoning, E execution, T revision, and RE joint R/E membership.}
\label{fig:method}
\end{figure*}

To address this question, we introduce Phase-Conditioned Reward Shaping (PCRS), a framework with two design choices: how to identify the computational phase of a reasoning chunk, and how to use that information in the reward pipeline. We investigate two classification routes: PCRS-H, which uses a hidden-state classifier inspired by ThinKV's thought-type distinction \citep{ramachandran2026thinkv}, and PCRS-T, which uses a frozen lightweight text classifier operating on chunk text. PCRS-T is our main implementation. It identifies three non-exclusive roles: reasoning and constraint formation (R), execution of intermediate computations (E), and revision or changes in approach (T). Labels can overlap: RE denotes a chunk assigned both R and E. The T~classifier reduces existing penalties on eligible tokens in chunks containing R, while an exact-repetition rule strengthens penalties on chunks containing repeated sentences. Figure~\ref{fig:method} presents the workflow from response sampling and classification to reward allocation, \mbox{policy optimization, and evaluation.}

We organize the experiments around these two design choices. First, we compare the completed PCRS-H and PCRS-T configurations with our DECS reproduction \citep{jiang2025decs} on six mathematical and scientific reasoning benchmarks comprising 1,379 questions. A repetition-only control tests whether repetition penalties alone recover the full configuration's gains; we then compare Default, E-priority, T-priority, and R-milder weights within PCRS-T. With default weights, PCRS-T reduces mean response length by 3.13\% while maintaining nearly identical sampled Pass@1 point estimates (62.07\%). Oracle Pass@16, which measures whether at least one of 16 sampled responses is correct, increases from 86.08\% to 87.67\%. Paired reasoning traces connect these aggregate outcomes to valid intermediate computations and repeated checking.

Our main contribution is a phase-conditioned reward framework that separates classification signals from reward allocation. We describe the hidden-state and text-based routes and their reward integrations, evaluate the T~classifier against DECS and a repetition-only control, and examine reward-weight choices with the classifier held fixed. This organization connects chunk-level information to token-level penalty allocation and makes the trade-offs between accuracy, output length, and multi-attempt coverage explicit.

\section{Related Work}

\paragraph{Length control for efficient reasoning.}
Reinforcement learning improves language-model reasoning \citep{shao2024deepseekmath,deepseek2025r1}, while extended reasoning can introduce unnecessary computation \citep{chen2024overthink}. s1 controls response length through inference-time budget forcing \citep{muennighoff2025s1}; ThinkPrune progressively reduces token limits during reinforcement learning \citep{team2025thinkprune}. Other approaches balance reasoning cost and correctness \citep{arora2025training}, adapt length penalties to problem-level solve rates \citep{xiang2025alp}, or combine token-level rewards with necessary-prefix detection and \mbox{curriculum scheduling \citep{jiang2025decs}}.

\paragraph{Process-level feedback and reasoning behavior.}
Process supervision evaluates the correctness of individual reasoning steps \citep{lightman2023verify}. REA-RL uses a lightweight reflection model and reflection-aware rewards to preserve useful reflection during efficiency-oriented training \citep{deng2025rearl}. DuP-PO balances responses with and without thinking tokens through sampling, advantage control, and gradient updates \mbox{\citep{ding2025thinking}}.

\paragraph{Thought-aware inference and evaluation.}
ThinKV identifies thought types using attention sparsity and adjusts KV-cache precision and retention \citep{ramachandran2026thinkv}. Deep-Thinking Tokens uses changes in internal predictions across model layers for test-time sample selection \citep{chen2026deepthinking}. Self-consistency selects answers by aggregating sampled reasoning paths \citep{wang2023selfconsistency}, while pass@$k$ measures the probability of obtaining at least one correct response across multiple attempts \citep{chen2021code}. Our focus is training-time penalty allocation based on the functional roles of reasoning chunks.

\section{Method}
\label{sec:methods}

Phase-Conditioned Reward Shaping (PCRS) combines phase classification with training-time reward allocation. PCRS-H uses a hidden-state classifier with mutually exclusive R/E/T labels; PCRS-T uses a frozen text classifier with overlapping labels. Both use the shared reward mechanism below. Within PCRS-T, we compare Default, E-priority, T-priority, and R-milder weights; an unqualified PCRS-T result denotes Default. Implementation details appear in Appendix~\ref{app:implementation}.

\subsection{Base Rewards and Eligible Positions}

For a prompt $x$, the policy samples $G=4$ responses $y_i$. Let $m_{it}$ indicate whether position $t$ in response $i$ is valid, $L_i=\sum_t m_{it}$ denote the response length, and $H$ denote the padded response width used by the reward function. Let $c_i$ be the response score obtained by summing the incoming token rewards. Since KL-in-reward is disabled, $c_i$ is also the score passed to the reward function.

Following DECS \citep{jiang2025decs}, the underlying reward construction uses native reasoning separators and an answer-containing-chunk detector. Let $a_{it}$ denote the chunk-advantage array supplied to the reward function. In the implemented reward function, the positions subject to length penalties form an inclusive interval $I_i$, spanning the first and last negative entries of $a_{it}$. The interval is empty if no negative entry exists. Define $b_{it}$ to be true when $c_i>0$, $t\in I_i$, and $m_{it}=1$. The resulting token-level reward is
\begin{equation}
q^{D}_{it}=
\begin{cases}
c_i+\beta(1-L_i/H), & b_{it},\\
c_i(1+\beta), & \text{otherwise},
\end{cases}
\label{eq:decs}
\end{equation}
where $\beta=0.1$.

We denote the unpenalized reference reward by $u_i=c_i(1+\beta)$. For a response with unit correctness score $c_i=1$, the length-penalty gap at positions in $I_i$ is $u_i-q^D_{it}=\beta L_i/H$. PCRS adjusts this gap selectively, while leaving positions outside its eligibility conditions unchanged. The implemented reward function initializes padded positions to $u_i$; their treatment during \mbox{normalization is described below.}

\subsection{Two Routes to Phase Classification}
\label{sec:role_routes}

PCRS characterizes intermediate chunks using three computational roles: reasoning and constraint formation (R), execution of intermediate computations (E), and revision or changes in approach (T). We use \emph{phase} to refer to these roles, rather than to sequential stages of reasoning. The roles are not mutually exclusive: a chunk may perform multiple functions, and the same role may recur throughout a response. For example, RE combines reasoning and execution.

\paragraph{PCRS-H: hidden-state classifier.}
ThinKV identifies reasoning, execution, and transition through patterns in model-internal signals \citep{ramachandran2026thinkv}. Its calibration examines attention-sparsity distributions across layers, selects layers with distinct modes, and estimates thresholds from the density minima between those modes. We also investigated this attention-sparsity approach offline across all 28 layers of DeepSeek-R1-Distill-Qwen-1.5B, using 100 calibration questions and 50 held-out questions (Appendix~\ref{app:legacy_proxy}). The H~classifier uses normalized changes between adjacent hidden states to assign mutually exclusive R/E/T labels. Two ordered thresholds partition the signal into E, R, and T regions. Each response token receives one label, represented as a singleton set $A_{it}=\{c_{it}\}$, where $c_{it}\in\{R,E,T\}$. Both routes use the same phase-weight mapping and apply it to the existing penalty gap before group normalization. Appendix~\ref{app:legacy_proxy} gives the signal definition and three-class mapping.

\paragraph{PCRS-T: text classifier.}
The T~classifier is a frozen lightweight model operating on native chunks without modifying their boundaries or token IDs. For each chunk $s_j$, the classifier takes its text and the preceding 128 tokens, decoded using the policy tokenizer. The context is truncated to its last 800 characters. A term-frequency--inverse-document-frequency (TF--IDF) representation \citep{manning2008ir} combines word unigrams and bigrams with character 3--5-grams. Three independent binary logistic classifiers estimate the probability of each role:
$p_{jk}=\sigma\left(v_k^\top \phi(s_j,\operatorname{context}_j)+b_k\right)$.

\begin{figure*}[t]
\centering
\resizebox{.94\textwidth}{!}{\input{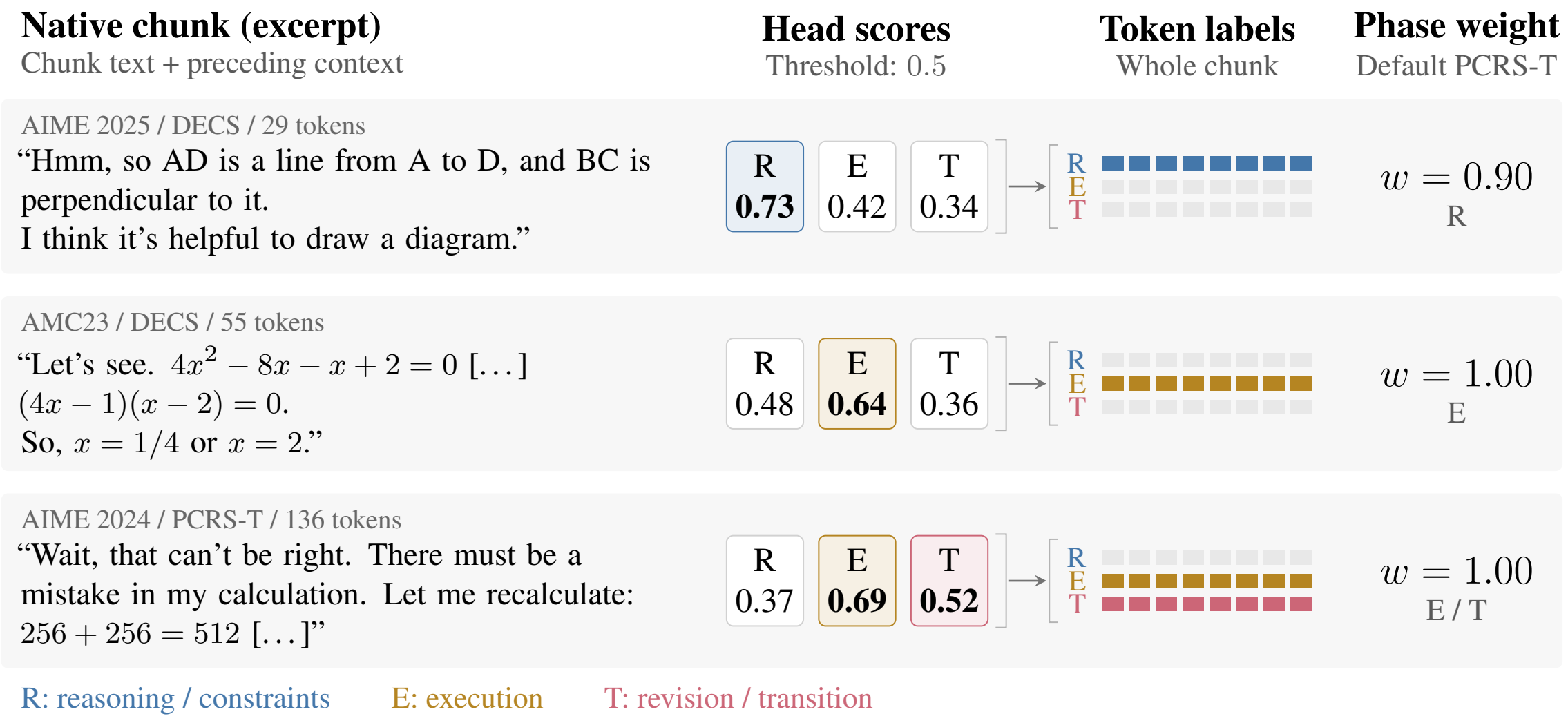}}
\caption{Chunk-level phase predictions and default penalty weights. Roles with scores $\geq 0.5$ are assigned to all tokens in the chunk; multiple roles may be active. Weights precede repetition and protection rules and apply at eligible positions. Token lanes are schematic; ellipses mark omitted text.}
\label{fig:role_examples}
\end{figure*}

The predicted role set is
$A_j=\{k\in\{R,E,T\}:p_{jk}\geq 0.5\}$.
A chunk may receive one or more roles, or none if all probabilities fall below the threshold. The predicted set is assigned to all original token positions within the chunk. Figure~\ref{fig:role_examples} illustrates this assignment using archived response chunks. The R/E/T vocabulary is related to ThinKV \citep{ramachandran2026thinkv}, but PCRS uses non-exclusive text-based predictions rather than attention-based assignments.

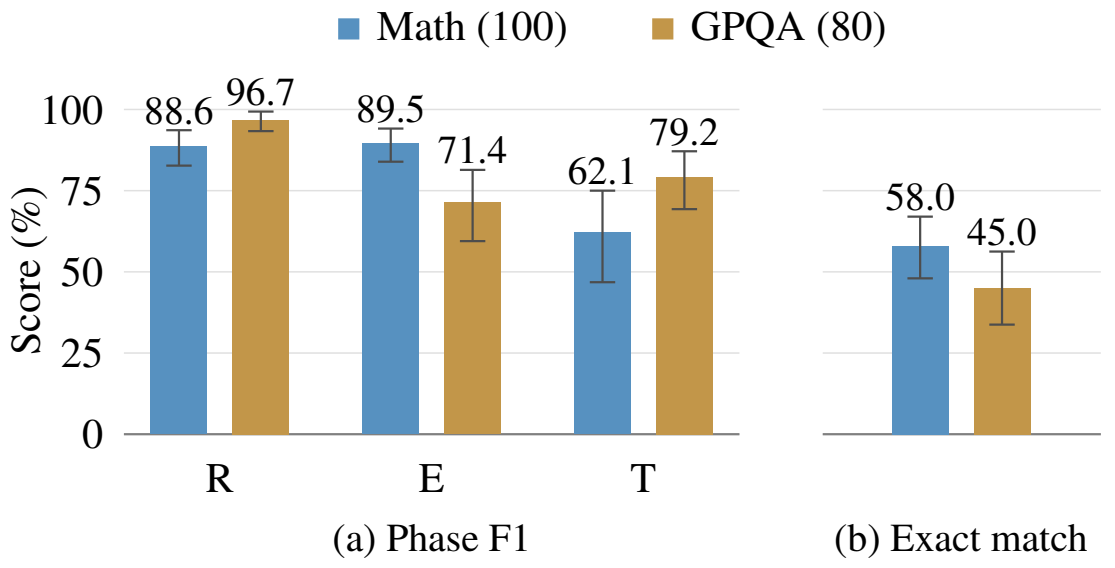
\begin{figure}[t]
\centering
\resizebox{\columnwidth}{!}{\begin{tikzpicture}[x=1cm,y=1cm,font=\rmfamily\fontsize{8.5}{10}\selectfont]
\path[use as bounding box] (-.25,-.48) rectangle (7.90,3.58);
\definecolor{phaseblue}{HTML}{397DB5}
\definecolor{phasegold}{HTML}{B78429}
\node[inner sep=0pt] at (4.18,3.35) {\textcolor{phaseblue!85}{\rule{1.5mm}{1.5mm}}\, Math (100)\qquad\textcolor{phasegold!85}{\rule{1.5mm}{1.5mm}}\, GPQA (80)};
\node[inner sep=0pt,font=\fontsize{8}{9}\selectfont] at (2.9,-.28) {(a) Phase F1};
\node[inner sep=0pt,font=\fontsize{8}{9}\selectfont] at (6.65,-.28) {(b) Exact match};
\node[inner sep=0pt,anchor=east] at (.58,0.48000) {0};
\draw[black!12,line width=.3pt] (0.72,0.48000)--(5.08,0.48000);
\draw[black!12,line width=.3pt] (5.67,0.48000)--(7.64,0.48000);
\node[inner sep=0pt,anchor=east] at (.58,1.05500) {25};
\draw[black!12,line width=.3pt] (0.72,1.05500)--(5.08,1.05500);
\draw[black!12,line width=.3pt] (5.67,1.05500)--(7.64,1.05500);
\node[inner sep=0pt,anchor=east] at (.58,1.63000) {50};
\draw[black!12,line width=.3pt] (0.72,1.63000)--(5.08,1.63000);
\draw[black!12,line width=.3pt] (5.67,1.63000)--(7.64,1.63000);
\node[inner sep=0pt,anchor=east] at (.58,2.20500) {75};
\draw[black!12,line width=.3pt] (0.72,2.20500)--(5.08,2.20500);
\draw[black!12,line width=.3pt] (5.67,2.20500)--(7.64,2.20500);
\node[inner sep=0pt,anchor=east] at (.58,2.78000) {100};
\draw[black!12,line width=.3pt] (0.72,2.78000)--(5.08,2.78000);
\draw[black!12,line width=.3pt] (5.67,2.78000)--(7.64,2.78000);
\node[inner sep=0pt,rotate=90] at (.05,1.64) {Score (\%)};
\draw[black!50,line width=.45pt] (.72,.48)--(5.08,.48);
\draw[black!50,line width=.45pt] (5.67,.48)--(7.64,.48);
\node[inner sep=0pt] at (1.4,.17) {R};
\fill[phaseblue!85] (0.91000,0.48) rectangle (1.31000,2.51713);
\draw[black!70,line width=.45pt] (1.11000,2.38228)--(1.11000,2.63316);
\draw[black!70,line width=.45pt] (1.02000,2.38228)--(1.20000,2.38228);
\draw[black!70,line width=.45pt] (1.02000,2.63316)--(1.20000,2.63316);
\node[inner sep=0pt,font=\fontsize{8}{9}\selectfont] at (1.11000,2.77316) {88.6};
\fill[phasegold!85] (1.49000,0.48) rectangle (1.89000,2.70386);
\draw[black!70,line width=.45pt] (1.69000,2.62669)--(1.69000,2.76517);
\draw[black!70,line width=.45pt] (1.60000,2.62669)--(1.78000,2.62669);
\draw[black!70,line width=.45pt] (1.60000,2.76517)--(1.78000,2.76517);
\node[inner sep=0pt,font=\fontsize{8}{9}\selectfont] at (1.69000,2.90517) {96.7};
\node[inner sep=0pt] at (2.9,.17) {E};
\fill[phaseblue!85] (2.41000,0.48) rectangle (2.81000,2.53789);
\draw[black!70,line width=.45pt] (2.61000,2.41006)--(2.61000,2.64472);
\draw[black!70,line width=.45pt] (2.52000,2.41006)--(2.70000,2.41006);
\draw[black!70,line width=.45pt] (2.52000,2.64472)--(2.70000,2.64472);
\node[inner sep=0pt,font=\fontsize{8}{9}\selectfont] at (2.61000,2.78472) {89.5};
\fill[phasegold!85] (2.99000,0.48) rectangle (3.39000,2.12287);
\draw[black!70,line width=.45pt] (3.19000,1.84759)--(3.19000,2.35208);
\draw[black!70,line width=.45pt] (3.10000,1.84759)--(3.28000,1.84759);
\draw[black!70,line width=.45pt] (3.10000,2.35208)--(3.28000,2.35208);
\node[inner sep=0pt,font=\fontsize{8}{9}\selectfont] at (3.19000,2.49208) {71.4};
\node[inner sep=0pt] at (4.4,.17) {T};
\fill[phaseblue!85] (3.91000,0.48) rectangle (4.31000,1.90757);
\draw[black!70,line width=.45pt] (4.11000,1.55658)--(4.11000,2.20500);
\draw[black!70,line width=.45pt] (4.02000,1.55658)--(4.20000,1.55658);
\draw[black!70,line width=.45pt] (4.02000,2.20500)--(4.20000,2.20500);
\node[inner sep=0pt,font=\fontsize{8}{9}\selectfont] at (4.11000,2.34500) {62.1};
\fill[phasegold!85] (4.49000,0.48) rectangle (4.89000,2.30081);
\draw[black!70,line width=.45pt] (4.69000,2.07408)--(4.69000,2.48397);
\draw[black!70,line width=.45pt] (4.60000,2.07408)--(4.78000,2.07408);
\draw[black!70,line width=.45pt] (4.60000,2.48397)--(4.78000,2.48397);
\node[inner sep=0pt,font=\fontsize{8}{9}\selectfont] at (4.69000,2.62397) {79.2};
\fill[phaseblue!85] (6.16000,0.48) rectangle (6.56000,1.81400);
\draw[black!70,line width=.45pt] (6.36000,1.58400)--(6.36000,2.02100);
\draw[black!70,line width=.45pt] (6.27000,1.58400)--(6.45000,1.58400);
\draw[black!70,line width=.45pt] (6.27000,2.02100)--(6.45000,2.02100);
\node[inner sep=0pt,font=\fontsize{8}{9}\selectfont] at (6.36000,2.16100) {58.0};
\fill[phasegold!85] (6.74000,0.48) rectangle (7.14000,1.51500);
\draw[black!70,line width=.45pt] (6.94000,1.25625)--(6.94000,1.77375);
\draw[black!70,line width=.45pt] (6.85000,1.25625)--(7.03000,1.25625);
\draw[black!70,line width=.45pt] (6.85000,1.77375)--(7.03000,1.77375);
\node[inner sep=0pt,font=\fontsize{8}{9}\selectfont] at (6.94000,1.91375) {45.0};
\end{tikzpicture}}
\caption{Agreement of the frozen T~classifier: (a) phase F1 and (b) exact label-set agreement. Error bars show 95\% question-bootstrap intervals. Reference labels combine majority votes from three model-assisted annotators with subsequent manual inspection by an author. All four reward settings share this classifier.}
\label{fig:role_diagnostics}
\end{figure}

The classifier is trained on 300 MATH chunks \citep{hendrycks2021math} and selected using 96 development chunks (macro-F1 0.7835). Against model-assisted reference annotations manually inspected by an author, it achieves macro-F1 of 0.7167 on 585 held-out MATH500 chunks. Under a separate cross-domain evaluation protocol, macro-F1 is 0.8004 on 100 mathematics chunks and 0.8243 on 80 GPQA chunks. These scores quantify agreement with the reference role annotations. Figure~\ref{fig:role_diagnostics} reports phase-level F1 and exact label-set agreement; Appendix~\ref{app:implementation} details the validation protocol. The classifier remains frozen throughout policy training and is \mbox{not used at inference.}

\subsection{Shared Phase-Conditioned Rewards}

PCRS uses the predicted roles to control the strength of existing length penalties. For the H~classifier, the label set $A_{it}$ contains the single predicted role at each response position. For the T~classifier, $A_{it}$ is the label set of the native chunk containing that position. Both routes \mbox{use the following rule:}
\begin{equation}
w_{it}=
\begin{cases}
1, & \text{protected region},\\
1.1, & \text{exact repetition},\\
\displaystyle\frac{1}{|A_{it}|}
\sum_{k\in A_{it}}\lambda_k,
& A_{it}\neq\varnothing,\\
1, & A_{it}=\varnothing.
\end{cases}
\label{eq:role_weight}
\end{equation}
The default configuration uses
$(\lambda_R,\lambda_E,\lambda_T)=(0.9,1,1)$.

An exact repetition occurs when a normalized sentence matches an earlier sentence and contains at least eight regex-extracted word, numeral, or operator units. This rule detects exact textual repetition rather than semantic similarity. Repetition takes precedence over role-based weights. The chunk containing a closing-thinking or final-answer marker, together with all subsequent tokens, is protected and receives weight 1 regardless of repetition or predicted roles.

The default configuration reduces the length penalty for chunks containing R while leaving E-only and T-only chunks unchanged. For example, an R-only chunk receives weight 0.9, an RE chunk receives 0.95, and a chunk containing all three roles receives $29/30$. Exact repetition receives weight 1.1. For the T~classifier, the chunk weight is broadcast to its original token positions. For the H~classifier, the singleton set gives $w_{it}=\lambda_{c_{it}}$: R receives 0.9, while E and T receive 1, before repetition and protection rules. Mixed labels RE, RT, ET, and RET occur only with the T~classifier; M is their aggregate reporting category, not \mbox{a separately weighted class.}

Rather than scaling the complete token reward, PCRS adjusts only the existing penalty gap $\Delta_{it}=u_i-q^D_{it}$. A position is eligible for adjustment when $c_i>0$, $m_{it}=1$, $\Delta_{it}>10^{-7}$, and $w_{it}\neq 1$. Let $e_{it}$ denote these conditions. We set
\begin{equation}
q^{S}_{it}=
\begin{cases}
u_i-w_{it}\Delta_{it},
& e_{it},\\
q^{D}_{it}, & \text{otherwise}.
\end{cases}
\label{eq:sret}
\end{equation}

Thus, PCRS rescales the existing penalty gap by $w_{it}$: values below 1 reduce the penalty, while values above 1 increase it. In the default configuration, R-only chunks receive a 10\% penalty reduction, RE chunks receive a 5\% reduction, and exact repetitions receive a 10\% increase. Incorrect responses, non-penalized positions, protected regions, and padding retain their original raw rewards. The configured weights lie in $[0.9,1.1]$.

\paragraph{Policy optimization.}
The adjusted token rewards are normalized across responses to the same prompt before updating the policy. The implemented group normalization computes its moments over the initialized padded reward arrays and applies the valid-response mask afterward. Consequently, changing a reward at one position can also affect the normalized advantages of other responses in the group. The policy is trained using proximal policy optimization (PPO) \citep{schulman2017ppo} with a dual-clipped token-mean loss \citep{ye2020dualclip} (dual-clip constant 3.0), one optimization epoch, and no explicit KL or entropy regularization. Appendix~\ref{app:implementation} provides the complete normalization and optimization objectives.

At each training update, PCRS samples responses, constructs the underlying token rewards, predicts chunk roles, detects exact repetitions, and applies Equation~\ref{eq:sret} at eligible positions before normalization and policy optimization. We separately record role-label distributions over all valid response tokens and the number of positions whose raw rewards are changed. Additional reward-weight configurations \mbox{are described in Appendix~\ref{app:weight_results}.}

\subsection{Reward-Weight Settings within PCRS-T}
Within PCRS-T, we compare four settings of $(\lambda_R,\lambda_E,\lambda_T)$: Default $(0.9,1,1)$, E-priority $(0.9,0.95,1.05)$, T-priority $(0.9,1.05,0.95)$, and R-milder $(0.95,1,1)$. They share the frozen T~classifier, threshold 0.5, label averaging, repetition multiplier 1.1, and protected-region rules. See \mbox{Appendix~\ref{app:weight_results} for detailed results.}

\section{Experimental Setup}
\label{sec:setup}

\paragraph{Training setup.}
All checkpoints are initialized from DeepSeek-R1-Distill-Qwen-1.5B \citep{deepseek2025r1}. We use a fixed quarter-data subset of DeepScaleR \citep{luo2025deepscaler}, the training dataset used by DECS \citep{jiang2025decs}. This subset contains 10,112 training and 492 validation examples. Our standard training setup uses four A100-80GB actor GPUs and one A100 judge, with two data epochs (52 optimizer updates). We use FSDP \citep{zhao2023fsdp}, a learning rate of $2\times10^{-6}$, batch size 128, and four rollouts per prompt. The maximum prompt and response lengths are 1,024 and 16,384 tokens, respectively. Group filtering and adaptive curriculum scheduling are enabled.

Our evaluated configurations comprise DECS, PCRS-H, and four weight settings for PCRS-T: Default, E-priority, T-priority, and R-milder. The primary comparison evaluates default PCRS-T against DECS across five paired training seeds, using the same initial model and data split. The comparison between PCRS-H and PCRS-T uses the shared reward interface with exclusive and overlapping phase labels, respectively (Section~\ref{sec:role_routes}); the weight comparison stays within PCRS-T. For the component ablation, repetition-only retains the exact-repetition multiplier of $1.1$ but sets every phase multiplier to $1.0$, keeping the classifier, chunk boundaries, and protection rules unchanged. Full configurations and checkpoint provenance appear in Appendix~\ref{app:provenance} and Table~\ref{tab:config}.

\paragraph{Benchmarks and evaluation.}
We evaluate on six benchmarks totaling 1,379 questions: MATH500 (500) \citep{lightman2023verify,hendrycks2021math}, AIME 2024 (30), AIME 2025 (30), AMC23 (40) \citep{maaAMC}, OlympiadBench (581) \citep{he2024olympiad}, and GPQA-Diamond (198) \citep{rein2023gpqa}. These are the fixed evaluation subsets used in our experiments.

Each checkpoint generates 16 responses per question using vLLM 0.11.2 \citep{kwon2023vllm} with the \texttt{instruct\_default} prompt. Training uses temperature 0.6 and top-$p=1.0$; evaluation uses temperature 0.6 and top-$p=0.95$.

\subsection{Evaluation Metrics}
\label{sec:metrics}

We primarily measure sampled Pass@1 and mean response length, with oracle Pass@16 as a supplementary measure of multi-attempt coverage. For question $i$, let $c_{ij}\in\{0,1\}$ indicate whether sampled response $j$ is correct, and let $\ell_{ij}$ denote its generated token count, including reasoning and the final answer. With 16 responses per question, we compute
$\widehat{\mathrm{Pass@1}}_i
=\frac{1}{16}\sum_{j=1}^{16}c_{ij}$, 
$\overline{\mathrm{Tokens}}_i
=\frac{1}{16}\sum_{j=1}^{16}\ell_{ij}$, 
$\widehat{\mathrm{Pass@16}}_i =\mathbb{I}\left[\sum_{j=1}^{16}c_{ij}>0\right]$.
Pass@1 estimates single-response success, while oracle Pass@16 measures whether at least one of the 16 sampled responses is correct \citep{chen2021code}. Mean tokens include correct, \mbox{incorrect, and truncated responses.}

We aggregate results using both question-weighted and benchmark-equal averages. Paired uncertainty is estimated using 10,000 question-bootstrap draws, retaining all 16 responses per question; multi-benchmark comparisons use benchmark-stratified resampling.

\paragraph{Case studies.}
We compare reasoning traces on the same questions, reporting selected responses alongside correctness and mean length over all 16 samples. Main cases select the first response from the largest original answer group; supplementary cases use deterministically selected samples, independently of correctness or length. Selection protocols and additional examples are provided in Appendices~\ref{app:caseprotocol} and~\ref{app:trchecks}.
\section{Results}
\label{sec:results}
\subsection{Classification Routes and Overall Results}
On all 1,379 questions across six benchmarks, overall sampled Pass@1 is approximately 62.07\% for both \mbox{PCRS-T} and our DECS reproduction; \mbox{PCRS-T} reduces mean response length from 7,187.80 to 6,962.68 tokens. Supplementary oracle Pass@16 increases from 86.08\% to 87.67\%. Table~\ref{tab:primary_full} reports paired differences under both aggregation schemes: both show fewer tokens and higher coverage. Appendix~\ref{app:fullresults} reports the route and weight configurations and per-benchmark results, and Appendix~\ref{app:fixed_budget} \mbox{describes the common-output-budget replay.}

\begin{table*}[t]\centering\small
\begin{tabular}{lrrr}\toprule
Full six-benchmark suite & $\Delta$ Pass@1 (points) & $\Delta$ Pass@16 (points) & $\Delta$ mean tokens\\\midrule
Question-weighted & $-0.005\;[-0.535,0.539]$ & $+1.595\;[0.435,2.828]$ & $-225.12\;[-275.06,-174.74]$\\
Six-benchmark-equal & $-0.23\;[-1.32,0.84]$ & $+3.71\;[0.42,7.13]$ & $-205.19\;[-312.97,-101.69]$\\\bottomrule
\end{tabular}
\caption{Default \mbox{PCRS-T} minus quarter-DECS on all 1,379 questions across six benchmarks. Brackets give 95\% benchmark-stratified paired question-bootstrap intervals. Pass@16 is oracle coverage over 16 responses.}
\label{tab:primary_full}\end{table*}

\mbox{PCRS-H} records 62.07\% sampled Pass@1, 7,106.23 mean tokens, and 86.80\% Pass@16. Default \mbox{PCRS-T} records 62.07\%, 6,962.68 tokens, and 87.67\%, respectively. Thus, default \mbox{PCRS-T} uses fewer tokens and reaches higher multi-attempt coverage. The classification signals for this comparison are specified in Section~\ref{sec:role_routes}; both routes use the shared penalty-gap reward interface.

\begin{figure*}[t]\centering
\resizebox{\textwidth}{!}{%
\begin{tikzpicture}[x=1cm,y=1cm,font=\rmfamily\fontsize{10}{12}\selectfont]
\definecolor{summaryBlue}{HTML}{397DB5}
\definecolor{summaryGold}{HTML}{B78429}
\definecolor{summaryPink}{HTML}{AA6187}
\begin{scope}[xshift=0.8cm]
\node[anchor=west,font=\bfseries] at (0,5.73) {(a) Accuracy--output trade-off};
\node[rotate=90] at (-1.50,3.35) {Pass@1 (\%)};
\draw[black!12,line width=.35pt] (0.000,1.25) -- (0.000,5.500);
\node[anchor=north] at (0.000,1.12) {6,900};
\draw[black!12,line width=.35pt] (1.939,1.25) -- (1.939,5.500);
\node[anchor=north] at (1.939,1.12) {7,000};
\draw[black!12,line width=.35pt] (3.879,1.25) -- (3.879,5.500);
\node[anchor=north] at (3.879,1.12) {7,100};
\draw[black!12,line width=.35pt] (5.818,1.25) -- (5.818,5.500);
\node[anchor=north] at (5.818,1.12) {7,200};
\draw[black!12,line width=.35pt] (0,1.250) -- (6.4,1.250);
\node[anchor=east] at (-.14,1.250) {61.5};
\draw[black!12,line width=.35pt] (0,2.174) -- (6.4,2.174);
\node[anchor=east] at (-.14,2.174) {61.75};
\draw[black!12,line width=.35pt] (0,3.098) -- (6.4,3.098);
\node[anchor=east] at (-.14,3.098) {62};
\draw[black!12,line width=.35pt] (0,4.022) -- (6.4,4.022);
\node[anchor=east] at (-.14,4.022) {62.25};
\draw[black!12,line width=.35pt] (0,4.946) -- (6.4,4.946);
\node[anchor=east] at (-.14,4.946) {62.5};
\draw[black!50,line width=.5pt] (0,5.5) -- (0,1.25) -- (6.4,1.25);
\node at (3.2,.48) {Mean response tokens};
\draw[draw=black!65,fill=black!65,line width=1.1pt,rounded corners=.4pt] (5.500,3.275) rectangle (5.664,3.439);
\node[anchor=center,text=black!65,inner sep=1.5pt,align=center,font=\fontsize{9}{10}\selectfont] at (5.85,3.68) {DECS};
\draw[draw=summaryPink,fill=summaryPink,line width=1.1pt,line join=round] (4.000,3.477) -- (4.105,3.357) -- (4.000,3.237) -- (3.895,3.357) -- cycle;
\node[anchor=center,text=summaryPink,inner sep=1.5pt,align=center,font=\fontsize{9}{10}\selectfont] at (4.0,2.98) {\mbox{PCRS-H}};
\draw[draw=summaryBlue,fill=summaryBlue,line width=1.1pt] (1.216,3.357) circle (.09);
\node[anchor=center,text=summaryBlue,inner sep=1.5pt,align=center,font=\fontsize{9}{10}\selectfont] at (1.15,3.89) {\mbox{PCRS-T}\\(Default)};
\draw[draw=summaryGold,fill=summaryGold,line width=1.1pt,line join=round] (4.937,1.813) -- (5.049,1.618) -- (4.825,1.618) -- cycle;
\node[anchor=center,text=summaryGold,inner sep=1.5pt,align=center,font=\fontsize{9}{10}\selectfont] at (4.7,2.20) {\mbox{PCRS-T}\\(E-priority)};
\draw[draw=summaryBlue,fill=white,line width=1.1pt,line join=round] (3.169,4.437) -- (3.281,4.242) -- (3.057,4.242) -- cycle;
\node[anchor=center,text=summaryBlue,inner sep=1.5pt,align=center,font=\fontsize{9}{10}\selectfont] at (4.25,4.25) {\mbox{PCRS-T}\\(T-priority)};
\draw[draw=summaryBlue,fill=white,line width=1.1pt,line join=round] (2.886,4.807) -- (2.991,4.687) -- (2.886,4.567) -- (2.781,4.687) -- cycle;
\node[anchor=center,text=summaryBlue,inner sep=1.5pt,align=center,font=\fontsize{9}{10}\selectfont] at (1.95,4.98) {\mbox{PCRS-T}\\(R-milder)};
\end{scope}
\begin{scope}[xshift=8.850000000000001cm]
\node[anchor=west,font=\bfseries] at (0,5.73) {(b) Coverage--output trade-off};
\node[rotate=90] at (-1.50,3.35) {Pass@16 (\%)};
\draw[black!12,line width=.35pt] (0.000,1.25) -- (0.000,5.500);
\node[anchor=north] at (0.000,1.12) {6,900};
\draw[black!12,line width=.35pt] (1.939,1.25) -- (1.939,5.500);
\node[anchor=north] at (1.939,1.12) {7,000};
\draw[black!12,line width=.35pt] (3.879,1.25) -- (3.879,5.500);
\node[anchor=north] at (3.879,1.12) {7,100};
\draw[black!12,line width=.35pt] (5.818,1.25) -- (5.818,5.500);
\node[anchor=north] at (5.818,1.12) {7,200};
\draw[black!12,line width=.35pt] (0,1.804) -- (6.4,1.804);
\node[anchor=east] at (-.14,1.804) {86};
\draw[black!12,line width=.35pt] (0,2.728) -- (6.4,2.728);
\node[anchor=east] at (-.14,2.728) {86.5};
\draw[black!12,line width=.35pt] (0,3.652) -- (6.4,3.652);
\node[anchor=east] at (-.14,3.652) {87};
\draw[black!12,line width=.35pt] (0,4.576) -- (6.4,4.576);
\node[anchor=east] at (-.14,4.576) {87.5};
\draw[black!12,line width=.35pt] (0,5.500) -- (6.4,5.500);
\node[anchor=east] at (-.14,5.500) {88};
\draw[black!50,line width=.5pt] (0,5.5) -- (0,1.25) -- (6.4,1.25);
\node at (3.2,.48) {Mean response tokens};
\draw[draw=black!65,fill=black!65,line width=1.1pt,rounded corners=.4pt] (5.500,1.870) rectangle (5.664,2.034);
\node[anchor=center,text=black!65,inner sep=1.5pt,align=center,font=\fontsize{9}{10}\selectfont] at (5.72,2.30) {DECS};
\draw[draw=summaryPink,fill=summaryPink,line width=1.1pt,line join=round] (4.000,3.403) -- (4.105,3.283) -- (4.000,3.163) -- (3.895,3.283) -- cycle;
\node[anchor=center,text=summaryPink,inner sep=1.5pt,align=center,font=\fontsize{9}{10}\selectfont] at (4.60,3.65) {\mbox{PCRS-H}};
\draw[draw=summaryBlue,fill=summaryBlue,line width=1.1pt] (1.216,4.890) circle (.09);
\node[anchor=center,text=summaryBlue,inner sep=1.5pt,align=center,font=\fontsize{9}{10}\selectfont] at (2.0,5.00) {\mbox{PCRS-T}\\(Default)};
\draw[draw=summaryGold,fill=summaryGold,line width=1.1pt,line join=round] (4.937,1.924) -- (5.049,1.729) -- (4.825,1.729) -- cycle;
\node[anchor=center,text=summaryGold,inner sep=1.5pt,align=center,font=\fontsize{9}{10}\selectfont] at (3.75,2.10) {\mbox{PCRS-T}\\(E-priority)};
\draw[draw=summaryBlue,fill=white,line width=1.1pt,line join=round] (3.169,3.144) -- (3.281,2.949) -- (3.057,2.949) -- cycle;
\node[anchor=center,text=summaryBlue,inner sep=1.5pt,align=center,font=\fontsize{9}{10}\selectfont] at (1.90,2.68) {\mbox{PCRS-T}\\(T-priority)};
\draw[draw=summaryBlue,fill=white,line width=1.1pt,line join=round] (2.886,3.532) -- (2.991,3.412) -- (2.886,3.292) -- (2.781,3.412) -- cycle;
\node[anchor=center,text=summaryBlue,inner sep=1.5pt,align=center,font=\fontsize{9}{10}\selectfont] at (1.62,3.65) {\mbox{PCRS-T}\\(R-milder)};
\end{scope}
\end{tikzpicture}}
\caption{Accuracy--output and coverage--output trade-offs. Points show question-weighted sampled Pass@1 and oracle Pass@16 for DECS, PCRS-H, and four PCRS-T weight settings on 1,379 questions, with 16 responses per question. Paired intervals are reported in Appendix~\ref{app:paired_uncertainty}.}
\label{fig:current_results}\end{figure*}
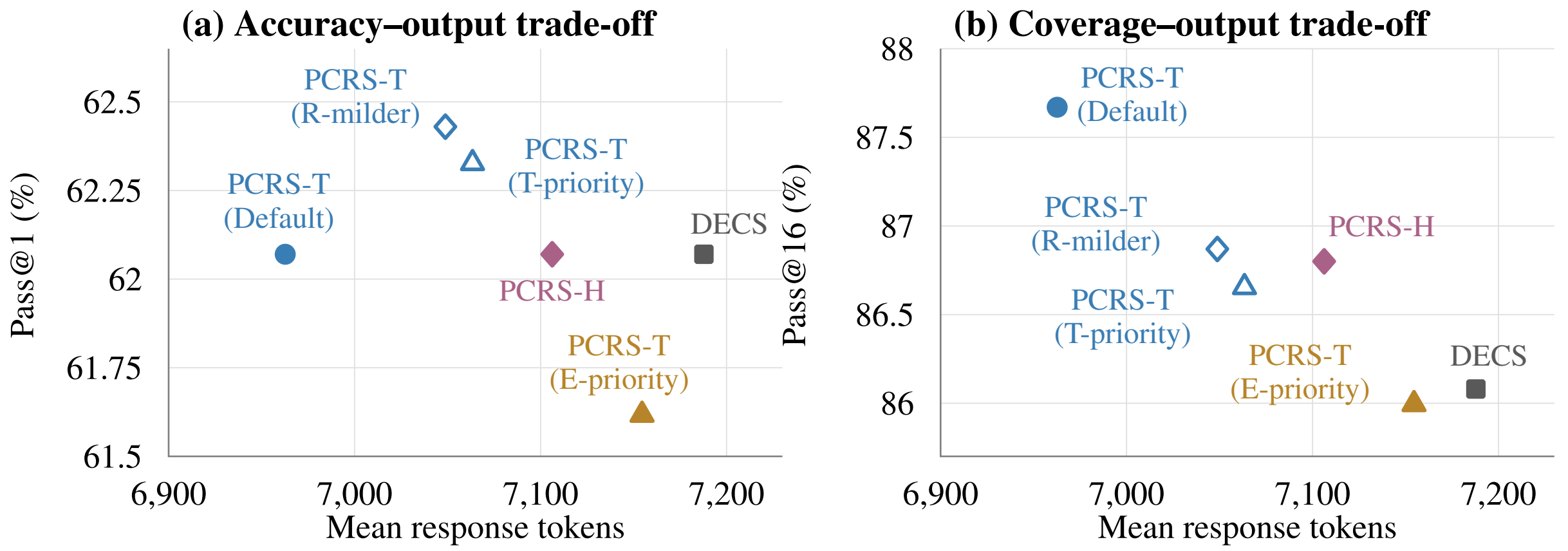

Figure~\ref{fig:current_results} places the six route and weight configurations on the same accuracy--output and coverage--output axes. Default \mbox{PCRS-T} has the lowest mean token use and highest Pass@16 point estimate; T-priority and R-milder trade longer responses for higher Pass@1 point estimates.

\paragraph{Variation across benchmarks.}
The largest mean token reductions occur on MATH500 and OlympiadBench. Sampled Pass@1 point estimates are higher on AIME 2024 and OlympiadBench, while the magnitude and direction of accuracy differences vary across benchmarks. Table~\ref{tab:results} reports all benchmark values, and Figure~\ref{fig:paired_evidence} gives paired intervals. Benchmark-equal Pass@1 is 49.53\% for \mbox{PCRS-T} and 49.75\% for DECS.

\subsection{Component Ablation}\label{sec:component_ablation}
The component comparison tests whether exact-repetition penalties alone account for the gains of default \mbox{PCRS-T}. Table~\ref{tab:repeat_control} compares DECS, repetition-only, and the full configuration on the same six benchmarks. Repetition-only produces 7,177.99 mean tokens and 86.15\% oracle Pass@16, close to DECS's 7,187.80 tokens and 86.08\%. Default \mbox{PCRS-T} uses 215.30 fewer tokens per response than repetition-only (3.00\%) and gains 1.52 percentage points in Pass@16. The paired 95\% intervals are $[-267.66,-161.96]$ tokens and $[0.36,2.68]$ points. The sampled Pass@1 difference is $+0.29$ points, with interval $[-0.23,0.82]$.

\begin{table}[t]\centering\small
\setlength{\tabcolsep}{4pt}
\begin{tabular}{lrrr}\toprule
Configuration & Pass@1 & Tokens & Pass@16\\\midrule
DECS & 62.07 & 7,187.80 & 86.08\\
Repetition-only & 61.77 & 7,177.99 & 86.15\\
\mbox{PCRS-T} & 62.07 & 6,962.68 & 87.67\\\bottomrule
\end{tabular}
\caption{Component ablation on 1,379 questions, with 16 responses per question. Pass@1 and Pass@16 are percentages; tokens denote mean response length. See the control definition and settings in Appendix~\ref{app:repeat_control}.}
\label{tab:repeat_control}\end{table}

Repetition handling alone does not recover the full configuration's observed output-length and coverage gains. The comparison supports using phase-conditioned allocation alongside repetition handling; per-benchmark results and paired differences are reported in Appendix~\ref{app:repeat_control}.

\begin{figure*}[t]
\centering
\input{figures/pcrs_case_compact}
\caption{Q211 (OlympiadBench): selected DECS and default PCRS-T response excerpts. Both answers are correct. Author annotations mark an error (orange), repeated checking (purple), and valid checks or repair (blue); [\ldots] marks omitted text.}
\label{fig:case_compact}
\end{figure*}

\subsection{Reward Weights within \mbox{PCRS-T}}
Among the four completed \mbox{PCRS-T} settings, the default setting has the lowest full-suite mean token use and the highest oracle Pass@16. T-priority and R-milder have higher full-suite Pass@1 point estimates and generate 100.74 and 86.11 additional tokens per response, respectively. Figure~\ref{fig:current_results} summarizes these operating points; Figure~\ref{fig:weight_results} in Appendix~\ref{app:weight_results} gives the metric-by-metric comparisons, with all changed weight factors \mbox{recorded alongside each configuration.}

Across the four settings, R increases by 0.35--1.03 percentage points from the early to late windows; M decreases for Default and T-priority but increases for E-priority and R-milder. Appendix~\ref{app:rolechanges} gives full label histories; Appendix~\ref{app:implementation} details reward normalization.

\subsection{Coverage at a common output budget}
Table~\ref{tab:fixed_budget} reports all six per-question output budgets on the full six-benchmark suite. Replay processes every response in its original generation order and counts a success only when the response finishes within the budget. An overflowing response consumes the remaining budget. We evaluate oracle coverage using up to 16 \mbox{cached responses per question.}

On the full six-benchmark suite, \mbox{PCRS-T} has higher coverage point estimates at five of the six tested budgets: the difference is $+2.32$ points at 8,192 tokens and $+1.60$ points at 262,144 tokens; at 65,536 tokens it is $-0.15$ points. Table~\ref{tab:fixed_budget} reports paired intervals at every budget, and Appendix~\ref{app:fixed_budget} specifies the replay protocol.

\begin{table}[t]
\centering\small\setlength{\tabcolsep}{3pt}
\begin{tabular}{@{}rrrr@{}}\toprule
Budget & \multicolumn{2}{c}{Coverage (\%)} & \shortstack{\mbox{PCRS-T} $-$ DECS}\\
(tokens) & DECS & \mbox{PCRS-T} & pp [95\% interval]\\\midrule
8,192 & 53.30 & 55.62 & $2.32\;[0.29,4.42]$\\
16,384 & 65.26 & 67.15 & $1.89\;[-0.22,3.99]$\\
32,768 & 74.04 & 75.27 & $1.23\;[-0.58,3.05]$\\
65,536 & 80.49 & 80.35 & $-0.15\;[-1.67,1.38]$\\
131,072 & 84.55 & 85.64 & $1.09\;[-0.22,2.39]$\\
262,144 & 86.08 & 87.67 & $1.60\;[0.36,2.83]$\\
\bottomrule\end{tabular}
\caption{Question-weighted oracle coverage at shared output-token budgets on 1,379 questions. Differences are default \mbox{PCRS-T} minus DECS in percentage points, with paired 95\% intervals.}
\label{tab:fixed_budget}
\end{table}

\subsection{Scientific reasoning}
On GPQA-Diamond, default \mbox{PCRS-T} records sampled Pass@1 of 32.70\% versus 34.09\% for DECS. Oracle coverage is 183/198 versus 182/198, with a mean-length difference of $-5.45$ tokens. The paired 95\% interval for the Pass@1 difference is $[-3.25,0.54]$ percentage points.

\begin{samepage}
T-priority records 35.51\% GPQA Pass@1 with 7,874.82 mean tokens; default \mbox{PCRS-T} records 32.70\% with 8,043.49 tokens.\par
\end{samepage}

\subsection{A closer look at verification}
\label{sec:cases}
Figure~\ref{fig:case_compact} contrasts DECS's repeated checks after a substitution error with default \mbox{PCRS-T}'s direct candidate verification on Q211. Both selected answers are correct. Across all 16 samples, \mbox{PCRS-T} answers 16 correctly versus 13 for DECS, with a mean length of 6,692 versus 7,685 tokens. See Figure~\ref{fig:atlasQ211} for all four configurations.
\section{Discussion}
\paragraph{Less repetition after an error.}
Q211 in Figure~\ref{fig:case_compact} shows what can differ within a shorter CoT. After swapping the variables during substitution, DECS repeatedly revisits the resulting contradiction before correcting the assignment. Default PCRS-T keeps the variables correctly assigned and proceeds to enumerate and check the candidate pairs. The shorter response avoids the extended detour around the substitution error, using 2,754 tokens compared with 9,559 for DECS.

\paragraph{Retaining essential checks.}
The shorter PCRS-T response still derives the constraint, checks all six candidate pairs, and identifies the maximum. These steps establish why the final answer is valid. In this example, the change in CoT is a more direct path from the constraint to candidate verification, with fewer tokens spent revisiting an earlier mistake. This suggests a useful goal for efficient reasoning: reduce repeated work while retaining checks that establish the solution.
\section{Conclusion}
We introduce PCRS, a framework that separates phase classification from training-time reward allocation. We investigate PCRS-H and PCRS-T, and compare four reward-weight settings within PCRS-T. Across six benchmarks, default PCRS-T reduces response length while achieving similar sampled single-response accuracy and higher oracle multi-attempt coverage than our DECS reproduction. The repetition-only control does not recover the full configuration's output-length and coverage gains. The classifier is used only during training. These findings identify phase-informed penalty allocation as a practical direction for improving reasoning efficiency.%
\clearpage
\section*{Limitations}
The checkpoint comparison evaluates the classifier and reward rule jointly. E-priority and T-priority each change two weight factors, so these comparisons characterize configurations rather than isolate individual weight effects. Training phase-label curves describe exclusive buckets on changing batches before reward eligibility; missing per-step denominators and pooled multilabel buckets restrict marginal-phase analysis. Trace cases examine selected mathematical decisions and do not estimate mechanism prevalence or training causality. Appendix~\ref{app:provenance} records the available provenance, and \mbox{Appendix~\ref{app:gaps} states artifact availability.}
\section*{Ethical Considerations}
The study evaluates benchmark reasoning outputs. Phase annotations describe textual function, and application decisions require task-specific validation of answer quality. Artifact redistribution must respect the licenses of the datasets, models, and traces. Numerical results are computed from stored outputs. Quoted \mbox{excerpts mark omissions.}
\FloatBarrier
\bibliography{references}
\clearpage
\appendix
\makeatletter
\setlength{\@fptop}{0pt}
\setlength{\@dblfptop}{0pt}
\setlength{\@fpsep}{14pt plus 2pt minus 2pt}
\setlength{\@dblfpsep}{14pt plus 2pt minus 2pt}
\makeatother
\setcounter{dbltopnumber}{4}
\renewcommand{\dbltopfraction}{0.9}
\section{Implementation and Reproducibility}
\label{app:implementation}
\subsection{PCRS-H: Labels and Shared Rewards}
\label{app:legacy_proxy}
\paragraph{Phase-signal sources.}
The H~classifier uses normalized changes between adjacent hidden states to assign one R/E/T label per response position. The T~classifier uses native chunk text and preceding context to predict overlapping labels. Table~\ref{tab:role_routes} summarizes the two routes and \mbox{their shared reward interface.}

\begin{table*}[t]
\centering\small
\setlength{\tabcolsep}{5pt}
\renewcommand{\arraystretch}{1.16}
\begin{tabular}{@{}p{0.15\textwidth}p{0.385\textwidth}p{0.385\textwidth}@{}}
\toprule
Property & PCRS-H & PCRS-T\\
\midrule
Input & Adjacent hidden-state changes, normalized within the input sequence & Native chunk text and preceding context; frozen word/character TF--IDF features\\
Granularity & Response-token-aligned labels & Native-chunk labels broadcast to original token positions\\
Labels & Mutually exclusive R/E/T; thresholds 1.0 and 1.4 & Overlapping R/E/T from three logistic heads; threshold 0.5; empty sets allowed\\
Phase weight & Weight of the single predicted role & Mean weight of the active roles\\
Compute required & Hidden-state extraction, adjacent differences, and normalization & CPU text features and sparse linear heads\\
Reward insertion & Before group normalization: scale eligible gaps & Before group normalization: scale eligible gaps\\
\bottomrule
\end{tabular}
\caption{Two phase-classification routes sharing reward allocation, repetition detection, and protection rules.}
\label{tab:role_routes}
\end{table*}

\paragraph{Offline attention-sparsity analysis.}
We examined the attention-sparsity calibration strategy of ThinKV \citep{ramachandran2026thinkv} on DeepSeek-R1-Distill-Qwen-1.5B. The analysis used 100 MATH-source calibration questions and 50 held-out questions, one response per question with a 4,096-token cap, and all 28 layers. For each question and layer, we estimated the sparsity distribution and examined its modes and intervening minima. At the primary KDE bandwidth of 0.02, the largest three-mode coverage was 28 of 100 calibration questions; the criterion of a common three-mode layer across calibration questions yielded no selected layers. This offline analysis used attention sparsity; the evaluated H~classifier training configuration uses the \mbox{hidden-state statistic defined below.}

\paragraph{Three-class signal.}
Let $h_s$ denote the hidden-state vector used by PCRS-H at input position $s$. As described in Section~\ref{sec:role_routes}, the normalized change between adjacent positions is
\begin{equation}
\begin{aligned}
d_1&=0,\quad d_s=\|h_s-h_{s-1}\|_2,\\
z_s&=\frac{d_s}{\max(\operatorname{mean}_{s'}d_{s'},10^{-6})}.
\end{aligned}
\end{equation}
We set the two ordered thresholds to $\theta_1=1.0$ and $\theta_2=1.4$. Following the mode-and-valley threshold-selection strategy of ThinKV \citep{ramachandran2026thinkv}, we inspected the three peaks and the two intervening valleys of the classification-signal distribution and selected these two valley-based cutoffs for the H~classifier. The classifier assigns
\begin{equation}
c_s=\begin{cases}
E,&z_s<\theta_1,\\
R,&\theta_1\le z_s<\theta_2,\\
T,&z_s\ge\theta_2.
\end{cases}
\end{equation}
The prediction at the position preceding each response token supplies its label. The resulting singleton set $A_{it}=\{c_{it}\}$ enters Equation~\ref{eq:role_weight}; the phase multiplier reduces to $\lambda_{c_{it}}$. The T~classifier uses the same equation with its \mbox{potentially overlapping label set.}

\paragraph{Shared reward integration.}
Both routes adjust the penalty gap $\Delta_{it}=u_i-q^D_{it}$ at eligible token positions using Equation~\ref{eq:sret}. The default phase weights are $(\lambda_R,\lambda_E,\lambda_T)=(0.9,1,1)$. Protected positions receive weight 1; elsewhere, exact repetition takes precedence and receives weight 1.1. Group normalization and policy optimization follow the reward adjustment for both routes. Thus, exclusive classification is the singleton case of the shared allocation rule, while the T~classifier additionally represents mixed-label combinations.

\subsection{PCRS-T: text classifier settings}
Features use sublinear term frequency and per-view $\ell_2$ normalization, with at most 15,000 word and 20,000 character features. Each head uses balanced class weights, $C=1$, the LIBLINEAR solver \citep{fan2008liblinear}, a 2,000-iteration cap, and random state 20260916; only the 300 training chunks are fitted. Inference uses the frozen coefficients. Head scores are uncalibrated phase-membership estimates.

In the export audit, portable and direct inference probabilities agree within $1.9\times10^{-15}$.

\paragraph{Author inspection of annotations.}
Following model annotation, an author manually reviewed the phase-annotation batches. The archived evaluation scores below retain the original model-reference labels and aggregation rules. The reference protocols and question-cluster uncertainty are specified for each cohort.

\paragraph{Held-out model-reference evaluation.}
\label{app:heldout_roles}
With the classifier and threshold 0.5 frozen, we evaluated 585 archived chunks from 100 held-out MATH500 questions and 200 DECS/PCRS-H responses. Neither questions nor exact chunk texts overlap the 120-question training or 30-question development pools. Three same-family model-annotation contexts assigned primary R/E/T labels without seeing classifier predictions; 60 chunks were double-annotated. Primary labels were fixed before scoring and retained through author inspection. Archived responses were re-tokenized, so this evaluation tests label agreement rather than original generation-token alignment.

Macro-F1 is 71.67\% (question-cluster bootstrap 95\% interval 66.88--75.34\%), micro-F1 80.66\%, and exact label-set agreement 48.38\% (283/585). Table~\ref{tab:heldout_roles} gives phase-level results. T is the main weakness: 27 true positives, 35 false negatives, and 20 false positives. Double annotation yields 53/60 exact agreements and macro-F1 of 71.61\% when secondary labels replace the corresponding primary labels.

A supplementary blinded audit used three fresh model-annotation contexts on 60 randomly sampled and 40 diagnostic chunks. Full label sets agree on 92/100 chunks. Against per-phase majority labels, macro-F1 is 66.98\% on the random subset, which contains only seven T-positive chunks. Nineteen of 20 diagnostic T false negatives remain T by majority vote. Eight disputed boundary cases remain unresolved; this audit does not replace the frozen 585-chunk reference.

\begin{table}[t]
\centering\small
\begin{tabular}{lrrrr}\toprule
Phase & Precision & Recall & F1 & Support\\\midrule
R & 83.17 & 83.80 & 83.48 & 395\\
E & 79.42 & 84.73 & 81.99 & 419\\
T & 57.45 & 43.55 & 49.54 & 62\\\bottomrule
\end{tabular}
\caption{Frozen classifier agreement on 585 held-out chunks against blinded model reference labels. Scores are percentages; support counts reference-positive chunks. Labels may overlap.}
\label{tab:heldout_roles}
\end{table}

\paragraph{Cross-domain model-reference evaluation.}
\label{app:crossdomain_roles}
The classifier uses 300 training chunks from 120 MATH questions and 96 development chunks from 30 separate questions, all from base-model responses. Cross-domain evaluation fixes one response and one native chunk per question for 180 questions: 20 each from MATH500, AIME 2024, AIME 2025, AMC23, and OlympiadBench, plus 80 from GPQA-Diamond. Each benchmark balances DECS and default PCRS responses. Selection is independent of correctness and classifier output; weights, vocabulary, threshold, and rubric remain frozen. Normalized question texts and exact chunk texts exclude the training/development pools and earlier held-out cohort, without testing semantic near-duplicates. Chunk boundaries and preceding contexts were replayed against the source responses.

Three fresh same-family model-annotation contexts independently labeled every chunk, seeing the question and chunk context but no predictions, model identities, correctness, or other labels. The pre-specified reference uses a two-of-three majority per phase, followed by author review. Complete label sets agree on 122/180 chunks; all 58 disagreements are retained.

\begin{table}[t]
\centering\small\setlength{\tabcolsep}{4pt}
\begin{tabular}{lrrrrrr}\toprule
Cohort & $n$ & Macro & R & E & T & Exact\\\midrule
Math & 100 & 80.04 & 88.57 & 89.47 & 62.07 & 58.00\\
GPQA & 80 & 82.43 & 96.69 & 71.43 & 79.17 & 45.00\\\bottomrule
\end{tabular}
\caption{Frozen cross-domain classifier agreement (\%). Macro is the mean of R/E/T F1; Exact requires the complete predicted label set to match the model-reference set. One chunk is sampled per question.}
\label{tab:crossdomain_roles}
\end{table}

\begin{figure*}[t]\centering
\resizebox{\textwidth}{!}{\begin{tikzpicture}[x=1cm,y=1cm,font=\fontsize{9}{11}\selectfont]
\definecolor{evB}{HTML}{777777}
\definecolor{evT}{HTML}{4477AA}
\definecolor{evS}{HTML}{008F82}
\definecolor{evE}{HTML}{CC6677}
\definecolor{evMath}{HTML}{4477AA}
\definecolor{evGPQA}{HTML}{B47716}
\draw[draw=black!12,line width=.35pt] (1.4500,0.0000) -- (1.4500,3.1500);
\node[inner sep=1pt,] at (1.4500,-0.2400) {40};
\draw[draw=black!12,line width=.35pt] (3.3333,0.0000) -- (3.3333,3.1500);
\node[inner sep=1pt,] at (3.3333,-0.2400) {60};
\draw[draw=black!12,line width=.35pt] (5.2167,0.0000) -- (5.2167,3.1500);
\node[inner sep=1pt,] at (5.2167,-0.2400) {80};
\draw[draw=black!12,line width=.35pt] (7.1000,0.0000) -- (7.1000,3.1500);
\node[inner sep=1pt,] at (7.1000,-0.2400) {100};
\draw[draw=black!55,line width=.45pt] (1.4500,0.0000) -- (7.1000,0.0000);
\node[inner sep=1pt,anchor=east] at (1.3200,2.6000) {R};
\draw[draw=evMath,line width=.8pt] (5.4716,2.7500) -- (6.4989,2.7500);
\draw[draw=evMath,line width=.7pt] (5.4716,2.7050) -- (5.4716,2.7950);
\draw[draw=evMath,line width=.7pt] (6.4989,2.7050) -- (6.4989,2.7950);
\fill[evMath] (6.0238,2.7500) circle (1.6pt);
\draw[draw=evGPQA,line width=.8pt] (6.4722,2.4500) -- (7.0392,2.4500);
\draw[draw=evGPQA,line width=.7pt] (6.4722,2.4050) -- (6.4722,2.4950);
\draw[draw=evGPQA,line width=.7pt] (7.0392,2.4050) -- (7.0392,2.4950);
\fill[evGPQA] (6.7362,2.3980) rectangle (6.8402,2.5020);
\node[inner sep=1pt,anchor=east] at (1.3200,1.5800) {E};
\draw[draw=evMath,line width=.8pt] (5.5854,1.7300) -- (6.5461,1.7300);
\draw[draw=evMath,line width=.7pt] (5.5854,1.6850) -- (5.5854,1.7750);
\draw[draw=evMath,line width=.7pt] (6.5461,1.6850) -- (6.5461,1.7750);
\fill[evMath] (6.1088,1.7300) circle (1.6pt);
\draw[draw=evGPQA,line width=.8pt] (3.2824,1.4300) -- (5.3481,1.4300);
\draw[draw=evGPQA,line width=.7pt] (3.2824,1.3850) -- (3.2824,1.4750);
\draw[draw=evGPQA,line width=.7pt] (5.3481,1.3850) -- (5.3481,1.4750);
\fill[evGPQA] (4.3575,1.3780) rectangle (4.4615,1.4820);
\node[inner sep=1pt,anchor=east] at (1.3200,0.5600) {T};
\draw[draw=evMath,line width=.8pt] (2.0911,0.7100) -- (4.7458,0.7100);
\draw[draw=evMath,line width=.7pt] (2.0911,0.6650) -- (2.0911,0.7550);
\draw[draw=evMath,line width=.7pt] (4.7458,0.6650) -- (4.7458,0.7550);
\fill[evMath] (3.5282,0.7100) circle (1.6pt);
\draw[draw=evGPQA,line width=.8pt] (4.2097,0.4100) -- (5.8880,0.4100);
\draw[draw=evGPQA,line width=.7pt] (4.2097,0.3650) -- (4.2097,0.4550);
\draw[draw=evGPQA,line width=.7pt] (5.8880,0.3650) -- (5.8880,0.4550);
\fill[evGPQA] (5.0862,0.3580) rectangle (5.1902,0.4620);
\node[inner sep=1pt,] at (4.2750,-0.6200) {Model-reference agreement (\%)};
\node[inner sep=1pt,] at (4.2750,-0.9400) {(a) Individual phase F1};
\draw[draw=black!12,line width=.35pt] (9.0500,0.0000) -- (9.0500,3.1500);
\node[inner sep=1pt,] at (9.0500,-0.2400) {20};
\draw[draw=black!12,line width=.35pt] (10.4625,0.0000) -- (10.4625,3.1500);
\node[inner sep=1pt,] at (10.4625,-0.2400) {40};
\draw[draw=black!12,line width=.35pt] (11.8750,0.0000) -- (11.8750,3.1500);
\node[inner sep=1pt,] at (11.8750,-0.2400) {60};
\draw[draw=black!12,line width=.35pt] (13.2875,0.0000) -- (13.2875,3.1500);
\node[inner sep=1pt,] at (13.2875,-0.2400) {80};
\draw[draw=black!12,line width=.35pt] (14.7000,0.0000) -- (14.7000,3.1500);
\node[inner sep=1pt,] at (14.7000,-0.2400) {100};
\draw[draw=black!55,line width=.45pt] (9.0500,0.0000) -- (14.7000,0.0000);
\node[inner sep=1pt,anchor=east] at (8.9200,2.4000) {Macro-F1};
\draw[draw=evMath,line width=.8pt] (12.8616,2.5500) -- (13.6605,2.5500);
\draw[draw=evMath,line width=.7pt] (12.8616,2.5050) -- (12.8616,2.5950);
\draw[draw=evMath,line width=.7pt] (13.6605,2.5050) -- (13.6605,2.5950);
\fill[evMath] (13.2902,2.5500) circle (1.6pt);
\draw[draw=evGPQA,line width=.8pt] (13.0969,2.2500) -- (13.7587,2.2500);
\draw[draw=evGPQA,line width=.7pt] (13.0969,2.2050) -- (13.0969,2.2950);
\draw[draw=evGPQA,line width=.7pt] (13.7587,2.2050) -- (13.7587,2.2950);
\fill[evGPQA] (13.4070,2.1980) rectangle (13.5110,2.3020);
\node[inner sep=1pt,anchor=east] at (8.9200,0.9000) {Exact set};
\draw[draw=evMath,line width=.8pt] (11.0275,1.0500) -- (12.3694,1.0500);
\draw[draw=evMath,line width=.7pt] (11.0275,1.0050) -- (11.0275,1.0950);
\draw[draw=evMath,line width=.7pt] (12.3694,1.0050) -- (12.3694,1.0950);
\fill[evMath] (11.7338,1.0500) circle (1.6pt);
\draw[draw=evGPQA,line width=.8pt] (10.0211,0.7500) -- (11.6102,0.7500);
\draw[draw=evGPQA,line width=.7pt] (10.0211,0.7050) -- (10.0211,0.7950);
\draw[draw=evGPQA,line width=.7pt] (11.6102,0.7050) -- (11.6102,0.7950);
\fill[evGPQA] (10.7636,0.6980) rectangle (10.8676,0.8020);
\node[inner sep=1pt,] at (11.8750,-0.6200) {Model-reference agreement (\%)};
\node[inner sep=1pt,] at (11.8750,-0.9400) {(b) Aggregate and complete label set};
\node[inner sep=1pt] at (8.0,3.6200) {{\color{evMath}$\bullet$}\; Mathematics (100)\qquad {\color{evGPQA}\rule{4pt}{4pt}}\; GPQA (80)};
\end{tikzpicture}}
\caption{Frozen classifier transfer under the same three-reviewer model-reference protocol. Each question contributes one chunk; bars show 95\% question-bootstrap intervals conditional on the majority labels. GPQA retains high R agreement but lower E F1 and complete-set agreement than mathematics. Reference-positive R/E/T counts are 67/80/25 for mathematics and 74/41/42 for GPQA. Reference labels are obtained by majority vote from three model reviewers and subsequently reviewed by a human author.}
\label{fig:classifier_transfer}\end{figure*}
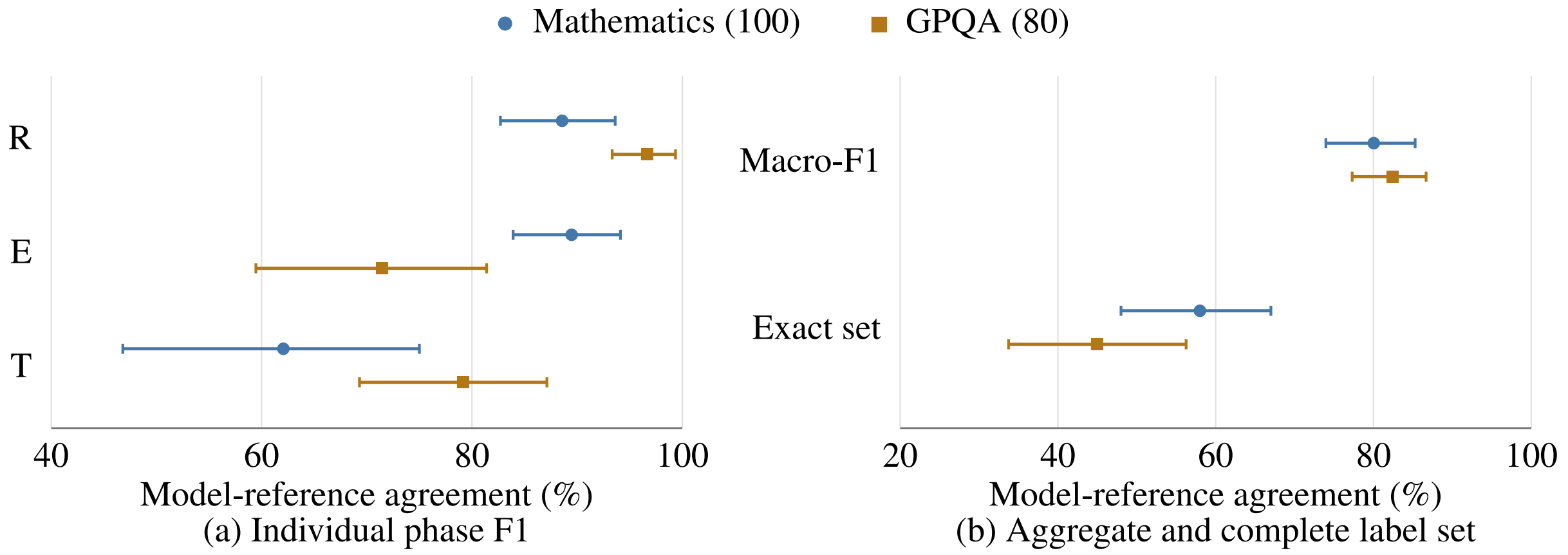

Table~\ref{tab:crossdomain_roles} reports cross-domain results; Figure~\ref{fig:classifier_transfer} shows their uncertainty. With 10,000 question-bootstrap draws stratified by benchmark, macro-F1 is 80.04\% for mathematics (95\% interval 73.97--85.28\%) and 82.43\% for GPQA (77.30--86.67\%). Scoring against individual reviewers gives 78.66--81.29\% and 80.98--81.77\%, respectively. Intervals condition on the majority reference labels.

GPQA has lower E F1 and exact label-set agreement than mathematics despite its higher macro-F1. The cross-domain and 585-chunk cohorts use different sampling and reference protocols. Classifier agreement and GPQA answer accuracy therefore measure distinct outcomes; relating them requires controlled phase-error interventions.

\subsection{Normalization and actor objective}
The maximum observed raw reward change is approximately 0.01; group normalization subsequently determines the training advantages.
At each position, the trainer computes the group mean $\mu_t$ and sample standard deviation $s_t$ over the full reward arrays, then applies the response mask:
\begin{equation}
\widehat A_{it}=m_{it}\frac{q_{it}-\mu_t}{s_t+10^{-6}}.
\label{eq:normalization}
\end{equation}
Padding is included in those group moments through the initialized reward arrays; it is masked after normalization. Changing a reward can change the group moments and hence the advantages of other responses whose raw rewards were unchanged. The actor uses the configured dual-clipped PPO token loss \citep{schulman2017ppo,ye2020dualclip}. With $\rho_{it}$ the current-to-rollout-policy probability ratio, define
\begin{equation}
\begin{aligned}
\ell^0_{it}=\max\bigl\{&-\rho_{it}\widehat A_{it},\\
&-\operatorname{clip}(\rho_{it},0.8,1.2)\widehat A_{it}\bigr\}.
\end{aligned}
\end{equation}
\begin{equation}
\ell_{it}=\begin{cases}
\min\{-3\widehat A_{it},\ell^0_{it}\},&\widehat A_{it}<0,\\
\ell^0_{it},&\widehat A_{it}\ge0.
\end{cases}
\end{equation}
\begin{equation}
\mathcal L=\frac{\sum_{i,t}m_{it}\ell_{it}}{\sum_{i,t}m_{it}}.
\end{equation}
The negative-advantage branch uses a dual-clip constant of 3.0. The resolved configuration has one PPO epoch and disables explicit KL loss and entropy regularization. Figure~\ref{fig:method} shows where PCRS adjusts rewards.

\subsection{Resolved configuration}
Table~\ref{tab:config} collects the training and decoding settings used for the reported checkpoints. Training uses four rollouts per prompt, whereas evaluation samples 16 responses per question; the response-generation cap is 16,384 tokens.
\begin{table}[htbp]
\centering\small
\begin{tabular}{lr}
\toprule
Setting & Value\\\midrule
Training / validation examples & 10,112 / 492\\
Epochs / observed updates & 2 / 52\\
Data seed & 1\\
Train / generation batch & 128 / 128\\
Rollouts per prompt & 4\\
Learning rate / weight decay & $2\times10^{-6}$ / 0.01\\
Warmup ratio / PPO epochs & 0 / 1\\
PPO clip lower / upper & 0.2 / 0.2\\
Negative-advantage dual clip & 3.0\\
PPO mini / per-GPU micro batch & 128 / 1\\
Actor loss aggregation & token mean\\
Explicit KL / entropy coefficient & off / 0\\
Length-reward $\beta$ & 0.1\\
Filter generation-batch cap & 10\\
Curriculum $\kappa$ learning rate & 0.2\\
Evaluation samples / seed & 16 / 0\\
Evaluation batch / tensor parallelism & 4 / 1\\
Evaluation maximum model length & 32,000\\
\bottomrule
\end{tabular}
\caption{Available training and evaluation settings. The response cap is 16,384; the $H$ used in the reward formula is the actual padded response width. Per-GPU micro-batch denotes the configured minimum; token caps govern dynamic batching.}
\label{tab:config}
\end{table}

\subsection{Implementation details}
\label{app:provenance}
The reported checkpoints use four training rollouts per prompt; earlier two-rollout experiments are excluded from this comparison. Group normalization includes initialized padding rewards and uses the sample standard deviation before masking (Eq.~\ref{eq:normalization}). The actor uses a negative-advantage dual-clip constant of 3 \citep{ye2020dualclip}, separate from the raw reward-adjustment bound. Dataset sizes and decoding settings are given \mbox{in Section~\ref{sec:setup} and Table~\ref{tab:config}.}

\subsection{Scope and data availability}
\label{app:gaps}
The benchmark results compare complete configurations; the repetition-only control does not by itself isolate interactions between phase weighting and repetition handling. Selected trace cases illustrate mathematical decisions rather than estimate their prevalence. Cached outputs, annotation records, and implementation configurations are retained privately and are not currently public.
\section{Supplementary Results}
The complete configuration record is retained here: Table~\ref{tab:current_summary} complements the full-suite overview in main-text Figure~\ref{fig:current_results}, and Table~\ref{tab:results} gives per-benchmark values. Weight comparisons and training diagnostics follow.

\label{app:fullresults}

\subsection{Repetition-only component comparison}
\label{app:repeat_control}
The repetition-only run uses the frozen DeepSeek-R1-Distill-Qwen-1.5B initialization, 10,112/492 training/validation split, nominal seed 1, learning rate $2\times10^{-6}$, batch size 128, four rollouts per prompt, and two data epochs. Its final checkpoint is update 52. The same frozen text classifier and native chunk boundaries remain in the reward path, but phase labels no longer change penalty multipliers: repeated content receives $1.1$ and other eligible content receives $1.0$, with the original protection rules retained. This control is distinct from the R-milder weight setting.

Training used four A100-80GB actors and an A100-40GB judge. Results are reported for the final checkpoint of each configuration under the shared evaluation protocol.

All six evaluation caches contain the same question tasks as the reference caches, with 16 responses per question. All configurations use the same scoring rules. The GPQA control yields 1,035 correct responses out of 3,168 and oracle coverage on 179/198 questions. Default PCRS-T has 1,036 correct responses and coverage on 183/198 questions.

\begin{table}[t]\centering\small
\setlength{\tabcolsep}{4pt}
\begin{tabular}{lrrr}\toprule
Benchmark & Pass@1 & Tokens & Pass@16\\\midrule
MATH500 & 85.11 & 4,504.46 & 97.20\\
AIME 2024 & 30.42 & 11,891.07 & 70.00\\
AIME 2025 & 24.58 & 11,628.34 & 46.67\\
AMC23 & 71.09 & 7,355.80 & 95.00\\
OlympiadBench & 54.51 & 8,662.12 & 77.45\\
GPQA-Diamond & 32.67 & 8,150.04 & 90.40\\\bottomrule
\end{tabular}
\caption{Repetition-only final-checkpoint results. Pass@1 and Pass@16 are percentages; tokens are mean complete-response length. All 16 responses, including invalid final answers, remain in the denominator.}
\label{tab:repeat_per_benchmark}\end{table}

The paired analysis retains each question's 16 responses together and uses 10,000 benchmark-stratified question-bootstrap draws (seed 20261007). Repetition-only minus DECS is $-0.30$ Pass@1 points $[-0.82,0.21]$, $+0.07$ Pass@16 points $[-1.16,1.38]$, and $-9.81$ tokens $[-58.95,39.81]$. Default PCRS-T minus repetition-only is reported in Section~\ref{sec:component_ablation}.

\subsection{Full-suite overview}
\paragraph{Full-suite output cost and sampled accuracy.}
On the full 1,379-question suite, PCRS-T reduces mean response length from 7,187.80 to 6,962.68 tokens (3.13\%). The paired interval for the mean token difference is $[-275.06,-174.74]$. Both methods achieve approximately 62.07\% sampled Pass@1, with a paired difference of $-0.005$ percentage points $[-0.535,0.539]$.

\paragraph{Full-suite coverage across 16 attempts.}
Supplementary Pass@16 rises from 86.0769\% to 87.6722\%, a gain of 1.595 points (95\% interval $[0.435,2.828]$). PCRS-T reaches a correct answer on more questions across the same number of attempts, with shorter responses on average. Selecting a successful sample would require a separate answer-selection rule. Appendix~\ref{app:paired_uncertainty} describes the paired analysis.

Figure~\ref{fig:current_results} plots sampled accuracy and oracle coverage against mean response length for all six completed configurations. Upper-left positions indicate fewer generated tokens and a higher score. The full-suite overview complements the main paired comparison in Table~\ref{tab:primary_full}.

\begin{table*}[t]\centering\small
\begin{tabular}{llrrr}\toprule
Weighting & Model & Pass@1 (\%) & Avg. tokens & Pass@16 (\%)\\\midrule
Question-weighted & DECS & 62.07 & 7187.80  & 86.08\\
 & PCRS-H & 62.07 & 7106.23  & 86.80\\
 & PCRS-T & 62.07 & 6962.68  & 87.67\\
 & PCRS-T (E-priority) & 61.62 & 7154.58  & 86.00\\
 & PCRS-T (T-priority) & 62.33 & 7063.42 & 86.66\\
 & PCRS-T (R-milder) & 62.43 & 7048.80 & 86.87\\
\addlinespace[2pt]
Benchmark-equal & DECS & 49.75 & 8685.45  & 77.11\\
 & PCRS-H & 49.97 & 8591.09  & 79.61\\
 & PCRS-T & 49.53 & 8480.26  & 80.82\\
 & PCRS-T (E-priority) & 48.87 & 8664.06  & 77.37\\
 & PCRS-T (T-priority) & 49.90 & 8591.51 & 78.97\\
 & PCRS-T (R-milder) & 50.29 & 8510.45 & 81.01\\
\bottomrule\end{tabular}
\caption{All six completed configurations on 1,379 questions: primary Pass@1 (estimated from 16 samples) and average tokens, with supplementary Pass@16. DECS denotes our quarter-data reproduction; PCRS-T denotes its Default setting; E-priority, T-priority, and R-milder vary the reward weights within PCRS-T. PCRS-H uses exclusive R/E/T labels; PCRS-T uses overlapping labels. Both share penalty-gap reward allocation.}
\label{tab:current_summary}\end{table*}

\begin{table*}[t]\centering\small
\begin{tabular}{llrrrr}\toprule
Benchmark & Model & $N$ & Pass@1 (\%) & Avg. tokens & Pass@16 (\%)\\\midrule
MATH500 & DECS & 500 & 85.10 & 4551.21 & 97.60 \\
 & PCRS-H &  & 85.46 & 4423.54 & 97.40 \\
 & PCRS-T &  & 85.04 & 4365.49 & 97.80 \\
 & PCRS-T (E-priority) &  & 85.06 & 4496.02 & 97.00 \\
\addlinespace[2pt]
AIME 2024 & DECS & 30 & 29.38 & 11660.99 & 63.33 \\
 & PCRS-H &  & 28.54 & 11623.01 & 66.67 \\
 & PCRS-T &  & 30.42 & 11391.56 & 73.33 \\
 & PCRS-T (E-priority) &  & 26.25 & 11804.23 & 63.33 \\
\addlinespace[2pt]
AIME 2025 & DECS & 30 & 22.50 & 11792.42 & 40.00 \\
 & PCRS-H &  & 25.62 & 11567.05 & 50.00 \\
 & PCRS-T &  & 22.08 & 11431.17 & 46.67 \\
 & PCRS-T (E-priority) &  & 23.54 & 11432.37 & 43.33 \\
\addlinespace[2pt]
AMC23 & DECS & 40 & 72.66 & 7377.62 & 92.50 \\
 & PCRS-H &  & 71.72 & 7285.04 & 95.00 \\
 & PCRS-T &  & 71.56 & 7303.26 & 95.00 \\
 & PCRS-T (E-priority) &  & 70.94 & 7516.88 & 95.00 \\
\addlinespace[2pt]
OlympiadBench & DECS & 581 & 54.80 & 8681.54 & 77.28 \\
 & PCRS-H &  & 54.44 & 8627.48 & 79.17 \\
 & PCRS-T &  & 55.36 & 8346.59 & 79.69 \\
 & PCRS-T (E-priority) &  & 54.29 & 8634.16 & 79.17 \\
\addlinespace[2pt]
GPQA-Diamond & DECS & 198 & 34.09 & 8048.94 & 91.92 \\
 & PCRS-H &  & 34.03 & 8020.46 & 89.39 \\
 & PCRS-T &  & 32.70 & 8043.49 & 92.42 \\
 & PCRS-T (E-priority) &  & 33.14 & 8100.70 & 86.36 \\
\bottomrule\end{tabular}
\caption{Per-benchmark results for four final checkpoints after 52 updates. Pass@1 and average complete-response tokens are the primary metrics; Pass@16 is supplementary. Each question has 16 responses. DECS denotes our quarter-data reproduction; PCRS-T denotes its Default setting; E-priority is its execution-priority setting. PCRS-H uses hidden-state proxy labels. Paired PCRS-T--DECS intervals appear in Table~\ref{tab:paired_intervals}.}
\label{tab:results}\end{table*}

\begin{figure*}[t]\centering
\resizebox{\textwidth}{!}{%
\begin{tikzpicture}[x=1cm,y=1cm,font=\rmfamily\fontsize{9}{11}\selectfont]
\definecolor{trBlue}{HTML}{397DB5}
\definecolor{trGold}{HTML}{B78429}
\node[anchor=east] at (1.9,3.7) {T-priority};
\node[anchor=east] at (1.9,2.95) {R-milder};
\node at (3.9000000000000004,4.45) {$\Delta$ Pass@1 (pp)};
\draw[black!15,densely dotted] (2.1,2.50)--(2.1,4.03);
\node[anchor=north] at (2.1,2.36) {-1};
\draw[black!15,densely dotted] (3.7363636363636363,2.50)--(3.7363636363636363,4.03);
\node[anchor=north] at (3.7363636363636363,2.36) {0};
\draw[black!15,densely dotted] (5.372727272727273,2.50)--(5.372727272727273,4.03);
\node[anchor=north] at (5.372727272727273,2.36) {1};
\draw[black!60,dashed,line width=.65pt] (3.7363636363636363,2.50)--(3.7363636363636363,4.03);
\draw[trBlue,line width=1pt] (3.294545454545455,3.7)--(5.012727272727273,3.7);
\draw[trBlue] (3.294545454545455,3.6300000000000003)--(3.294545454545455,3.77) (5.012727272727273,3.6300000000000003)--(5.012727272727273,3.77);
\fill[trBlue] (4.161818181818182,3.7) circle (.062);
\draw[trGold,line width=1pt] (3.409090909090909,2.95)--(5.241818181818181,2.95);
\draw[trGold] (3.409090909090909,2.8800000000000003)--(3.409090909090909,3.02) (5.241818181818181,2.8800000000000003)--(5.241818181818181,3.02);
\fill[trGold] (4.325454545454545,2.95) circle (.062);
\draw[black!50] (2.1,2.50)--(5.7,2.50);
\node at (8.8,4.45) {$\Delta$ mean tokens};
\draw[black!15,densely dotted] (7.163636363636364,2.50)--(7.163636363636364,4.03);
\node[anchor=north] at (7.163636363636364,2.36) {0};
\draw[black!15,densely dotted] (8.8,2.50)--(8.8,4.03);
\node[anchor=north] at (8.8,2.36) {100};
\draw[black!15,densely dotted] (10.436363636363637,2.50)--(10.436363636363637,4.03);
\node[anchor=north] at (10.436363636363637,2.36) {200};
\draw[black!60,dashed,line width=.65pt] (7.163636363636364,2.50)--(7.163636363636364,4.03);
\draw[trBlue,line width=1pt] (7.967909090909091,3.7)--(9.684290909090908,3.7);
\draw[trBlue] (7.967909090909091,3.6300000000000003)--(7.967909090909091,3.77) (9.684290909090908,3.6300000000000003)--(9.684290909090908,3.77);
\fill[trBlue] (8.812109090909091,3.7) circle (.062);
\draw[trGold,line width=1pt] (7.711654545454546,2.95)--(9.432127272727273,2.95);
\draw[trGold] (7.711654545454546,2.8800000000000003)--(7.711654545454546,3.02) (9.432127272727273,2.8800000000000003)--(9.432127272727273,3.02);
\fill[trGold] (8.572709090909092,2.95) circle (.062);
\draw[black!50] (7,2.50)--(10.6,2.50);
\node at (13.700000000000001,4.45) {$\Delta$ Pass@16 (pp)};
\draw[black!15,densely dotted] (12.3,2.50)--(12.3,4.03);
\node[anchor=north] at (12.3,2.36) {-3};
\draw[black!15,densely dotted] (13.1,2.50)--(13.1,4.03);
\node[anchor=north] at (13.1,2.36) {-2};
\draw[black!15,densely dotted] (13.9,2.50)--(13.9,4.03);
\node[anchor=north] at (13.9,2.36) {-1};
\draw[black!15,densely dotted] (14.700000000000001,2.50)--(14.700000000000001,4.03);
\node[anchor=north] at (14.700000000000001,2.36) {0};
\draw[black!15,densely dotted] (15.5,2.50)--(15.5,4.03);
\node[anchor=north] at (15.5,2.36) {1};
\draw[black!60,dashed,line width=.65pt] (14.700000000000001,2.50)--(14.700000000000001,4.03);
\draw[trBlue,line width=1pt] (12.956,3.7)--(14.82,3.7);
\draw[trBlue] (12.956,3.6300000000000003)--(12.956,3.77) (14.82,3.6300000000000003)--(14.82,3.77);
\fill[trBlue] (13.884,3.7) circle (.062);
\draw[trGold,line width=1pt] (13.076,2.95)--(15.052,2.95);
\draw[trGold] (13.076,2.8800000000000003)--(13.076,3.02) (15.052,2.8800000000000003)--(15.052,3.02);
\fill[trGold] (14.06,2.95) circle (.062);
\draw[black!50] (11.9,2.50)--(15.5,2.50);
\end{tikzpicture}}
\caption{T-priority and R-milder minus default PCRS-T: paired mean differences and 95\% percentile intervals, using 10,000 question-bootstrap draws with all 16 responses retained together (seed 20261009). All estimates use the full six-benchmark suite of 1,379 questions. Exact values appear in Table~\ref{tab:weight_uncertainty}.}
\label{fig:tr_differences}\end{figure*}
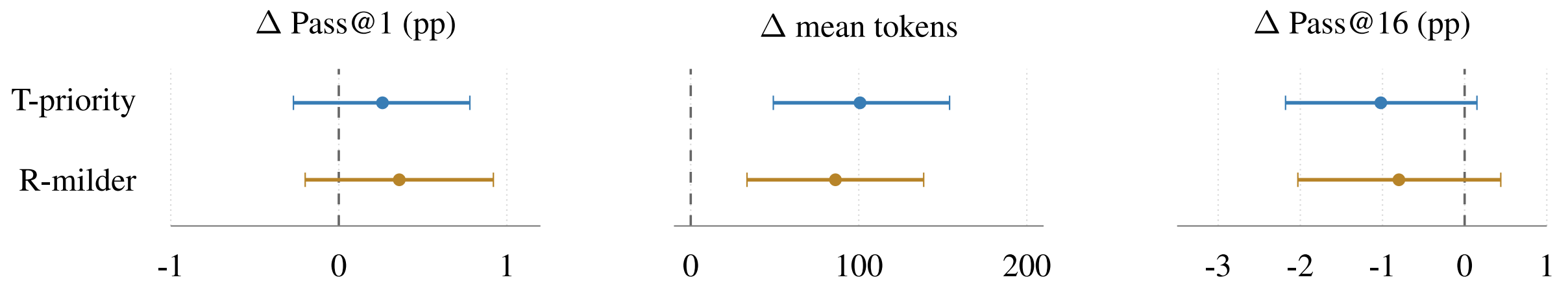
\subsection{Weight Sensitivity for the Text Classifier}
\label{app:weight_results}
\paragraph{Reward-weight settings.}
We evaluate four settings of $(\lambda_R,\lambda_E,\lambda_T)$: default PCRS-T $(0.9,1,1)$, E-priority $(0.9,0.95,1.05)$, T-priority $(0.9,1.05,0.95)$, and R-milder $(0.95,1,1)$. All retain the same classifier, threshold, and exact-repetition factor 1.1, and all complete 52 updates. E-priority starts from the pretrained model rather than continuing the default PCRS-T checkpoint. All four settings use four actor GPUs. T-priority and R-milder are separate training runs. These results characterize sensitivity across the completed configurations.

Figure~\ref{fig:role_weight_map} shows why a phase's nominal factor is not its effect on every token carrying that label. For example, E-priority assigns E a factor of 0.95, but an ET chunk receives $(0.95+1.05)/2=1$. The E and T adjustments also cancel within RET chunks. R-milder moves the weights of R-containing chunks closer to 1. These are consequences of the averaging rule; whether they improve answer quality must be measured separately.

\begin{figure*}[t]\centering
\resizebox{\textwidth}{!}{\begin{tikzpicture}[x=1cm,y=1cm,font=\rmfamily\fontsize{9}{11}\selectfont]
\definecolor{hp0}{HTML}{397DB5}
\definecolor{hp1}{HTML}{B78429}
\definecolor{hp2}{HTML}{AA6187}
\definecolor{hp3}{HTML}{788D3F}
\node[inner sep=1pt,anchor=west,font=\rmfamily\fontsize{9}{11}\selectfont\bfseries] at (0.0000,4.8500) {Native chunk};
\definecolor{cBCC9D6}{HTML}{BCC9D6}
\fill[cBCC9D6] (2.0500,4.6900) rectangle (2.2300,4.9100);
\definecolor{cBCC9D6}{HTML}{BCC9D6}
\fill[cBCC9D6] (2.2800,4.6900) rectangle (2.4600,4.9100);
\definecolor{cBCC9D6}{HTML}{BCC9D6}
\fill[cBCC9D6] (2.5100,4.6900) rectangle (2.6900,4.9100);
\definecolor{cBCC9D6}{HTML}{BCC9D6}
\fill[cBCC9D6] (2.7400,4.6900) rectangle (2.9200,4.9100);
\definecolor{cBCC9D6}{HTML}{BCC9D6}
\fill[cBCC9D6] (2.9700,4.6900) rectangle (3.1500,4.9100);
\definecolor{cBCC9D6}{HTML}{BCC9D6}
\fill[cBCC9D6] (3.2000,4.6900) rectangle (3.3800,4.9100);
\node[inner sep=1pt,] at (3.6500,4.8500) {$\longrightarrow$};
\node[inner sep=1pt,anchor=west] at (4.1500,4.8500) {Frozen R / E / T heads};
\node[inner sep=1pt,] at (8.0500,4.8500) {$\longrightarrow$};
\node[inner sep=1pt,anchor=west] at (8.5500,4.8500) {Overlapping labels};
\node[inner sep=1pt,] at (11.8000,4.8500) {$\longrightarrow$};
\node[inner sep=1pt,anchor=west,font=\rmfamily\fontsize{9}{11}\selectfont\bfseries] at (12.3000,4.8500) {Token weight};
\node[inner sep=1pt,,font=\rmfamily\fontsize{9}{11}\selectfont\bfseries] at (6.5000,4.1400) {Mean of active phase factors};
\node[inner sep=1pt,,font=\rmfamily\fontsize{9}{11}\selectfont\bfseries] at (13.8000,4.1400) {Overrides / default};
\node[inner sep=1pt,] at (3.6200,3.5800) {R};
\node[inner sep=1pt,] at (4.8600,3.5800) {E};
\node[inner sep=1pt,] at (6.1000,3.5800) {T};
\node[inner sep=1pt,] at (7.3400,3.5800) {RE};
\node[inner sep=1pt,] at (8.5800,3.5800) {RT};
\node[inner sep=1pt,] at (9.8200,3.5800) {ET};
\node[inner sep=1pt,] at (11.0600,3.5800) {RET};
\node[inner sep=1pt,,font=\rmfamily\fontsize{8}{10}\selectfont] at (12.3000,3.5800) {Repeat};
\node[inner sep=1pt,,font=\rmfamily\fontsize{8}{10}\selectfont] at (13.5400,3.5800) {None};
\node[inner sep=1pt,,font=\rmfamily\fontsize{8}{10}\selectfont] at (14.7800,3.5800) {Protected};
\definecolor{c397DB5}{HTML}{397DB5}
\fill[c397DB5] (0.0400,2.8400) rectangle (0.1400,2.9400);
\node[inner sep=1pt,anchor=west] at (0.3000,2.8900) {Default};
\definecolor{c9CBEDA}{HTML}{9CBEDA}
\fill[c9CBEDA] (3.0500,2.6600) rectangle (4.1900,3.1400);
\node[inner sep=1pt,] at (3.6200,2.9000) {0.90};
\definecolor{cF1F1F1}{HTML}{F1F1F1}
\fill[cF1F1F1] (4.2900,2.6600) rectangle (5.4300,3.1400);
\node[inner sep=1pt,] at (4.8600,2.9000) {1.00};
\definecolor{cF1F1F1}{HTML}{F1F1F1}
\fill[cF1F1F1] (5.5300,2.6600) rectangle (6.6700,3.1400);
\node[inner sep=1pt,] at (6.1000,2.9000) {1.00};
\definecolor{cCDDEEC}{HTML}{CDDEEC}
\fill[cCDDEEC] (6.7700,2.6600) rectangle (7.9100,3.1400);
\node[inner sep=1pt,] at (7.3400,2.9000) {0.95};
\definecolor{cCDDEEC}{HTML}{CDDEEC}
\fill[cCDDEEC] (8.0100,2.6600) rectangle (9.1500,3.1400);
\node[inner sep=1pt,] at (8.5800,2.9000) {0.95};
\definecolor{cF1F1F1}{HTML}{F1F1F1}
\fill[cF1F1F1] (9.2500,2.6600) rectangle (10.3900,3.1400);
\node[inner sep=1pt,] at (9.8200,2.9000) {1.00};
\definecolor{cDEE9F3}{HTML}{DEE9F3}
\fill[cDEE9F3] (10.4900,2.6600) rectangle (11.6300,3.1400);
\node[inner sep=1pt,] at (11.0600,2.9000) {0.967};
\definecolor{cDBC294}{HTML}{DBC294}
\fill[cDBC294] (11.7300,2.6600) rectangle (12.8700,3.1400);
\node[inner sep=1pt,] at (12.3000,2.9000) {1.10};
\definecolor{cF1F1F1}{HTML}{F1F1F1}
\fill[cF1F1F1] (12.9700,2.6600) rectangle (14.1100,3.1400);
\node[inner sep=1pt,] at (13.5400,2.9000) {1.00};
\definecolor{cF1F1F1}{HTML}{F1F1F1}
\fill[cF1F1F1] (14.2100,2.6600) rectangle (15.3500,3.1400);
\node[inner sep=1pt,] at (14.7800,2.9000) {1.00};
\definecolor{cB78429}{HTML}{B78429}
\fill[cB78429] (0.0400,2.1900) rectangle (0.1400,2.2900);
\node[inner sep=1pt,anchor=west] at (0.3000,2.2400) {E-priority};
\definecolor{c9CBEDA}{HTML}{9CBEDA}
\fill[c9CBEDA] (3.0500,2.0100) rectangle (4.1900,2.4900);
\node[inner sep=1pt,] at (3.6200,2.2500) {0.90};
\definecolor{cCDDEEC}{HTML}{CDDEEC}
\fill[cCDDEEC] (4.2900,2.0100) rectangle (5.4300,2.4900);
\node[inner sep=1pt,] at (4.8600,2.2500) {0.95};
\definecolor{cEDE0C9}{HTML}{EDE0C9}
\fill[cEDE0C9] (5.5300,2.0100) rectangle (6.6700,2.4900);
\node[inner sep=1pt,] at (6.1000,2.2500) {1.05};
\definecolor{cB5CEE3}{HTML}{B5CEE3}
\fill[cB5CEE3] (6.7700,2.0100) rectangle (7.9100,2.4900);
\node[inner sep=1pt,] at (7.3400,2.2500) {0.925};
\definecolor{cE6EFF6}{HTML}{E6EFF6}
\fill[cE6EFF6] (8.0100,2.0100) rectangle (9.1500,2.4900);
\node[inner sep=1pt,] at (8.5800,2.2500) {0.975};
\definecolor{cF1F1F1}{HTML}{F1F1F1}
\fill[cF1F1F1] (9.2500,2.0100) rectangle (10.3900,2.4900);
\node[inner sep=1pt,] at (9.8200,2.2500) {1.00};
\definecolor{cDEE9F3}{HTML}{DEE9F3}
\fill[cDEE9F3] (10.4900,2.0100) rectangle (11.6300,2.4900);
\node[inner sep=1pt,] at (11.0600,2.2500) {0.967};
\definecolor{cDBC294}{HTML}{DBC294}
\fill[cDBC294] (11.7300,2.0100) rectangle (12.8700,2.4900);
\node[inner sep=1pt,] at (12.3000,2.2500) {1.10};
\definecolor{cF1F1F1}{HTML}{F1F1F1}
\fill[cF1F1F1] (12.9700,2.0100) rectangle (14.1100,2.4900);
\node[inner sep=1pt,] at (13.5400,2.2500) {1.00};
\definecolor{cF1F1F1}{HTML}{F1F1F1}
\fill[cF1F1F1] (14.2100,2.0100) rectangle (15.3500,2.4900);
\node[inner sep=1pt,] at (14.7800,2.2500) {1.00};
\definecolor{cAA6187}{HTML}{AA6187}
\fill[cAA6187] (0.0400,1.5400) rectangle (0.1400,1.6400);
\node[inner sep=1pt,anchor=west] at (0.3000,1.5900) {T-priority};
\definecolor{c9CBEDA}{HTML}{9CBEDA}
\fill[c9CBEDA] (3.0500,1.3600) rectangle (4.1900,1.8400);
\node[inner sep=1pt,] at (3.6200,1.6000) {0.90};
\definecolor{cEDE0C9}{HTML}{EDE0C9}
\fill[cEDE0C9] (4.2900,1.3600) rectangle (5.4300,1.8400);
\node[inner sep=1pt,] at (4.8600,1.6000) {1.05};
\definecolor{cCDDEEC}{HTML}{CDDEEC}
\fill[cCDDEEC] (5.5300,1.3600) rectangle (6.6700,1.8400);
\node[inner sep=1pt,] at (6.1000,1.6000) {0.95};
\definecolor{cE6EFF6}{HTML}{E6EFF6}
\fill[cE6EFF6] (6.7700,1.3600) rectangle (7.9100,1.8400);
\node[inner sep=1pt,] at (7.3400,1.6000) {0.975};
\definecolor{cB5CEE3}{HTML}{B5CEE3}
\fill[cB5CEE3] (8.0100,1.3600) rectangle (9.1500,1.8400);
\node[inner sep=1pt,] at (8.5800,1.6000) {0.925};
\definecolor{cF1F1F1}{HTML}{F1F1F1}
\fill[cF1F1F1] (9.2500,1.3600) rectangle (10.3900,1.8400);
\node[inner sep=1pt,] at (9.8200,1.6000) {1.00};
\definecolor{cDEE9F3}{HTML}{DEE9F3}
\fill[cDEE9F3] (10.4900,1.3600) rectangle (11.6300,1.8400);
\node[inner sep=1pt,] at (11.0600,1.6000) {0.967};
\definecolor{cDBC294}{HTML}{DBC294}
\fill[cDBC294] (11.7300,1.3600) rectangle (12.8700,1.8400);
\node[inner sep=1pt,] at (12.3000,1.6000) {1.10};
\definecolor{cF1F1F1}{HTML}{F1F1F1}
\fill[cF1F1F1] (12.9700,1.3600) rectangle (14.1100,1.8400);
\node[inner sep=1pt,] at (13.5400,1.6000) {1.00};
\definecolor{cF1F1F1}{HTML}{F1F1F1}
\fill[cF1F1F1] (14.2100,1.3600) rectangle (15.3500,1.8400);
\node[inner sep=1pt,] at (14.7800,1.6000) {1.00};
\definecolor{c788D3F}{HTML}{788D3F}
\fill[c788D3F] (0.0400,0.8900) rectangle (0.1400,0.9900);
\node[inner sep=1pt,anchor=west] at (0.3000,0.9400) {R-milder};
\definecolor{cCDDEEC}{HTML}{CDDEEC}
\fill[cCDDEEC] (3.0500,0.7100) rectangle (4.1900,1.1900);
\node[inner sep=1pt,] at (3.6200,0.9500) {0.95};
\definecolor{cF1F1F1}{HTML}{F1F1F1}
\fill[cF1F1F1] (4.2900,0.7100) rectangle (5.4300,1.1900);
\node[inner sep=1pt,] at (4.8600,0.9500) {1.00};
\definecolor{cF1F1F1}{HTML}{F1F1F1}
\fill[cF1F1F1] (5.5300,0.7100) rectangle (6.6700,1.1900);
\node[inner sep=1pt,] at (6.1000,0.9500) {1.00};
\definecolor{cE6EFF6}{HTML}{E6EFF6}
\fill[cE6EFF6] (6.7700,0.7100) rectangle (7.9100,1.1900);
\node[inner sep=1pt,] at (7.3400,0.9500) {0.975};
\definecolor{cE6EFF6}{HTML}{E6EFF6}
\fill[cE6EFF6] (8.0100,0.7100) rectangle (9.1500,1.1900);
\node[inner sep=1pt,] at (8.5800,0.9500) {0.975};
\definecolor{cF1F1F1}{HTML}{F1F1F1}
\fill[cF1F1F1] (9.2500,0.7100) rectangle (10.3900,1.1900);
\node[inner sep=1pt,] at (9.8200,0.9500) {1.00};
\definecolor{cEFF4F9}{HTML}{EFF4F9}
\fill[cEFF4F9] (10.4900,0.7100) rectangle (11.6300,1.1900);
\node[inner sep=1pt,] at (11.0600,0.9500) {0.983};
\definecolor{cDBC294}{HTML}{DBC294}
\fill[cDBC294] (11.7300,0.7100) rectangle (12.8700,1.1900);
\node[inner sep=1pt,] at (12.3000,0.9500) {1.10};
\definecolor{cF1F1F1}{HTML}{F1F1F1}
\fill[cF1F1F1] (12.9700,0.7100) rectangle (14.1100,1.1900);
\node[inner sep=1pt,] at (13.5400,0.9500) {1.00};
\definecolor{cF1F1F1}{HTML}{F1F1F1}
\fill[cF1F1F1] (14.2100,0.7100) rectangle (15.3500,1.1900);
\node[inner sep=1pt,] at (14.7800,0.9500) {1.00};
\definecolor{cBBBBBB}{HTML}{BBBBBB}
\draw[draw=cBBBBBB,line width=.4pt] (11.6700,0.6400)--(11.6700,3.3000);
\definecolor{c9CBEDA}{HTML}{9CBEDA}
\fill[c9CBEDA] (0.1500,0.1200) rectangle (0.3500,0.3200);
\node[inner sep=1pt,anchor=west] at (0.5000,0.2200) {Lighter penalty};
\definecolor{cF1F1F1}{HTML}{F1F1F1}
\fill[cF1F1F1] (4.0000,0.1200) rectangle (4.2000,0.3200);
\node[inner sep=1pt,anchor=west] at (4.3500,0.2200) {Unchanged};
\definecolor{cDBC294}{HTML}{DBC294}
\fill[cDBC294] (7.3500,0.1200) rectangle (7.5500,0.3200);
\node[inner sep=1pt,anchor=west] at (7.7000,0.2200) {Stronger penalty};
\node[inner sep=1pt,anchor=west] at (12.1500,0.2200) {DECS gap = 1};
\end{tikzpicture}}
\caption{From chunk labels to token weights within PCRS-T. Each of three classifier heads can be active, yielding seven nonempty phase sets. Cells give the configured multiplier of the DECS penalty gap, averaged over active phases and broadcast to the original tokens. Repetition overrides phase weights; protection overrides repetition. Unlabeled and protected chunks use 1. Colors show penalty multipliers: blue indicates a smaller penalty and gold a larger one. Only eligible positions receive a raw reward change. RET factors use \mbox{three decimal places.}}
\label{fig:role_weight_map}
\end{figure*}

\begin{figure*}[t]\centering
\resizebox{\textwidth}{!}{\begin{tikzpicture}[x=1cm,y=1cm,font=\rmfamily\fontsize{9}{11}\selectfont]
\definecolor{hp0}{HTML}{397DB5}
\definecolor{hp1}{HTML}{B78429}
\definecolor{hp2}{HTML}{AA6187}
\definecolor{hp3}{HTML}{788D3F}
\node[anchor=center,inner sep=0pt] at (8.1,8.00) {%
{\color{hp0}\rule{1.5mm}{1.5mm}}\hspace{0.35em}Default\hspace{0.8em}%
{\color{hp1}\rule{1.5mm}{1.5mm}}\hspace{0.35em}E-priority\hspace{0.8em}%
{\color{hp2}\rule{1.5mm}{1.5mm}}\hspace{0.35em}T-priority\hspace{0.8em}%
{\color{hp3}\rule{1.5mm}{1.5mm}}\hspace{0.35em}R-milder};
\node[inner sep=1pt,anchor=west,font=\rmfamily\fontsize{9}{11}\selectfont\bfseries] at (0.0000,7.4400) {PCRS-T settings: all 1,379 questions};
\definecolor{cDDDDDD}{HTML}{DDDDDD}
\draw[draw=cDDDDDD,line width=.4pt] (0.8000,4.6200)--(4.8000,4.6200);
\node[inner sep=1pt,anchor=east,font=\rmfamily\fontsize{8}{10}\selectfont] at (0.6800,4.6200) {61};
\definecolor{cDDDDDD}{HTML}{DDDDDD}
\draw[draw=cDDDDDD,line width=.4pt] (0.8000,5.6350)--(4.8000,5.6350);
\node[inner sep=1pt,anchor=east,font=\rmfamily\fontsize{8}{10}\selectfont] at (0.6800,5.6350) {62};
\definecolor{cDDDDDD}{HTML}{DDDDDD}
\draw[draw=cDDDDDD,line width=.4pt] (0.8000,6.6500)--(4.8000,6.6500);
\node[inner sep=1pt,anchor=east,font=\rmfamily\fontsize{8}{10}\selectfont] at (0.6800,6.6500) {63};
\definecolor{c999999}{HTML}{999999}
\draw[draw=c999999,line width=.4pt] (0.8000,4.6200)--(4.8000,4.6200);
\definecolor{c999999}{HTML}{999999}
\draw[draw=c999999,line width=.4pt] (0.8000,4.6200)--(0.8000,6.6500);
\definecolor{c397DB5}{HTML}{397DB5}
\fill[c397DB5] (1.0200,4.6200) rectangle (1.5800,5.7055);
\node[inner sep=1pt,,font=\rmfamily\fontsize{8}{10}\selectfont] at (1.3000,5.8855) {62.07};
\node[inner sep=1pt,anchor=east,rotate=90] at (1.3000,4.40) {Default};
\definecolor{cB78429}{HTML}{B78429}
\fill[cB78429] (2.0200,4.6200) rectangle (2.5800,5.2455);
\node[inner sep=1pt,,font=\rmfamily\fontsize{8}{10}\selectfont] at (2.3000,5.4255) {61.62};
\node[inner sep=1pt,anchor=east,rotate=90] at (2.3000,4.40) {E-priority};
\definecolor{cAA6187}{HTML}{AA6187}
\fill[cAA6187] (3.0200,4.6200) rectangle (3.5800,5.9677);
\node[inner sep=1pt,,font=\rmfamily\fontsize{8}{10}\selectfont] at (3.3000,6.1477) {62.33};
\node[inner sep=1pt,anchor=east,rotate=90] at (3.3000,4.40) {T-priority};
\definecolor{c788D3F}{HTML}{788D3F}
\fill[c788D3F] (4.0200,4.6200) rectangle (4.5800,6.0689);
\node[inner sep=1pt,,font=\rmfamily\fontsize{8}{10}\selectfont] at (4.3000,6.2489) {62.43};
\node[inner sep=1pt,anchor=east,rotate=90] at (4.3000,4.40) {R-milder};
\node[inner sep=1pt,] at (2.8000,7.0500) {Pass@1 (\%)};
\definecolor{cDDDDDD}{HTML}{DDDDDD}
\draw[draw=cDDDDDD,line width=.4pt] (11.4000,4.6200)--(15.4000,4.6200);
\node[inner sep=1pt,anchor=east,font=\rmfamily\fontsize{8}{10}\selectfont] at (11.2800,4.6200) {6,800};
\definecolor{cDDDDDD}{HTML}{DDDDDD}
\draw[draw=cDDDDDD,line width=.4pt] (11.4000,5.4320)--(15.4000,5.4320);
\node[inner sep=1pt,anchor=east,font=\rmfamily\fontsize{8}{10}\selectfont] at (11.2800,5.4320) {7,000};
\definecolor{cDDDDDD}{HTML}{DDDDDD}
\draw[draw=cDDDDDD,line width=.4pt] (11.4000,6.2440)--(15.4000,6.2440);
\node[inner sep=1pt,anchor=east,font=\rmfamily\fontsize{8}{10}\selectfont] at (11.2800,6.2440) {7,200};
\definecolor{c999999}{HTML}{999999}
\draw[draw=c999999,line width=.4pt] (11.4000,4.6200)--(15.4000,4.6200);
\definecolor{c999999}{HTML}{999999}
\draw[draw=c999999,line width=.4pt] (11.4000,4.6200)--(11.4000,6.6500);
\definecolor{c397DB5}{HTML}{397DB5}
\fill[c397DB5] (11.6200,4.6200) rectangle (12.1800,5.2805);
\node[inner sep=1pt,,font=\rmfamily\fontsize{8}{10}\selectfont] at (11.9000,5.4605) {6,963};
\node[inner sep=1pt,anchor=east,rotate=90] at (11.9000,4.40) {Default};
\definecolor{cB78429}{HTML}{B78429}
\fill[cB78429] (12.6200,4.6200) rectangle (13.1800,6.0596);
\node[inner sep=1pt,,font=\rmfamily\fontsize{8}{10}\selectfont] at (12.9000,6.2396) {7,155};
\node[inner sep=1pt,anchor=east,rotate=90] at (12.9000,4.40) {E-priority};
\definecolor{cAA6187}{HTML}{AA6187}
\fill[cAA6187] (13.6200,4.6200) rectangle (14.1800,5.6895);
\node[inner sep=1pt,,font=\rmfamily\fontsize{8}{10}\selectfont] at (13.9000,5.8695) {7,063};
\node[inner sep=1pt,anchor=east,rotate=90] at (13.9000,4.40) {T-priority};
\definecolor{c788D3F}{HTML}{788D3F}
\fill[c788D3F] (14.6200,4.6200) rectangle (15.1800,5.6301);
\node[inner sep=1pt,,font=\rmfamily\fontsize{8}{10}\selectfont] at (14.9000,5.8101) {7,049};
\node[inner sep=1pt,anchor=east,rotate=90] at (14.9000,4.40) {R-milder};
\node[inner sep=1pt,] at (13.4000,7.0500) {Mean tokens};
\definecolor{cDDDDDD}{HTML}{DDDDDD}
\draw[draw=cDDDDDD,line width=.4pt] (6.1000,4.6200)--(10.1000,4.6200);
\node[inner sep=1pt,anchor=east,font=\rmfamily\fontsize{8}{10}\selectfont] at (5.9800,4.6200) {85};
\definecolor{cDDDDDD}{HTML}{DDDDDD}
\draw[draw=cDDDDDD,line width=.4pt] (6.1000,5.6350)--(10.1000,5.6350);
\node[inner sep=1pt,anchor=east,font=\rmfamily\fontsize{8}{10}\selectfont] at (5.9800,5.6350) {87};
\definecolor{cDDDDDD}{HTML}{DDDDDD}
\draw[draw=cDDDDDD,line width=.4pt] (6.1000,6.6500)--(10.1000,6.6500);
\node[inner sep=1pt,anchor=east,font=\rmfamily\fontsize{8}{10}\selectfont] at (5.9800,6.6500) {89};
\definecolor{c999999}{HTML}{999999}
\draw[draw=c999999,line width=.4pt] (6.1000,4.6200)--(10.1000,4.6200);
\definecolor{c999999}{HTML}{999999}
\draw[draw=c999999,line width=.4pt] (6.1000,4.6200)--(6.1000,6.6500);
\definecolor{c397DB5}{HTML}{397DB5}
\fill[c397DB5] (6.3200,4.6200) rectangle (6.8800,5.9762);
\node[inner sep=1pt,,font=\rmfamily\fontsize{8}{10}\selectfont] at (6.6000,6.1562) {87.67};
\node[inner sep=1pt,anchor=east,rotate=90] at (6.6000,4.40) {Default};
\definecolor{cB78429}{HTML}{B78429}
\fill[cB78429] (7.3200,4.6200) rectangle (7.8800,5.1297);
\node[inner sep=1pt,,font=\rmfamily\fontsize{8}{10}\selectfont] at (7.6000,5.3097) {86.00};
\node[inner sep=1pt,anchor=east,rotate=90] at (7.6000,4.40) {E-priority};
\definecolor{cAA6187}{HTML}{AA6187}
\fill[cAA6187] (8.3200,4.6200) rectangle (8.8800,5.4609);
\node[inner sep=1pt,,font=\rmfamily\fontsize{8}{10}\selectfont] at (8.6000,5.6409) {86.66};
\node[inner sep=1pt,anchor=east,rotate=90] at (8.6000,4.40) {T-priority};
\definecolor{c788D3F}{HTML}{788D3F}
\fill[c788D3F] (9.3200,4.6200) rectangle (9.8800,5.5713);
\node[inner sep=1pt,,font=\rmfamily\fontsize{8}{10}\selectfont] at (9.6000,5.7513) {86.87};
\node[inner sep=1pt,anchor=east,rotate=90] at (9.6000,4.40) {R-milder};
\node[inner sep=1pt,] at (8.1000,7.0500) {Pass@16 (\%)};
\end{tikzpicture}}
\caption{Four reward-weight settings within PCRS-T on identical evaluation questions, with 16 responses per question. Bars show absolute question-weighted values on focused, nonzero vertical scales. Default has the lowest token use and highest full-suite coverage; R-milder has the highest Pass@1 point estimate. Columns show Pass@1, Pass@16, and mean tokens. Table~\ref{tab:weight_uncertainty} reports paired intervals.}
\label{fig:weight_results}
\end{figure*}
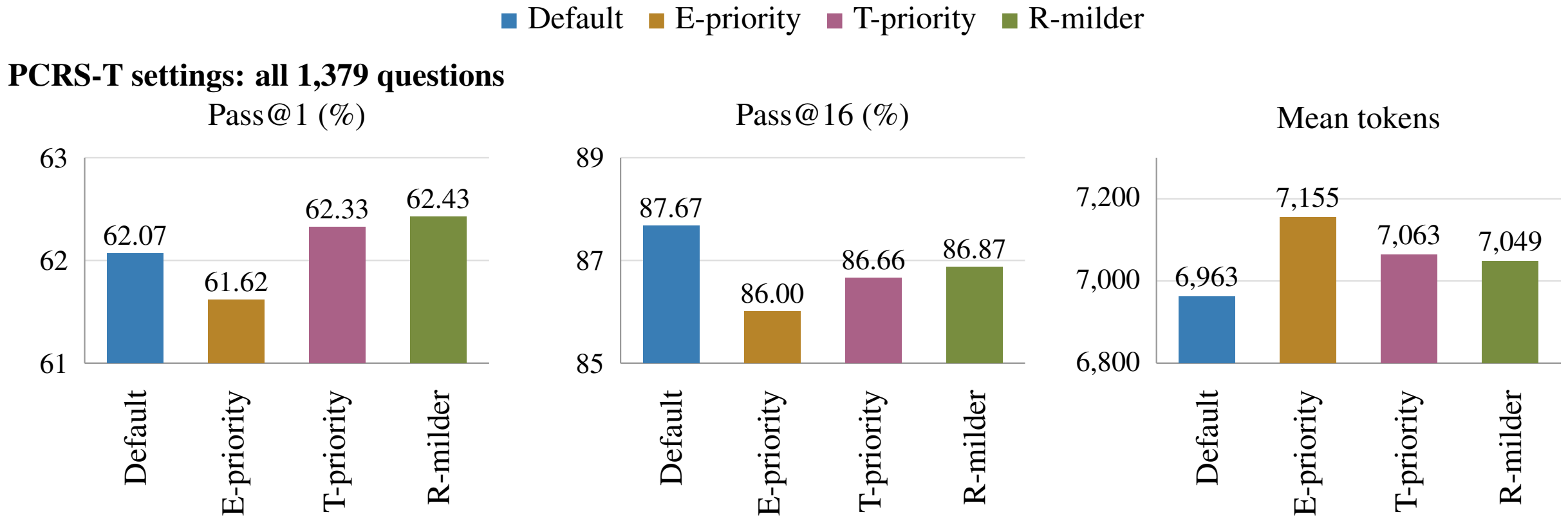

\paragraph{Observed trade-offs.}
\label{sec:weights}
Figure~\ref{fig:weight_results} compares all four completed PCRS-T weight settings. Default PCRS-T uses the fewest tokens (6,962.68) and has the highest question-weighted Pass@16 (87.67\%). E-priority uses 7,154.58 tokens, with lower Pass@1 (61.62\%) and Pass@16 (86.00\%). Relative to default PCRS-T, E-priority therefore improves none of these three outcomes under full-suite question weighting.

Under full-suite question weighting, the two new settings have higher Pass@1 point estimates but use more tokens and have lower Pass@16. T-priority and R-milder reach Pass@1 of 62.33\% and 62.43\%, compared with 62.07\% for default PCRS-T, but use 100.74 and 86.11 more tokens per response. Their Pass@16 values are 86.66\% and 86.87\%. The paired question-bootstrap intervals for the token increases are $[49.15,154.04]$ and $[33.49,138.63]$; Table~\ref{tab:weight_uncertainty} reports intervals for all three metrics.

On GPQA, T-priority reaches 35.51\% Pass@1, versus 32.70\% for default PCRS-T, with lower mean length (7,874.82 versus 8,043.49 tokens); both cover 183/198 questions. Appendix~\ref{app:weight_results} reports all six benchmarks and \mbox{both full-suite aggregation schemes.}

E-priority and T-priority each change two phase factors. R-milder changes the R factor from 0.9 to 0.95. Among the evaluated configurations, default PCRS-T has the lowest full-suite token use and highest coverage, while R-milder has the highest Pass@1 point estimate.

\paragraph{Complete evaluations and paired contrasts.}
\begin{table*}[t]\centering\small
\begin{lrbox}{\WeightTableBox}\begin{tabular}{lrrrrrr}\toprule
& \multicolumn{3}{c}{T-priority} & \multicolumn{3}{c}{R-milder}\\
\cmidrule(lr){2-4}\cmidrule(lr){5-7}
Benchmark / aggregate & Pass@1 & Tokens & Pass@16 & Pass@1 & Tokens & Pass@16\\\midrule
MATH500 & 85.82 & 4,423.27 & 97.40 & 85.70 & 4,400.78 & 98.00\\
AIME 2024 & 30.21 & 11,765.69 & 70.00 & 31.46 & 11,487.52 & 76.67\\
AIME 2025 & 22.08 & 11,581.08 & 43.33 & 25.21 & 11,365.92 & 50.00\\
AMC23 & 71.41 & 7,340.33 & 92.50 & 70.00 & 7,231.04 & 92.50\\
OlympiadBench & 54.36 & 8,563.84 & 78.14 & 54.95 & 8,518.85 & 77.97\\
GPQA-Diamond & 35.51 & 7,874.82 & 92.42 & 34.41 & 8,058.60 & 90.91\\
\midrule
Question-weighted & 62.33 & 7,063.42 & 86.66 & 62.43 & 7,048.80 & 86.87\\
Benchmark-equal & 49.90 & 8,591.51 & 78.97 & 50.29 & 8,510.45 & 81.01\\
\bottomrule\end{tabular}\end{lrbox}
\ifdim\wd\WeightTableBox>\textwidth\resizebox{\textwidth}{!}{\usebox{\WeightTableBox}}\else\usebox{\WeightTableBox}\fi
\caption{Results for the PCRS-T T-priority and R-milder settings. Accuracy and coverage are percentages; tokens count the complete response. All six benchmark inputs match those in Table~\ref{tab:results}. Question-weighted and benchmark-equal rows use all 1,379 questions.}
\label{tab:tr_results}\end{table*}
\begin{table*}[t]\centering\small
\begin{lrbox}{\WeightTableBox}\begin{tabular}{llrrr}\toprule
Questions & Contrast & $\Delta$ Pass@1 (pp) & $\Delta$ tokens & $\Delta$ Pass@16 (pp)\\\midrule
All 1,379 & T-priority minus Default & $+0.26$ [$ -0.27, 0.78 $] & $+100.74$ [$ 49.15, 154.04 $] & $-1.02$ [$ -2.18, 0.15 $]\\
All 1,379 & R-milder minus Default & $+0.36$ [$ -0.20, 0.92 $] & $+86.11$ [$ 33.49, 138.63 $] & $-0.80$ [$ -2.03, 0.44 $]\\
\bottomrule\end{tabular}\end{lrbox}
\ifdim\wd\WeightTableBox>\textwidth\resizebox{\textwidth}{!}{\usebox{\WeightTableBox}}\else\usebox{\WeightTableBox}\fi
\caption{Paired differences from default PCRS-T and 95\% percentile intervals. We resample questions with all 16 responses kept together, using 10,000 draws and seed 20261009.}
\label{tab:weight_uncertainty}\end{table*}
T-priority and R-milder are evaluated on the same questions and prompts as the other configurations.

\subsection{Uncertainty and Length Tails}
\label{app:paired_uncertainty}
The paired analysis uses 88,256 responses across 24 model--benchmark cells, with identical questions and prompts across models. Each of 10,000 bootstrap draws (seed 20261007) resamples questions within each benchmark, retaining all 16 responses and the same selected questions across models. Overall estimates retain either original benchmark sample weights or equal benchmark weights. Intervals are pointwise percentiles without multiplicity adjustment, conditional on the fixed checkpoints and generations; they do not measure \mbox{training-seed or fresh-generation variation.}

\begin{table*}[t]
\centering\small\setlength{\tabcolsep}{5pt}
\begin{tabular}{lrrr}\toprule
Scope & $\Delta$ Pass@1 (points) & $\Delta$ Pass@16 (points) & $\Delta$ mean tokens\\\midrule
MATH500 & $-0.06\;[-0.74,0.61]$ & $0.20\;[-0.80,1.20]$ & $-185.72\;[-258.22,-114.49]$\\
AIME'24 & $1.04\;[-3.12,5.00]$ & $10.00\;[-6.67,26.67]$ & $-269.42\;[-641.95,103.49]$\\
AIME'25 & $-0.42\;[-3.33,2.08]$ & $6.67\;[0.00,16.67]$ & $-361.25\;[-742.07,21.13]$\\
AMC23 & $-1.09\;[-4.69,2.34]$ & $2.50\;[0.00,7.50]$ & $-74.36\;[-349.91,209.97]$\\
OlympiadBench & $0.56\;[-0.28,1.40]$ & $2.41\;[0.52,4.48]$ & $-334.95\;[-418.31,-250.88]$\\
GPQA & $-1.39\;[-3.25,0.54]$ & $0.51\;[-4.04,5.05]$ & $-5.45\;[-136.77,124.05]$\\
Question-weighted & $-0.00\;[-0.53,0.54]$ & $1.60\;[0.44,2.83]$ & $-225.12\;[-275.06,-174.74]$\\
Benchmark-equal & $-0.23\;[-1.32,0.84]$ & $3.71\;[0.42,7.13]$ & $-205.19\;[-312.97,-101.69]$\\
\bottomrule\end{tabular}
\caption{Paired PCRS-T--DECS differences and 95\% question-bootstrap intervals. A negative token difference \mbox{denotes lower output volume.}}
\label{tab:paired_intervals}
\end{table*}

Question-weighted PCRS-T--DECS Pass@16 increases by 1.60 points with interval $[0.44,2.83]$, while the Pass@1 interval spans gains and losses. The mean output-volume difference is reported together with the sampled-accuracy interval. The question-weighted median response length is 5,304 tokens for DECS and 5,017.5 for PCRS-T; the proportions at or above 16,384 are 14.20\% and 12.39\%. All four models have P90/P95/P99 of 16,384, reflecting accumulation near the configured cap rather than identical tail distributions. Across all four caches, 87 responses exceed 16,384 (maximum 16,450); all recorded counts are retained. Reaching or exceeding the cap is a diagnostic, not \mbox{a verified termination reason.}

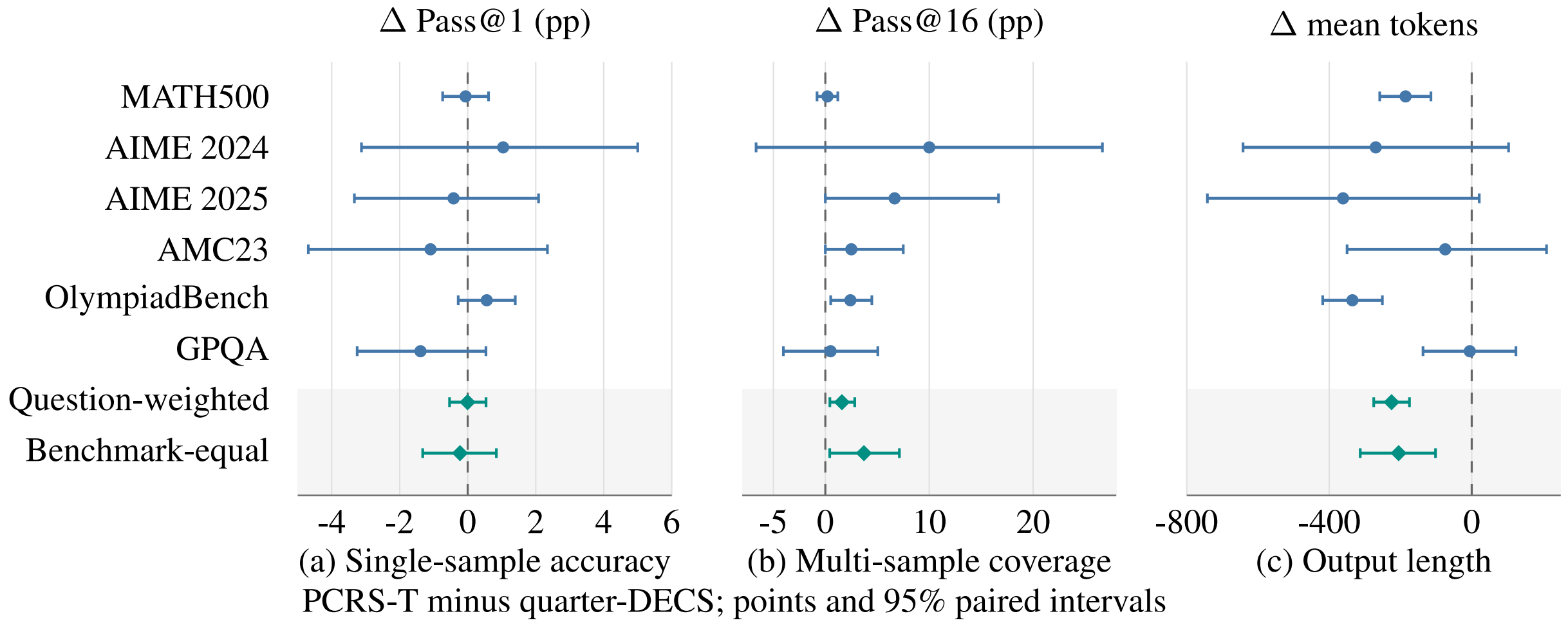
\begin{figure*}[t]\centering
\resizebox{\textwidth}{!}{\begin{tikzpicture}[x=1cm,y=1cm,font=\fontsize{9}{11}\selectfont]
\definecolor{evB}{HTML}{777777}
\definecolor{evT}{HTML}{4477AA}
\definecolor{evS}{HTML}{008F82}
\definecolor{evE}{HTML}{CC6677}
\definecolor{evMath}{HTML}{4477AA}
\definecolor{evGPQA}{HTML}{B47716}
\node[inner sep=1pt,anchor=east] at (2.7800,3.6800) {MATH500};
\node[inner sep=1pt,anchor=east] at (2.7800,3.2100) {AIME 2024};
\node[inner sep=1pt,anchor=east] at (2.7800,2.7400) {AIME 2025};
\node[inner sep=1pt,anchor=east] at (2.7800,2.2700) {AMC23};
\node[inner sep=1pt,anchor=east] at (2.7800,1.8000) {OlympiadBench};
\node[inner sep=1pt,anchor=east] at (2.7800,1.3300) {GPQA};
\node[inner sep=1pt,anchor=east] at (2.7800,0.8600) {Question-weighted};
\node[inner sep=1pt,anchor=east] at (2.7800,0.3900) {Benchmark-equal};
\fill[black!4] (3.000,0) rectangle (6.450,.98);
\draw[draw=black!12,line width=.35pt] (3.3136,0.0000) -- (3.3136,4.0000);
\node[inner sep=1pt,] at (3.3136,-0.2400) {-4};
\draw[draw=black!12,line width=.35pt] (3.9409,0.0000) -- (3.9409,4.0000);
\node[inner sep=1pt,] at (3.9409,-0.2400) {-2};
\draw[draw=black!12,line width=.35pt] (4.5682,0.0000) -- (4.5682,4.0000);
\node[inner sep=1pt,] at (4.5682,-0.2400) {0};
\draw[draw=black!12,line width=.35pt] (5.1955,0.0000) -- (5.1955,4.0000);
\node[inner sep=1pt,] at (5.1955,-0.2400) {2};
\draw[draw=black!12,line width=.35pt] (5.8227,0.0000) -- (5.8227,4.0000);
\node[inner sep=1pt,] at (5.8227,-0.2400) {4};
\draw[draw=black!12,line width=.35pt] (6.4500,0.0000) -- (6.4500,4.0000);
\node[inner sep=1pt,] at (6.4500,-0.2400) {6};
\draw[draw=black!60,dashed,line width=.55pt] (4.5682,0.0000) -- (4.5682,4.0000);
\draw[draw=black!55,line width=.45pt] (3.0000,0.0000) -- (6.4500,0.0000);
\draw[draw=evT,line width=.8pt] (4.3369,3.6800) -- (4.7603,3.6800);
\draw[draw=evT,line width=.7pt] (4.3369,3.6350) -- (4.3369,3.7250);
\draw[draw=evT,line width=.7pt] (4.7603,3.6350) -- (4.7603,3.7250);
\fill[evT] (4.5486,3.6800) circle (1.6pt);
\draw[draw=evT,line width=.8pt] (3.5881,3.2100) -- (6.1364,3.2100);
\draw[draw=evT,line width=.7pt] (3.5881,3.1650) -- (3.5881,3.2550);
\draw[draw=evT,line width=.7pt] (6.1364,3.1650) -- (6.1364,3.2550);
\fill[evT] (4.8949,3.2100) circle (1.6pt);
\draw[draw=evT,line width=.8pt] (3.5227,2.7400) -- (5.2216,2.7400);
\draw[draw=evT,line width=.7pt] (3.5227,2.6950) -- (3.5227,2.7850);
\draw[draw=evT,line width=.7pt] (5.2216,2.6950) -- (5.2216,2.7850);
\fill[evT] (4.4375,2.7400) circle (1.6pt);
\draw[draw=evT,line width=.8pt] (3.0980,2.2700) -- (5.3033,2.2700);
\draw[draw=evT,line width=.7pt] (3.0980,2.2250) -- (3.0980,2.3150);
\draw[draw=evT,line width=.7pt] (5.3033,2.2250) -- (5.3033,2.3150);
\fill[evT] (4.2251,2.2700) circle (1.6pt);
\draw[draw=evT,line width=.8pt] (4.4805,1.8000) -- (5.0068,1.8000);
\draw[draw=evT,line width=.7pt] (4.4805,1.7550) -- (4.4805,1.8450);
\draw[draw=evT,line width=.7pt] (5.0068,1.7550) -- (5.0068,1.8450);
\fill[evT] (4.7436,1.8000) circle (1.6pt);
\draw[draw=evT,line width=.8pt] (3.5485,1.3300) -- (4.7365,1.3300);
\draw[draw=evT,line width=.7pt] (3.5485,1.2850) -- (3.5485,1.3750);
\draw[draw=evT,line width=.7pt] (4.7365,1.2850) -- (4.7365,1.3750);
\fill[evT] (4.1326,1.3300) circle (1.6pt);
\draw[draw=evS,line width=.8pt] (4.4004,0.8600) -- (4.7373,0.8600);
\draw[draw=evS,line width=.7pt] (4.4004,0.8150) -- (4.4004,0.9050);
\draw[draw=evS,line width=.7pt] (4.7373,0.8150) -- (4.7373,0.9050);
\fill[evS] (4.4918,0.8600) -- (4.5668,0.9350) -- (4.6418,0.8600) -- (4.5668,0.7850) -- cycle;
\draw[draw=evS,line width=.8pt] (4.1530,0.3900) -- (4.8324,0.3900);
\draw[draw=evS,line width=.7pt] (4.1530,0.3450) -- (4.1530,0.4350);
\draw[draw=evS,line width=.7pt] (4.8324,0.3450) -- (4.8324,0.4350);
\fill[evS] (4.4221,0.3900) -- (4.4971,0.4650) -- (4.5721,0.3900) -- (4.4971,0.3150) -- cycle;
\node[inner sep=1pt,] at (4.7250,4.3500) {$\Delta$ Pass@1 (pp)};
\node[inner sep=1pt,] at (4.7250,-0.6300) {(a) Single-sample accuracy};
\fill[black!4] (7.100,0) rectangle (10.550,.98);
\draw[draw=black!12,line width=.35pt] (7.3875,0.0000) -- (7.3875,4.0000);
\node[inner sep=1pt,] at (7.3875,-0.2400) {-5};
\draw[draw=black!12,line width=.35pt] (7.8667,0.0000) -- (7.8667,4.0000);
\node[inner sep=1pt,] at (7.8667,-0.2400) {0};
\draw[draw=black!12,line width=.35pt] (8.8250,0.0000) -- (8.8250,4.0000);
\node[inner sep=1pt,] at (8.8250,-0.2400) {10};
\draw[draw=black!12,line width=.35pt] (9.7833,0.0000) -- (9.7833,4.0000);
\node[inner sep=1pt,] at (9.7833,-0.2400) {20};
\draw[draw=black!60,dashed,line width=.55pt] (7.8667,0.0000) -- (7.8667,4.0000);
\draw[draw=black!55,line width=.45pt] (7.1000,0.0000) -- (10.5500,0.0000);
\draw[draw=evT,line width=.8pt] (7.7900,3.6800) -- (7.9817,3.6800);
\draw[draw=evT,line width=.7pt] (7.7900,3.6350) -- (7.7900,3.7250);
\draw[draw=evT,line width=.7pt] (7.9817,3.6350) -- (7.9817,3.7250);
\fill[evT] (7.8858,3.6800) circle (1.6pt);
\draw[draw=evT,line width=.8pt] (7.2278,3.2100) -- (10.4222,3.2100);
\draw[draw=evT,line width=.7pt] (7.2278,3.1650) -- (7.2278,3.2550);
\draw[draw=evT,line width=.7pt] (10.4222,3.1650) -- (10.4222,3.2550);
\fill[evT] (8.8250,3.2100) circle (1.6pt);
\draw[draw=evT,line width=.8pt] (7.8667,2.7400) -- (9.4639,2.7400);
\draw[draw=evT,line width=.7pt] (7.8667,2.6950) -- (7.8667,2.7850);
\draw[draw=evT,line width=.7pt] (9.4639,2.6950) -- (9.4639,2.7850);
\fill[evT] (8.5056,2.7400) circle (1.6pt);
\draw[draw=evT,line width=.8pt] (7.8667,2.2700) -- (8.5854,2.2700);
\draw[draw=evT,line width=.7pt] (7.8667,2.2250) -- (7.8667,2.3150);
\draw[draw=evT,line width=.7pt] (8.5854,2.2250) -- (8.5854,2.3150);
\fill[evT] (8.1062,2.2700) circle (1.6pt);
\draw[draw=evT,line width=.8pt] (7.9162,1.8000) -- (8.2955,1.8000);
\draw[draw=evT,line width=.7pt] (7.9162,1.7550) -- (7.9162,1.8450);
\draw[draw=evT,line width=.7pt] (8.2955,1.7550) -- (8.2955,1.8450);
\fill[evT] (8.0976,1.8000) circle (1.6pt);
\draw[draw=evT,line width=.8pt] (7.4795,1.3300) -- (8.3507,1.3300);
\draw[draw=evT,line width=.7pt] (7.4795,1.2850) -- (7.4795,1.3750);
\draw[draw=evT,line width=.7pt] (8.3507,1.2850) -- (8.3507,1.3750);
\fill[evT] (7.9151,1.3300) circle (1.6pt);
\draw[draw=evS,line width=.8pt] (7.9084,0.8600) -- (8.1377,0.8600);
\draw[draw=evS,line width=.7pt] (7.9084,0.8150) -- (7.9084,0.9050);
\draw[draw=evS,line width=.7pt] (8.1377,0.8150) -- (8.1377,0.9050);
\fill[evS] (7.9446,0.8600) -- (8.0196,0.9350) -- (8.0946,0.8600) -- (8.0196,0.7850) -- cycle;
\draw[draw=evS,line width=.8pt] (7.9071,0.3900) -- (8.5496,0.3900);
\draw[draw=evS,line width=.7pt] (7.9071,0.3450) -- (7.9071,0.4350);
\draw[draw=evS,line width=.7pt] (8.5496,0.3450) -- (8.5496,0.4350);
\fill[evS] (8.1475,0.3900) -- (8.2225,0.4650) -- (8.2975,0.3900) -- (8.2225,0.3150) -- cycle;
\node[inner sep=1pt,] at (8.8250,4.3500) {$\Delta$ Pass@16 (pp)};
\node[inner sep=1pt,] at (8.8250,-0.6300) {(b) Multi-sample coverage};
\fill[black!4] (11.200,0) rectangle (14.650,.98);
\draw[draw=black!12,line width=.35pt] (11.2000,0.0000) -- (11.2000,4.0000);
\node[inner sep=1pt,] at (11.2000,-0.2400) {-800};
\draw[draw=black!12,line width=.35pt] (12.5143,0.0000) -- (12.5143,4.0000);
\node[inner sep=1pt,] at (12.5143,-0.2400) {-400};
\draw[draw=black!12,line width=.35pt] (13.8286,0.0000) -- (13.8286,4.0000);
\node[inner sep=1pt,] at (13.8286,-0.2400) {0};
\draw[draw=black!60,dashed,line width=.55pt] (13.8286,0.0000) -- (13.8286,4.0000);
\draw[draw=black!55,line width=.45pt] (11.2000,0.0000) -- (14.6500,0.0000);
\draw[draw=evT,line width=.8pt] (12.9801,3.6800) -- (13.4524,3.6800);
\draw[draw=evT,line width=.7pt] (12.9801,3.6350) -- (12.9801,3.7250);
\draw[draw=evT,line width=.7pt] (13.4524,3.6350) -- (13.4524,3.7250);
\fill[evT] (13.2184,3.6800) circle (1.6pt);
\draw[draw=evT,line width=.8pt] (11.7193,3.2100) -- (14.1686,3.2100);
\draw[draw=evT,line width=.7pt] (11.7193,3.1650) -- (11.7193,3.2550);
\draw[draw=evT,line width=.7pt] (14.1686,3.1650) -- (14.1686,3.2550);
\fill[evT] (12.9433,3.2100) circle (1.6pt);
\draw[draw=evT,line width=.8pt] (11.3903,2.7400) -- (13.8980,2.7400);
\draw[draw=evT,line width=.7pt] (11.3903,2.6950) -- (11.3903,2.7850);
\draw[draw=evT,line width=.7pt] (13.8980,2.6950) -- (13.8980,2.7850);
\fill[evT] (12.6416,2.7400) circle (1.6pt);
\draw[draw=evT,line width=.8pt] (12.6789,2.2700) -- (14.5185,2.2700);
\draw[draw=evT,line width=.7pt] (12.6789,2.2250) -- (12.6789,2.3150);
\draw[draw=evT,line width=.7pt] (14.5185,2.2250) -- (14.5185,2.3150);
\fill[evT] (13.5842,2.2700) circle (1.6pt);
\draw[draw=evT,line width=.8pt] (12.4541,1.8000) -- (13.0043,1.8000);
\draw[draw=evT,line width=.7pt] (12.4541,1.7550) -- (12.4541,1.8450);
\draw[draw=evT,line width=.7pt] (13.0043,1.7550) -- (13.0043,1.8450);
\fill[evT] (12.7280,1.8000) circle (1.6pt);
\draw[draw=evT,line width=.8pt] (13.3792,1.3300) -- (14.2362,1.3300);
\draw[draw=evT,line width=.7pt] (13.3792,1.2850) -- (13.3792,1.3750);
\draw[draw=evT,line width=.7pt] (14.2362,1.2850) -- (14.2362,1.3750);
\fill[evT] (13.8107,1.3300) circle (1.6pt);
\draw[draw=evS,line width=.8pt] (12.9248,0.8600) -- (13.2544,0.8600);
\draw[draw=evS,line width=.7pt] (12.9248,0.8150) -- (12.9248,0.9050);
\draw[draw=evS,line width=.7pt] (13.2544,0.8150) -- (13.2544,0.9050);
\fill[evS] (13.0139,0.8600) -- (13.0889,0.9350) -- (13.1639,0.8600) -- (13.0889,0.7850) -- cycle;
\draw[draw=evS,line width=.8pt] (12.8002,0.3900) -- (13.4945,0.3900);
\draw[draw=evS,line width=.7pt] (12.8002,0.3450) -- (12.8002,0.4350);
\draw[draw=evS,line width=.7pt] (13.4945,0.3450) -- (13.4945,0.4350);
\fill[evS] (13.0794,0.3900) -- (13.1544,0.4650) -- (13.2294,0.3900) -- (13.1544,0.3150) -- cycle;
\node[inner sep=1pt,] at (12.9250,4.3500) {$\Delta$ mean tokens};
\node[inner sep=1pt,] at (12.9250,-0.6300) {(c) Output length};
\node[inner sep=1pt,anchor=west] at (3.0000,-1.0000) {PCRS-T minus quarter-DECS; points and 95\% paired intervals};
\end{tikzpicture}}
\caption{Benchmark heterogeneity in the PCRS-T--DECS comparison. Error bars are paired question-bootstrap intervals for the fixed checkpoints; shaded rows show the two full-suite aggregation schemes. Rightward values indicate higher accuracy or coverage in the first two panels, whereas leftward values indicate fewer \mbox{tokens in the third.}}
\label{fig:paired_evidence}\end{figure*}

\subsection{Data and Output-Budget Analysis}
\label{app:fixed_budget}
The data contain 10,112 training rows and 492 validation rows. Exact-question matching uses Unicode NFKC, case folding, and whitespace normalization, retaining mathematical content and removing the terminal generation instruction. Training contains 10,023 unique normalized questions and 89 duplicate rows, with no train--validation matches.

For an offline common-budget comparison, we consume responses in their original generation order and count only complete responses whose cumulative output length fits the per-question budget. An overflowing response consumes the remaining budget but contributes no correct answer. No response is skipped or sorted by length or correctness, and generation stops at the 16-response cache ceiling. Oracle coverage is the fraction of questions with at least one correct completed response. It is neither accuracy from an answer-selection policy nor measured wall-clock efficiency. Table~\ref{tab:fixed_budget} reports all six budget points; the paired percentile intervals use 10,000 benchmark-stratified question draws (seed 20261008), without multiplicity adjustment.

\begin{figure*}[t]\centering
\resizebox{\textwidth}{!}{\begin{tikzpicture}[x=1cm,y=1cm,font=\fontsize{9}{11}\selectfont]
\definecolor{evB}{HTML}{777777}
\definecolor{evT}{HTML}{4477AA}
\definecolor{evS}{HTML}{008F82}
\definecolor{evE}{HTML}{CC6677}
\definecolor{evMath}{HTML}{4477AA}
\definecolor{evGPQA}{HTML}{B47716}
\draw[draw=black!12,line width=.35pt] (1.0000,0.4056) -- (7.1500,0.4056);
\node[inner sep=1pt,anchor=east] at (0.8500,0.4056) {50};
\draw[draw=black!12,line width=.35pt] (1.0000,1.2167) -- (7.1500,1.2167);
\node[inner sep=1pt,anchor=east] at (0.8500,1.2167) {60};
\draw[draw=black!12,line width=.35pt] (1.0000,2.0278) -- (7.1500,2.0278);
\node[inner sep=1pt,anchor=east] at (0.8500,2.0278) {70};
\draw[draw=black!12,line width=.35pt] (1.0000,2.8389) -- (7.1500,2.8389);
\node[inner sep=1pt,anchor=east] at (0.8500,2.8389) {80};
\draw[draw=black!12,line width=.35pt] (1.0000,3.6500) -- (7.1500,3.6500);
\node[inner sep=1pt,anchor=east] at (0.8500,3.6500) {90};
\draw[draw=black!55,line width=.45pt] (1.0000,0.0000) -- (7.1500,0.0000);
\draw[draw=black!55,line width=.45pt] (1.0000,0.0000) -- (1.0000,3.6500);
\draw[draw=black!55,line width=.45pt] (1.0000,0.0000) -- (1.0000,-0.0700);
\node[inner sep=1pt,] at (1.0000,-0.2400) {8k};
\draw[draw=black!55,line width=.45pt] (2.2300,0.0000) -- (2.2300,-0.0700);
\node[inner sep=1pt,] at (2.2300,-0.2400) {16k};
\draw[draw=black!55,line width=.45pt] (3.4600,0.0000) -- (3.4600,-0.0700);
\node[inner sep=1pt,] at (3.4600,-0.2400) {32k};
\draw[draw=black!55,line width=.45pt] (4.6900,0.0000) -- (4.6900,-0.0700);
\node[inner sep=1pt,] at (4.6900,-0.2400) {64k};
\draw[draw=black!55,line width=.45pt] (5.9200,0.0000) -- (5.9200,-0.0700);
\node[inner sep=1pt,] at (5.9200,-0.2400) {128k};
\draw[draw=black!55,line width=.45pt] (7.1500,0.0000) -- (7.1500,-0.0700);
\node[inner sep=1pt,] at (7.1500,-0.2400) {256k};
\node[inner sep=1pt,] at (4.0750,-0.6300) {Output-token budget per question};
\node[inner sep=1pt,] at (4.0750,-0.9600) {(a) All four checkpoints};
\node[inner sep=1pt,rotate=90] at (0.2700,1.8250) {Oracle coverage (\%)};
\draw[draw=evB,line width=.85pt,dashed] (1.0000,0.6732) -- (2.2300,1.6437);
\draw[draw=evB,line width=.85pt,dashed] (2.2300,1.6437) -- (3.4600,2.3554);
\draw[draw=evB,line width=.85pt,dashed] (3.4600,2.3554) -- (4.6900,2.8789);
\draw[draw=evB,line width=.85pt,dashed] (4.6900,2.8789) -- (5.9200,3.2083);
\draw[draw=evB,line width=.85pt,dashed] (5.9200,3.2083) -- (7.1500,3.3318);
\fill[evB] (1.0000,0.6732) circle (1.6pt);
\fill[evB] (2.2300,1.6437) circle (1.6pt);
\fill[evB] (3.4600,2.3554) circle (1.6pt);
\fill[evB] (4.6900,2.8789) circle (1.6pt);
\fill[evB] (5.9200,3.2083) circle (1.6pt);
\fill[evB] (7.1500,3.3318) circle (1.6pt);
\draw[draw=evT,line width=.85pt,densely dotted] (1.0000,0.8202) -- (2.2300,1.6907);
\draw[draw=evT,line width=.85pt,densely dotted] (2.2300,1.6907) -- (3.4600,2.3319);
\draw[draw=evT,line width=.85pt,densely dotted] (3.4600,2.3319) -- (4.6900,2.8436);
\draw[draw=evT,line width=.85pt,densely dotted] (4.6900,2.8436) -- (5.9200,3.1906);
\draw[draw=evT,line width=.85pt,densely dotted] (5.9200,3.1906) -- (7.1500,3.3906);
\fill[evT] (0.9480,0.7682) rectangle (1.0520,0.8722);
\fill[evT] (2.1780,1.6387) rectangle (2.2820,1.7427);
\fill[evT] (3.4080,2.2799) rectangle (3.5120,2.3839);
\fill[evT] (4.6380,2.7916) rectangle (4.7420,2.8956);
\fill[evT] (5.8680,3.1386) rectangle (5.9720,3.2426);
\fill[evT] (7.0980,3.3386) rectangle (7.2020,3.4426);
\draw[draw=evE,line width=.85pt,dashdotted] (1.0000,0.6673) -- (2.2300,1.6202);
\draw[draw=evE,line width=.85pt,dashdotted] (2.2300,1.6202) -- (3.4600,2.3672);
\draw[draw=evE,line width=.85pt,dashdotted] (3.4600,2.3672) -- (4.6900,2.8024);
\draw[draw=evE,line width=.85pt,dashdotted] (4.6900,2.8024) -- (5.9200,3.1847);
\draw[draw=evE,line width=.85pt,dashdotted] (5.9200,3.1847) -- (7.1500,3.3259);
\fill[evE] (0.9480,0.6153) rectangle (1.0520,0.7193);
\fill[evE] (2.1780,1.5682) rectangle (2.2820,1.6722);
\fill[evE] (3.4080,2.3152) rectangle (3.5120,2.4192);
\fill[evE] (4.6380,2.7504) rectangle (4.7420,2.8544);
\fill[evE] (5.8680,3.1327) rectangle (5.9720,3.2367);
\fill[evE] (7.0980,3.2739) rectangle (7.2020,3.3779);
\draw[draw=evS,line width=.85pt,solid] (1.0000,0.8614) -- (2.2300,1.7966);
\draw[draw=evS,line width=.85pt,solid] (2.2300,1.7966) -- (3.4600,2.4554);
\draw[draw=evS,line width=.85pt,solid] (3.4600,2.4554) -- (4.6900,2.8671);
\draw[draw=evS,line width=.85pt,solid] (4.6900,2.8671) -- (5.9200,3.2965);
\draw[draw=evS,line width=.85pt,solid] (5.9200,3.2965) -- (7.1500,3.4612);
\fill[evS] (0.9250,0.8614) -- (1.0000,0.9364) -- (1.0750,0.8614) -- (1.0000,0.7864) -- cycle;
\fill[evS] (2.1550,1.7966) -- (2.2300,1.8716) -- (2.3050,1.7966) -- (2.2300,1.7216) -- cycle;
\fill[evS] (3.3850,2.4554) -- (3.4600,2.5304) -- (3.5350,2.4554) -- (3.4600,2.3804) -- cycle;
\fill[evS] (4.6150,2.8671) -- (4.6900,2.9421) -- (4.7650,2.8671) -- (4.6900,2.7921) -- cycle;
\fill[evS] (5.8450,3.2965) -- (5.9200,3.3715) -- (5.9950,3.2965) -- (5.9200,3.2215) -- cycle;
\fill[evS] (7.0750,3.4612) -- (7.1500,3.5362) -- (7.2250,3.4612) -- (7.1500,3.3862) -- cycle;
\draw[draw=evB,line width=.85pt] (1.0000,4.3000) -- (1.3600,4.3000);
\node[inner sep=1pt,anchor=west] at (1.4600,4.3000) {quarter-DECS};
\draw[draw=evT,line width=.85pt] (4.2500,4.3000) -- (4.6100,4.3000);
\node[inner sep=1pt,anchor=west] at (4.7100,4.3000) {PCRS-H};
\draw[draw=evS,line width=.85pt] (1.0000,3.9700) -- (1.3600,3.9700);
\node[inner sep=1pt,anchor=west] at (1.4600,3.9700) {default PCRS-T};
\draw[draw=evE,line width=.85pt] (4.2500,3.9700) -- (4.6100,3.9700);
\node[inner sep=1pt,anchor=west] at (4.7100,3.9700) {E-priority};
\draw[draw=black!12,line width=.35pt] (9.0000,0.4562) -- (15.1500,0.4562);
\node[inner sep=1pt,anchor=east] at (8.8500,0.4562) {-2};
\draw[draw=black!12,line width=.35pt] (9.0000,1.3687) -- (15.1500,1.3687);
\node[inner sep=1pt,anchor=east] at (8.8500,1.3687) {0};
\draw[draw=black!12,line width=.35pt] (9.0000,2.2812) -- (15.1500,2.2812);
\node[inner sep=1pt,anchor=east] at (8.8500,2.2812) {2};
\draw[draw=black!12,line width=.35pt] (9.0000,3.1938) -- (15.1500,3.1938);
\node[inner sep=1pt,anchor=east] at (8.8500,3.1938) {4};
\draw[draw=black!55,line width=.45pt] (9.0000,0.0000) -- (15.1500,0.0000);
\draw[draw=black!55,line width=.45pt] (9.0000,0.0000) -- (9.0000,3.6500);
\draw[draw=black!55,line width=.45pt] (9.0000,0.0000) -- (9.0000,-0.0700);
\node[inner sep=1pt,] at (9.0000,-0.2400) {8k};
\draw[draw=black!55,line width=.45pt] (10.2300,0.0000) -- (10.2300,-0.0700);
\node[inner sep=1pt,] at (10.2300,-0.2400) {16k};
\draw[draw=black!55,line width=.45pt] (11.4600,0.0000) -- (11.4600,-0.0700);
\node[inner sep=1pt,] at (11.4600,-0.2400) {32k};
\draw[draw=black!55,line width=.45pt] (12.6900,0.0000) -- (12.6900,-0.0700);
\node[inner sep=1pt,] at (12.6900,-0.2400) {64k};
\draw[draw=black!55,line width=.45pt] (13.9200,0.0000) -- (13.9200,-0.0700);
\node[inner sep=1pt,] at (13.9200,-0.2400) {128k};
\draw[draw=black!55,line width=.45pt] (15.1500,0.0000) -- (15.1500,-0.0700);
\node[inner sep=1pt,] at (15.1500,-0.2400) {256k};
\node[inner sep=1pt,] at (12.0750,-0.6300) {Output-token budget per question};
\node[inner sep=1pt,] at (12.0750,-0.9600) {(b) PCRS-T minus quarter-DECS};
\node[inner sep=1pt,rotate=90] at (8.2700,1.8250) {Coverage difference (pp)};
\draw[draw=black!60,dashed,line width=.55pt] (9.0000,1.3687) -- (15.1500,1.3687);
\draw[draw=evS,line width=.7pt] (8.9400,1.5011) -- (8.9400,3.3870);
\draw[draw=evS,line width=.7pt] (8.8850,1.5011) -- (8.9950,1.5011);
\draw[draw=evS,line width=.7pt] (8.8850,3.3870) -- (8.9950,3.3870);
\fill[evS] (8.8650,2.4275) -- (8.9400,2.5025) -- (9.0150,2.4275) -- (8.9400,2.3525) -- cycle;
\draw[draw=evS,line width=.7pt] (10.1700,1.2695) -- (10.1700,3.1885);
\draw[draw=evS,line width=.7pt] (10.1150,1.2695) -- (10.2250,1.2695);
\draw[draw=evS,line width=.7pt] (10.1150,3.1885) -- (10.2250,3.1885);
\fill[evS] (10.0950,2.2290) -- (10.1700,2.3040) -- (10.2450,2.2290) -- (10.1700,2.1540) -- cycle;
\draw[draw=evS,line width=.7pt] (11.4000,1.1041) -- (11.4000,2.7583);
\draw[draw=evS,line width=.7pt] (11.3450,1.1041) -- (11.4550,1.1041);
\draw[draw=evS,line width=.7pt] (11.3450,2.7583) -- (11.4550,2.7583);
\fill[evS] (11.3250,1.9312) -- (11.4000,2.0062) -- (11.4750,1.9312) -- (11.4000,1.8562) -- cycle;
\draw[draw=evS,line width=.7pt] (12.6300,0.6078) -- (12.6300,1.9974);
\draw[draw=evS,line width=.7pt] (12.5750,0.6078) -- (12.6850,0.6078);
\draw[draw=evS,line width=.7pt] (12.5750,1.9974) -- (12.6850,1.9974);
\fill[evS] (12.5550,1.3026) -- (12.6300,1.3776) -- (12.7050,1.3026) -- (12.6300,1.2276) -- cycle;
\draw[draw=evS,line width=.7pt] (13.8600,1.2695) -- (13.8600,2.4606);
\draw[draw=evS,line width=.7pt] (13.8050,1.2695) -- (13.9150,1.2695);
\draw[draw=evS,line width=.7pt] (13.8050,2.4606) -- (13.9150,2.4606);
\fill[evS] (13.7850,1.8650) -- (13.8600,1.9400) -- (13.9350,1.8650) -- (13.8600,1.7900) -- cycle;
\draw[draw=evS,line width=.7pt] (15.0900,1.5342) -- (15.0900,2.6591);
\draw[draw=evS,line width=.7pt] (15.0350,1.5342) -- (15.1450,1.5342);
\draw[draw=evS,line width=.7pt] (15.0350,2.6591) -- (15.1450,2.6591);
\fill[evS] (15.0150,2.0966) -- (15.0900,2.1716) -- (15.1650,2.0966) -- (15.0900,2.0216) -- cycle;
\fill[evS] (9.0750,4.3000) -- (9.1500,4.3750) -- (9.2250,4.3000) -- (9.1500,4.2250) -- cycle;
\node[inner sep=1pt,anchor=west] at (9.3600,4.3000) {All 1,379 questions};
\end{tikzpicture}}
\caption{Coverage under a common output-token budget. Responses are consumed in their original order; an incomplete response contributes no success. Left: all four frozen checkpoints. Right: paired PCRS-T--DECS differences with 95\% question-bootstrap intervals on all 1,379 questions. Every pre-specified budget is shown; $k=1{,}024$ tokens. The cache contains at most 16 responses per question. Oracle coverage measures available correct answers, not the accuracy of an answer selector.}
\label{fig:budget_evidence}\end{figure*}
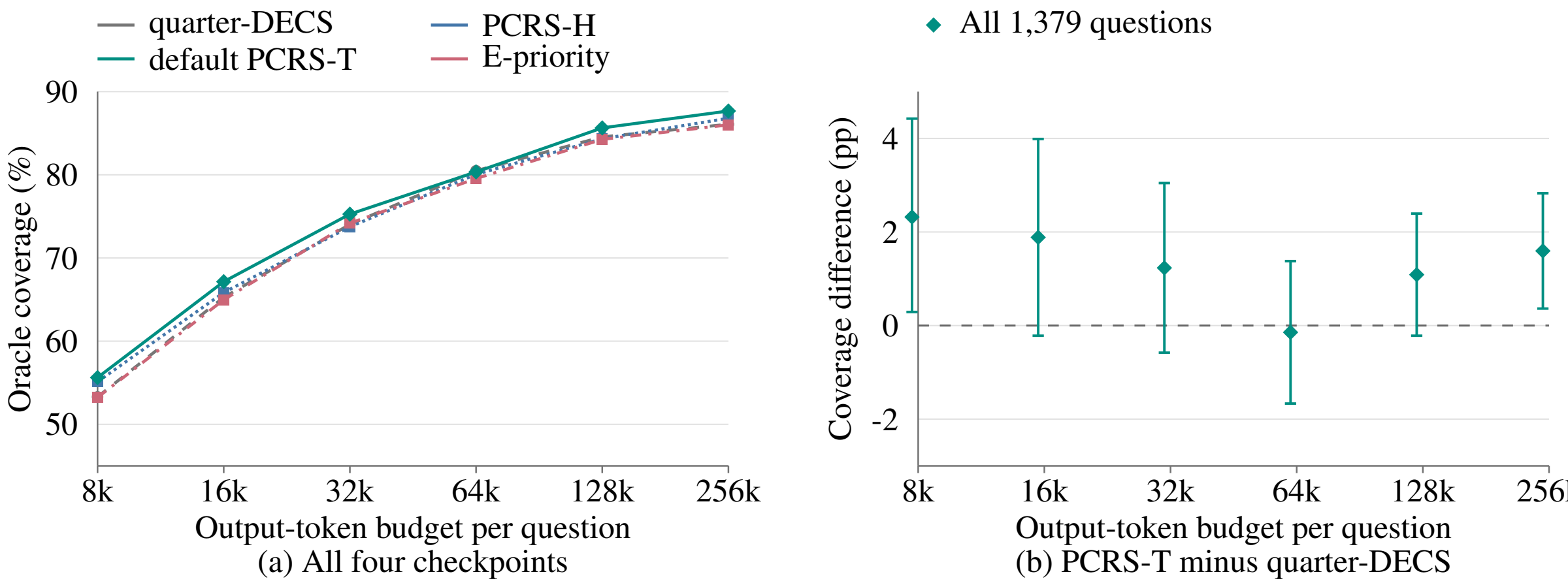

At 8,192 tokens, no response completes for 38.36\% of DECS questions and 36.04\% of PCRS-T questions. At 65,536 tokens, average completed-response counts are 10.46 and 10.63, while 37.42\% and 38.80\% of questions already reach the 16-response ceiling. These observations distinguish output cost, response availability, \mbox{and oracle correctness.}

\subsection{Training window values}
\label{app:training_role_history}
\label{sec:trainingroles}

During default PCRS-T training, the exclusive R bucket rises from 4.04\% in updates 1--5 to 4.86\% in updates 48--52, while the multilabel bucket changes from 68.95\% to 67.39\%. Figure~\ref{fig:trainingroles} shows all 52 updates; the definitions and \mbox{exact window values follow.}

We analyze the existing 52-update PCRS-T history. The five logged buckets are mutually exclusive: R, E, T, M (RE/RT/ET/RET combined), and U (unclassified or protected). The denominator is all valid response tokens, including incorrect responses and protected final answers, before reward-eligibility filtering. These fractions are not the fraction of tokens whose reward changes.

Multilabel tokens dominate, with M ranging from 65.66\% to 71.55\% across updates. Comparing the fixed early window (1--5) and late window (48--52), the equally weighted step-mean R share rises from 4.0422\% to 4.8566\% (+0.8143 percentage points), M falls from 68.9465\% to 67.3933\% ($-1.5532$ points), and U rises from 6.8556\% to 7.5841\% (+0.7286 points). E changes by just +0.0111 points between the windows, although its first-to-last difference is +0.6915 points. The endpoint difference is therefore a poor summary of the window comparison.

T averages 0.004118\% and 0.003334\% in the two windows and is zero in 17 of 52 updates, including the last. T can also occur in M; without separate RE/RT/ET/RET counts, marginal phase shares cannot be recovered. Missing per-step denominators prevent a token-weighted whole-run mean. Figure~\ref{fig:trainingroles} connects all unsmoothed observations from changing batches, not repeated measurements of fixed questions.

Appendix~\ref{app:rolechanges} compares the same fixed windows across the completed reward settings. Across the completed settings, the R bucket increases from the early to the late training window, while changes in E and multilabel \mbox{shares vary by configuration.}

Table~\ref{tab:rolewindows} records the underlying token shares, and Figure~\ref{fig:rolechanges} shows the same early and late windows for each completed reward setting.
\begin{table*}[t]
\centering
\small
\setlength{\tabcolsep}{4pt}
\begin{tabular}{lrrrrrr}
\toprule
Bucket & Step 1 & Step 52 & $\Delta$ (pp) & Mean 1--5 & Mean 48--52 & $\Delta$ (pp) \\
\midrule
R & 4.0757 & 4.3380 & +0.2623 & 4.0422 & 4.8566 & +0.8143 \\
E & 19.6729 & 20.3644 & +0.6915 & 20.1516 & 20.1627 & +0.0111 \\
T & 0.000911 & 0.000000 & -0.000911 & 0.004118 & 0.003334 & -0.000785 \\
M & 69.1865 & 68.0526 & -1.1339 & 68.9465 & 67.3933 & -1.5532 \\
U & 7.0640 & 7.2451 & +0.1811 & 6.8556 & 7.5841 & +0.7286 \\
\bottomrule
\end{tabular}
\caption{Training token phase-label shares (percent). Deltas are later minus earlier, in percentage points (pp).
Window means weight the five training updates equally, not by token count. T uses six decimal places to preserve its small nonzero values; other buckets use four.
M pools RE/RT/ET/RET; U includes \mbox{unclassified and protected tokens.}}
\label{tab:rolewindows}
\end{table*}

\subsection{Training-Window Comparisons}
\label{app:rolechanges}
Figure~\ref{fig:rolechanges} reports descriptive early-to-late changes for default PCRS-T and the three other completed weight settings. Each uses all 52 recorded updates and the same fixed five-update windows. These changing-batch comparisons summarize each run's rollout trajectories. Across settings, the exclusive R share increases, while \mbox{E and mixed-label shares vary.}

\FloatBarrier

\brokenpenalty=10000
\twocolumn[{%
\begin{minipage}{\textwidth}
\captionsetup{type=figure}
\centering
\resizebox{\textwidth}{!}{%
\begin{tikzpicture}[x=1cm,y=1cm,font=\rmfamily\fontsize{9}{11}\selectfont,line cap=round,line join=round]
\definecolor{trR}{HTML}{397DB5}
\definecolor{trE}{HTML}{B78429}
\definecolor{trT}{HTML}{AA6187}
\definecolor{trM}{HTML}{788D3F}
\definecolor{trU}{HTML}{737C87}
\definecolor{trgrid}{HTML}{DDDDDD}
\definecolor{trwindow}{HTML}{F1F2F4}
\definecolor{traxis}{HTML}{777777}
\definecolor{trblack}{HTML}{222222}
\node[anchor=center,inner sep=1pt,font=\rmfamily\fontsize{9}{11}\selectfont] at (8.45,7.155) {{\color{trR}\rule{5pt}{5pt}}\; R\quad {\color{trE}\rule{5pt}{5pt}}\; E\quad {\color{trT}\rule{5pt}{5pt}}\; T\quad {\color{trM}\rule{5pt}{5pt}}\; Multilabel\quad {\color{trU}\rule{5pt}{5pt}}\; Other / protected};
\node[anchor=center,inner sep=1pt,font=\rmfamily\fontsize{10}{11}\selectfont] at (2.95000,6.60000) {(a) Overall composition};
\node[anchor=center,inner sep=1pt,font=\rmfamily\fontsize{9}{11}\selectfont] at (2.95000,6.23000) {All valid tokens; full 0--100\% scale};
\fill[trR] (0.70000,4.40000) -- (0.78824,4.40000) -- (0.87647,4.40000) -- (0.96471,4.40000) -- (1.05294,4.40000) -- (1.14118,4.40000) -- (1.22941,4.40000) -- (1.31765,4.40000) -- (1.40588,4.40000) -- (1.49412,4.40000) -- (1.58235,4.40000) -- (1.67059,4.40000) -- (1.75882,4.40000) -- (1.84706,4.40000) -- (1.93529,4.40000) -- (2.02353,4.40000) -- (2.11176,4.40000) -- (2.20000,4.40000) -- (2.28824,4.40000) -- (2.37647,4.40000) -- (2.46471,4.40000) -- (2.55294,4.40000) -- (2.64118,4.40000) -- (2.72941,4.40000) -- (2.81765,4.40000) -- (2.90588,4.40000) -- (2.99412,4.40000) -- (3.08235,4.40000) -- (3.17059,4.40000) -- (3.25882,4.40000) -- (3.34706,4.40000) -- (3.43529,4.40000) -- (3.52353,4.40000) -- (3.61176,4.40000) -- (3.70000,4.40000) -- (3.78824,4.40000) -- (3.87647,4.40000) -- (3.96471,4.40000) -- (4.05294,4.40000) -- (4.14118,4.40000) -- (4.22941,4.40000) -- (4.31765,4.40000) -- (4.40588,4.40000) -- (4.49412,4.40000) -- (4.58235,4.40000) -- (4.67059,4.40000) -- (4.75882,4.40000) -- (4.84706,4.40000) -- (4.93529,4.40000) -- (5.02353,4.40000) -- (5.11176,4.40000) -- (5.20000,4.40000) -- (5.20000,4.46724) -- (5.11176,4.47327) -- (5.02353,4.47295) -- (4.93529,4.47612) -- (4.84706,4.48681) -- (4.75882,4.46745) -- (4.67059,4.47139) -- (4.58235,4.46742) -- (4.49412,4.46770) -- (4.40588,4.47396) -- (4.31765,4.47451) -- (4.22941,4.47055) -- (4.14118,4.47970) -- (4.05294,4.46486) -- (3.96471,4.47523) -- (3.87647,4.47247) -- (3.78824,4.47033) -- (3.70000,4.46735) -- (3.61176,4.48107) -- (3.52353,4.46574) -- (3.43529,4.46457) -- (3.34706,4.46402) -- (3.25882,4.46532) -- (3.17059,4.46825) -- (3.08235,4.46992) -- (2.99412,4.47209) -- (2.90588,4.46917) -- (2.81765,4.47657) -- (2.72941,4.47405) -- (2.64118,4.48446) -- (2.55294,4.46823) -- (2.46471,4.47180) -- (2.37647,4.46967) -- (2.28824,4.47811) -- (2.20000,4.46401) -- (2.11176,4.46432) -- (2.02353,4.47273) -- (1.93529,4.47828) -- (1.84706,4.46956) -- (1.75882,4.46926) -- (1.67059,4.46685) -- (1.58235,4.46708) -- (1.49412,4.45977) -- (1.40588,4.46693) -- (1.31765,4.46885) -- (1.22941,4.47279) -- (1.14118,4.46275) -- (1.05294,4.46501) -- (0.96471,4.46032) -- (0.87647,4.45868) -- (0.78824,4.46610) -- (0.70000,4.46317) -- cycle;
\fill[trE] (0.70000,4.46317) -- (0.78824,4.46610) -- (0.87647,4.45868) -- (0.96471,4.46032) -- (1.05294,4.46501) -- (1.14118,4.46275) -- (1.22941,4.47279) -- (1.31765,4.46885) -- (1.40588,4.46693) -- (1.49412,4.45977) -- (1.58235,4.46708) -- (1.67059,4.46685) -- (1.75882,4.46926) -- (1.84706,4.46956) -- (1.93529,4.47828) -- (2.02353,4.47273) -- (2.11176,4.46432) -- (2.20000,4.46401) -- (2.28824,4.47811) -- (2.37647,4.46967) -- (2.46471,4.47180) -- (2.55294,4.46823) -- (2.64118,4.48446) -- (2.72941,4.47405) -- (2.81765,4.47657) -- (2.90588,4.46917) -- (2.99412,4.47209) -- (3.08235,4.46992) -- (3.17059,4.46825) -- (3.25882,4.46532) -- (3.34706,4.46402) -- (3.43529,4.46457) -- (3.52353,4.46574) -- (3.61176,4.48107) -- (3.70000,4.46735) -- (3.78824,4.47033) -- (3.87647,4.47247) -- (3.96471,4.47523) -- (4.05294,4.46486) -- (4.14118,4.47970) -- (4.22941,4.47055) -- (4.31765,4.47451) -- (4.40588,4.47396) -- (4.49412,4.46770) -- (4.58235,4.46742) -- (4.67059,4.47139) -- (4.75882,4.46745) -- (4.84706,4.48681) -- (4.93529,4.47612) -- (5.02353,4.47295) -- (5.11176,4.47327) -- (5.20000,4.46724) -- (5.20000,4.78289) -- (5.11176,4.80286) -- (5.02353,4.78770) -- (4.93529,4.78810) -- (4.84706,4.77745) -- (4.75882,4.76873) -- (4.67059,4.80596) -- (4.58235,4.78071) -- (4.49412,4.74870) -- (4.40588,4.76460) -- (4.31765,4.77875) -- (4.22941,4.80773) -- (4.14118,4.74324) -- (4.05294,4.75637) -- (3.96471,4.77188) -- (3.87647,4.76965) -- (3.78824,4.76404) -- (3.70000,4.77606) -- (3.61176,4.78728) -- (3.52353,4.79283) -- (3.43529,4.82006) -- (3.34706,4.76978) -- (3.25882,4.80568) -- (3.17059,4.77422) -- (3.08235,4.81484) -- (2.99412,4.75213) -- (2.90588,4.79998) -- (2.81765,4.77580) -- (2.72941,4.76487) -- (2.64118,4.76789) -- (2.55294,4.78875) -- (2.46471,4.75252) -- (2.37647,4.79331) -- (2.28824,4.80865) -- (2.20000,4.78826) -- (2.11176,4.78403) -- (2.02353,4.76439) -- (1.93529,4.79226) -- (1.84706,4.78552) -- (1.75882,4.80522) -- (1.67059,4.73828) -- (1.58235,4.79880) -- (1.49412,4.76672) -- (1.40588,4.74242) -- (1.31765,4.76341) -- (1.22941,4.73969) -- (1.14118,4.75109) -- (1.05294,4.74349) -- (0.96471,4.79427) -- (0.87647,4.78317) -- (0.78824,4.78599) -- (0.70000,4.76810) -- cycle;
\fill[trT] (0.70000,4.76810) -- (0.78824,4.78599) -- (0.87647,4.78317) -- (0.96471,4.79427) -- (1.05294,4.74349) -- (1.14118,4.75109) -- (1.22941,4.73969) -- (1.31765,4.76341) -- (1.40588,4.74242) -- (1.49412,4.76672) -- (1.58235,4.79880) -- (1.67059,4.73828) -- (1.75882,4.80522) -- (1.84706,4.78552) -- (1.93529,4.79226) -- (2.02353,4.76439) -- (2.11176,4.78403) -- (2.20000,4.78826) -- (2.28824,4.80865) -- (2.37647,4.79331) -- (2.46471,4.75252) -- (2.55294,4.78875) -- (2.64118,4.76789) -- (2.72941,4.76487) -- (2.81765,4.77580) -- (2.90588,4.79998) -- (2.99412,4.75213) -- (3.08235,4.81484) -- (3.17059,4.77422) -- (3.25882,4.80568) -- (3.34706,4.76978) -- (3.43529,4.82006) -- (3.52353,4.79283) -- (3.61176,4.78728) -- (3.70000,4.77606) -- (3.78824,4.76404) -- (3.87647,4.76965) -- (3.96471,4.77188) -- (4.05294,4.75637) -- (4.14118,4.74324) -- (4.22941,4.80773) -- (4.31765,4.77875) -- (4.40588,4.76460) -- (4.49412,4.74870) -- (4.58235,4.78071) -- (4.67059,4.80596) -- (4.75882,4.76873) -- (4.84706,4.77745) -- (4.93529,4.78810) -- (5.02353,4.78770) -- (5.11176,4.80286) -- (5.20000,4.78289) -- (5.20000,4.78289) -- (5.11176,4.80292) -- (5.02353,4.78771) -- (4.93529,4.78818) -- (4.84706,4.77755) -- (4.75882,4.76873) -- (4.67059,4.80604) -- (4.58235,4.78073) -- (4.49412,4.74870) -- (4.40588,4.76467) -- (4.31765,4.77908) -- (4.22941,4.80773) -- (4.14118,4.74326) -- (4.05294,4.75637) -- (3.96471,4.77193) -- (3.87647,4.76977) -- (3.78824,4.76434) -- (3.70000,4.77616) -- (3.61176,4.78733) -- (3.52353,4.79283) -- (3.43529,4.82006) -- (3.34706,4.76980) -- (3.25882,4.80580) -- (3.17059,4.77435) -- (3.08235,4.81488) -- (2.99412,4.75213) -- (2.90588,4.79998) -- (2.81765,4.77580) -- (2.72941,4.76490) -- (2.64118,4.76805) -- (2.55294,4.78875) -- (2.46471,4.75257) -- (2.37647,4.79431) -- (2.28824,4.80865) -- (2.20000,4.78826) -- (2.11176,4.78408) -- (2.02353,4.76452) -- (1.93529,4.79232) -- (1.84706,4.78556) -- (1.75882,4.80527) -- (1.67059,4.73828) -- (1.58235,4.79880) -- (1.49412,4.76672) -- (1.40588,4.74244) -- (1.31765,4.76349) -- (1.22941,4.73982) -- (1.14118,4.75122) -- (1.05294,4.74359) -- (0.96471,4.79427) -- (0.87647,4.78335) -- (0.78824,4.78602) -- (0.70000,4.76812) -- cycle;
\fill[trM] (0.70000,4.76812) -- (0.78824,4.78602) -- (0.87647,4.78335) -- (0.96471,4.79427) -- (1.05294,4.74359) -- (1.14118,4.75122) -- (1.22941,4.73982) -- (1.31765,4.76349) -- (1.40588,4.74244) -- (1.49412,4.76672) -- (1.58235,4.79880) -- (1.67059,4.73828) -- (1.75882,4.80527) -- (1.84706,4.78556) -- (1.93529,4.79232) -- (2.02353,4.76452) -- (2.11176,4.78408) -- (2.20000,4.78826) -- (2.28824,4.80865) -- (2.37647,4.79431) -- (2.46471,4.75257) -- (2.55294,4.78875) -- (2.64118,4.76805) -- (2.72941,4.76490) -- (2.81765,4.77580) -- (2.90588,4.79998) -- (2.99412,4.75213) -- (3.08235,4.81488) -- (3.17059,4.77435) -- (3.25882,4.80580) -- (3.34706,4.76980) -- (3.43529,4.82006) -- (3.52353,4.79283) -- (3.61176,4.78733) -- (3.70000,4.77616) -- (3.78824,4.76434) -- (3.87647,4.76977) -- (3.96471,4.77193) -- (4.05294,4.75637) -- (4.14118,4.74326) -- (4.22941,4.80773) -- (4.31765,4.77908) -- (4.40588,4.76467) -- (4.49412,4.74870) -- (4.58235,4.78073) -- (4.67059,4.80604) -- (4.75882,4.76873) -- (4.84706,4.77755) -- (4.93529,4.78818) -- (5.02353,4.78771) -- (5.11176,4.80292) -- (5.20000,4.78289) -- (5.20000,5.83770) -- (5.11176,5.82069) -- (5.02353,5.82763) -- (4.93529,5.84293) -- (4.84706,5.83328) -- (4.75882,5.83138) -- (4.67059,5.84917) -- (4.58235,5.82767) -- (4.49412,5.83181) -- (4.40588,5.81582) -- (4.31765,5.83700) -- (4.22941,5.83230) -- (4.14118,5.81507) -- (4.05294,5.84161) -- (3.96471,5.84433) -- (3.87647,5.84427) -- (3.78824,5.83076) -- (3.70000,5.84455) -- (3.61176,5.82493) -- (3.52353,5.82266) -- (3.43529,5.84836) -- (3.34706,5.84208) -- (3.25882,5.84215) -- (3.17059,5.84969) -- (3.08235,5.84248) -- (2.99412,5.84027) -- (2.90588,5.83522) -- (2.81765,5.84164) -- (2.72941,5.83335) -- (2.64118,5.83611) -- (2.55294,5.82619) -- (2.46471,5.83162) -- (2.37647,5.84165) -- (2.28824,5.83120) -- (2.20000,5.85174) -- (2.11176,5.85706) -- (2.02353,5.83987) -- (1.93529,5.85323) -- (1.84706,5.84315) -- (1.75882,5.86365) -- (1.67059,5.82901) -- (1.58235,5.84543) -- (1.49412,5.85290) -- (1.40588,5.85139) -- (1.31765,5.86507) -- (1.22941,5.84647) -- (1.14118,5.82858) -- (1.05294,5.83435) -- (0.96471,5.84569) -- (0.87647,5.85119) -- (0.78824,5.84695) -- (0.70000,5.84051) -- cycle;
\fill[trU] (0.70000,5.84051) -- (0.78824,5.84695) -- (0.87647,5.85119) -- (0.96471,5.84569) -- (1.05294,5.83435) -- (1.14118,5.82858) -- (1.22941,5.84647) -- (1.31765,5.86507) -- (1.40588,5.85139) -- (1.49412,5.85290) -- (1.58235,5.84543) -- (1.67059,5.82901) -- (1.75882,5.86365) -- (1.84706,5.84315) -- (1.93529,5.85323) -- (2.02353,5.83987) -- (2.11176,5.85706) -- (2.20000,5.85174) -- (2.28824,5.83120) -- (2.37647,5.84165) -- (2.46471,5.83162) -- (2.55294,5.82619) -- (2.64118,5.83611) -- (2.72941,5.83335) -- (2.81765,5.84164) -- (2.90588,5.83522) -- (2.99412,5.84027) -- (3.08235,5.84248) -- (3.17059,5.84969) -- (3.25882,5.84215) -- (3.34706,5.84208) -- (3.43529,5.84836) -- (3.52353,5.82266) -- (3.61176,5.82493) -- (3.70000,5.84455) -- (3.78824,5.83076) -- (3.87647,5.84427) -- (3.96471,5.84433) -- (4.05294,5.84161) -- (4.14118,5.81507) -- (4.22941,5.83230) -- (4.31765,5.83700) -- (4.40588,5.81582) -- (4.49412,5.83181) -- (4.58235,5.82767) -- (4.67059,5.84917) -- (4.75882,5.83138) -- (4.84706,5.83328) -- (4.93529,5.84293) -- (5.02353,5.82763) -- (5.11176,5.82069) -- (5.20000,5.83770) -- (5.20000,5.95000) -- (5.11176,5.95000) -- (5.02353,5.95000) -- (4.93529,5.95000) -- (4.84706,5.95000) -- (4.75882,5.95000) -- (4.67059,5.95000) -- (4.58235,5.95000) -- (4.49412,5.95000) -- (4.40588,5.95000) -- (4.31765,5.95000) -- (4.22941,5.95000) -- (4.14118,5.95000) -- (4.05294,5.95000) -- (3.96471,5.95000) -- (3.87647,5.95000) -- (3.78824,5.95000) -- (3.70000,5.95000) -- (3.61176,5.95000) -- (3.52353,5.95000) -- (3.43529,5.95000) -- (3.34706,5.95000) -- (3.25882,5.95000) -- (3.17059,5.95000) -- (3.08235,5.95000) -- (2.99412,5.95000) -- (2.90588,5.95000) -- (2.81765,5.95000) -- (2.72941,5.95000) -- (2.64118,5.95000) -- (2.55294,5.95000) -- (2.46471,5.95000) -- (2.37647,5.95000) -- (2.28824,5.95000) -- (2.20000,5.95000) -- (2.11176,5.95000) -- (2.02353,5.95000) -- (1.93529,5.95000) -- (1.84706,5.95000) -- (1.75882,5.95000) -- (1.67059,5.95000) -- (1.58235,5.95000) -- (1.49412,5.95000) -- (1.40588,5.95000) -- (1.31765,5.95000) -- (1.22941,5.95000) -- (1.14118,5.95000) -- (1.05294,5.95000) -- (0.96471,5.95000) -- (0.87647,5.95000) -- (0.78824,5.95000) -- (0.70000,5.95000) -- cycle;
\node[anchor=east,inner sep=1pt,font=\rmfamily\fontsize{9}{11}\selectfont] at (0.58000,4.40000) {0};
\node[anchor=east,inner sep=1pt,font=\rmfamily\fontsize{9}{11}\selectfont] at (0.58000,5.17500) {50};
\node[anchor=east,inner sep=1pt,font=\rmfamily\fontsize{9}{11}\selectfont] at (0.58000,5.95000) {100};
\draw[traxis,line width=0.45pt] (0.70000,5.95000) -- (0.70000,4.40000) -- (5.20000,4.40000);
\draw[traxis,line width=0.4pt] (0.70000,4.40000) -- (0.70000,4.34500);
\node[anchor=center,inner sep=1pt,font=\rmfamily\fontsize{9}{11}\selectfont] at (0.70000,4.17000) {1};
\draw[traxis,line width=0.4pt] (2.02353,4.40000) -- (2.02353,4.34500);
\node[anchor=center,inner sep=1pt,font=\rmfamily\fontsize{9}{11}\selectfont] at (2.02353,4.17000) {16};
\draw[traxis,line width=0.4pt] (3.43529,4.40000) -- (3.43529,4.34500);
\node[anchor=center,inner sep=1pt,font=\rmfamily\fontsize{9}{11}\selectfont] at (3.43529,4.17000) {32};
\draw[traxis,line width=0.4pt] (5.20000,4.40000) -- (5.20000,4.34500);
\node[anchor=center,inner sep=1pt,font=\rmfamily\fontsize{9}{11}\selectfont] at (5.20000,4.17000) {52};
\node[anchor=center,inner sep=1pt,font=\rmfamily\fontsize{10}{11}\selectfont] at (8.45000,6.60000) {(b) R};
\node[anchor=center,inner sep=1pt,font=\rmfamily\fontsize{9}{11}\selectfont] at (8.45000,6.23000) {4.04\% $\to$ 4.86\%};
\fill[trwindow] (6.20000,4.40000) -- (6.55294,4.40000) -- (6.55294,5.95000) -- (6.20000,5.95000) -- cycle;
\fill[trwindow] (10.34706,4.40000) -- (10.70000,4.40000) -- (10.70000,5.95000) -- (10.34706,5.95000) -- cycle;
\draw[trgrid,line width=0.35pt] (6.20000,4.40000) -- (10.70000,4.40000);
\draw[trgrid,line width=0.35pt] (6.20000,5.02000) -- (10.70000,5.02000);
\draw[trgrid,line width=0.35pt] (6.20000,5.95000) -- (10.70000,5.95000);
\draw[trR,line width=0.7pt] (6.20000,4.75691) -- (6.28824,4.87387) -- (6.37647,4.57708) -- (6.46471,4.64270) -- (6.55294,4.83041) -- (6.64118,4.74005) -- (6.72941,5.14162) -- (6.81765,4.98407) -- (6.90588,4.90723) -- (6.99412,4.62080) -- (7.08235,4.91324) -- (7.17059,4.90398) -- (7.25882,5.00029) -- (7.34706,5.01245) -- (7.43529,5.36120) -- (7.52353,5.13901) -- (7.61176,4.80294) -- (7.70000,4.79054) -- (7.78824,5.35456) -- (7.87647,5.01690) -- (7.96471,5.10193) -- (8.05294,4.95914) -- (8.14118,5.60832) -- (8.22941,5.19209) -- (8.31765,5.29276) -- (8.40588,4.99697) -- (8.49412,5.11362) -- (8.58235,5.02691) -- (8.67059,4.95990) -- (8.75882,4.84294) -- (8.84706,4.79069) -- (8.93529,4.81297) -- (9.02353,4.85959) -- (9.11176,5.47285) -- (9.20000,4.92406) -- (9.28824,5.04336) -- (9.37647,5.12898) -- (9.46471,5.23902) -- (9.55294,4.82431) -- (9.64118,5.41816) -- (9.72941,5.05197) -- (9.81765,5.21050) -- (9.90588,5.18841) -- (9.99412,4.93815) -- (10.08235,4.92689) -- (10.17059,5.08551) -- (10.25882,4.92784) -- (10.34706,5.70222) -- (10.43529,5.27471) -- (10.52353,5.14798) -- (10.61176,5.16092) -- (10.70000,4.91953);
\node[anchor=east,inner sep=1pt,font=\rmfamily\fontsize{9}{11}\selectfont] at (6.08000,4.40000) {3.5};
\node[anchor=east,inner sep=1pt,font=\rmfamily\fontsize{9}{11}\selectfont] at (6.08000,5.02000) {4.5};
\node[anchor=east,inner sep=1pt,font=\rmfamily\fontsize{9}{11}\selectfont] at (6.08000,5.95000) {6};
\draw[traxis,line width=0.45pt] (6.20000,5.95000) -- (6.20000,4.40000) -- (10.70000,4.40000);
\draw[traxis,line width=0.4pt] (6.20000,4.40000) -- (6.20000,4.34500);
\node[anchor=center,inner sep=1pt,font=\rmfamily\fontsize{9}{11}\selectfont] at (6.20000,4.17000) {1};
\draw[traxis,line width=0.4pt] (7.52353,4.40000) -- (7.52353,4.34500);
\node[anchor=center,inner sep=1pt,font=\rmfamily\fontsize{9}{11}\selectfont] at (7.52353,4.17000) {16};
\draw[traxis,line width=0.4pt] (8.93529,4.40000) -- (8.93529,4.34500);
\node[anchor=center,inner sep=1pt,font=\rmfamily\fontsize{9}{11}\selectfont] at (8.93529,4.17000) {32};
\draw[traxis,line width=0.4pt] (10.70000,4.40000) -- (10.70000,4.34500);
\node[anchor=center,inner sep=1pt,font=\rmfamily\fontsize{9}{11}\selectfont] at (10.70000,4.17000) {52};
\node[anchor=center,inner sep=1pt,font=\rmfamily\fontsize{10}{11}\selectfont] at (13.95000,6.60000) {(c) E};
\node[anchor=center,inner sep=1pt,font=\rmfamily\fontsize{9}{11}\selectfont] at (13.95000,6.23000) {20.15\% $\to$ 20.16\%};
\fill[trwindow] (11.70000,4.40000) -- (12.05294,4.40000) -- (12.05294,5.95000) -- (11.70000,5.95000) -- cycle;
\fill[trwindow] (15.84706,4.40000) -- (16.20000,4.40000) -- (16.20000,5.95000) -- (15.84706,5.95000) -- cycle;
\draw[trgrid,line width=0.35pt] (11.70000,4.40000) -- (16.20000,4.40000);
\draw[trgrid,line width=0.35pt] (11.70000,5.17500) -- (16.20000,5.17500);
\draw[trgrid,line width=0.35pt] (11.70000,5.95000) -- (16.20000,5.95000);
\draw[trE,line width=0.7pt] (11.70000,5.11162) -- (11.78824,5.29862) -- (11.87647,5.35618) -- (11.96471,5.47437) -- (12.05294,4.78105) -- (12.14118,4.90427) -- (12.22941,4.63628) -- (12.31765,4.98202) -- (12.40588,4.74358) -- (12.49412,5.13687) -- (12.58235,5.44650) -- (12.67059,4.69283) -- (12.75882,5.49951) -- (12.84706,5.24945) -- (12.93529,5.22479) -- (13.02353,4.94585) -- (13.11176,5.29632) -- (13.20000,5.35304) -- (13.28824,5.43170) -- (13.37647,5.34548) -- (13.46471,4.80901) -- (13.55294,5.30652) -- (13.64118,4.84294) -- (13.72941,4.93527) -- (13.81765,5.04040) -- (13.90588,5.43510) -- (13.99412,4.80045) -- (14.08235,5.61146) -- (14.17059,5.12467) -- (14.25882,5.55444) -- (14.34706,5.12206) -- (14.43529,5.74357) -- (14.52353,5.38864) -- (14.61176,5.12765) -- (14.70000,5.15882) -- (14.78824,4.97132) -- (14.87647,5.01465) -- (14.96471,5.00819) -- (15.05294,4.94392) -- (15.14118,4.59424) -- (15.22941,5.51476) -- (15.31765,5.10301) -- (15.40588,4.93295) -- (15.49412,4.81245) -- (15.58235,5.21614) -- (15.67059,5.48210) -- (15.75882,5.06608) -- (15.84706,4.93312) -- (15.93529,5.19972) -- (16.02353,5.23436) -- (16.11176,5.41982) -- (16.20000,5.24559);
\node[anchor=east,inner sep=1pt,font=\rmfamily\fontsize{9}{11}\selectfont] at (11.58000,4.40000) {16};
\node[anchor=east,inner sep=1pt,font=\rmfamily\fontsize{9}{11}\selectfont] at (11.58000,5.17500) {20};
\node[anchor=east,inner sep=1pt,font=\rmfamily\fontsize{9}{11}\selectfont] at (11.58000,5.95000) {24};
\draw[traxis,line width=0.45pt] (11.70000,5.95000) -- (11.70000,4.40000) -- (16.20000,4.40000);
\draw[traxis,line width=0.4pt] (11.70000,4.40000) -- (11.70000,4.34500);
\node[anchor=center,inner sep=1pt,font=\rmfamily\fontsize{9}{11}\selectfont] at (11.70000,4.17000) {1};
\draw[traxis,line width=0.4pt] (13.02353,4.40000) -- (13.02353,4.34500);
\node[anchor=center,inner sep=1pt,font=\rmfamily\fontsize{9}{11}\selectfont] at (13.02353,4.17000) {16};
\draw[traxis,line width=0.4pt] (14.43529,4.40000) -- (14.43529,4.34500);
\node[anchor=center,inner sep=1pt,font=\rmfamily\fontsize{9}{11}\selectfont] at (14.43529,4.17000) {32};
\draw[traxis,line width=0.4pt] (16.20000,4.40000) -- (16.20000,4.34500);
\node[anchor=center,inner sep=1pt,font=\rmfamily\fontsize{9}{11}\selectfont] at (16.20000,4.17000) {52};
\node[anchor=center,inner sep=1pt,font=\rmfamily\fontsize{10}{11}\selectfont] at (2.95000,3.30000) {(d) T};
\node[anchor=center,inner sep=1pt,font=\rmfamily\fontsize{9}{11}\selectfont] at (2.95000,2.93000) {0.0041\% $\to$ 0.0033\%};
\fill[trwindow] (0.70000,1.10000) -- (1.05294,1.10000) -- (1.05294,2.65000) -- (0.70000,2.65000) -- cycle;
\fill[trwindow] (4.84706,1.10000) -- (5.20000,1.10000) -- (5.20000,2.65000) -- (4.84706,2.65000) -- cycle;
\draw[trgrid,line width=0.35pt] (0.70000,1.10000) -- (5.20000,1.10000);
\draw[trgrid,line width=0.35pt] (0.70000,1.87500) -- (5.20000,1.87500);
\draw[trgrid,line width=0.35pt] (0.70000,2.65000) -- (5.20000,2.65000);
\draw[trT,line width=0.7pt] (0.70000,1.12018) -- (0.78824,1.14242) -- (0.87647,1.35058) -- (0.96471,1.10000) -- (1.05294,1.24279) -- (1.14118,1.27563) -- (1.22941,1.28509) -- (1.31765,1.20879) -- (1.40588,1.13787) -- (1.49412,1.10000) -- (1.58235,1.10000) -- (1.67059,1.10000) -- (1.75882,1.17600) -- (1.84706,1.16795) -- (1.93529,1.18770) -- (2.02353,1.27447) -- (2.11176,1.17211) -- (2.20000,1.10000) -- (2.28824,1.10000) -- (2.37647,2.52849) -- (2.46471,1.17749) -- (2.55294,1.10000) -- (2.64118,1.32912) -- (2.72941,1.14412) -- (2.81765,1.10000) -- (2.90588,1.10000) -- (2.99412,1.10000) -- (3.08235,1.15528) -- (3.17059,1.28818) -- (3.25882,1.27062) -- (3.34706,1.13016) -- (3.43529,1.10000) -- (3.52353,1.10000) -- (3.61176,1.16400) -- (3.70000,1.24901) -- (3.78824,1.52521) -- (3.87647,1.27564) -- (3.96471,1.16873) -- (4.05294,1.10000) -- (4.14118,1.11944) -- (4.22941,1.10000) -- (4.31765,1.56323) -- (4.40588,1.20989) -- (4.49412,1.10000) -- (4.58235,1.11874) -- (4.67059,1.22121) -- (4.75882,1.10000) -- (4.84706,1.23935) -- (4.93529,1.22442) -- (5.02353,1.11733) -- (5.11176,1.18799) -- (5.20000,1.10000);
\node[anchor=east,inner sep=1pt,font=\rmfamily\fontsize{9}{11}\selectfont] at (0.58000,1.10000) {0};
\node[anchor=east,inner sep=1pt,font=\rmfamily\fontsize{9}{11}\selectfont] at (0.58000,1.87500) {0.035};
\node[anchor=east,inner sep=1pt,font=\rmfamily\fontsize{9}{11}\selectfont] at (0.58000,2.65000) {0.07};
\draw[traxis,line width=0.45pt] (0.70000,2.65000) -- (0.70000,1.10000) -- (5.20000,1.10000);
\draw[traxis,line width=0.4pt] (0.70000,1.10000) -- (0.70000,1.04500);
\node[anchor=center,inner sep=1pt,font=\rmfamily\fontsize{9}{11}\selectfont] at (0.70000,0.87000) {1};
\draw[traxis,line width=0.4pt] (2.02353,1.10000) -- (2.02353,1.04500);
\node[anchor=center,inner sep=1pt,font=\rmfamily\fontsize{9}{11}\selectfont] at (2.02353,0.87000) {16};
\draw[traxis,line width=0.4pt] (3.43529,1.10000) -- (3.43529,1.04500);
\node[anchor=center,inner sep=1pt,font=\rmfamily\fontsize{9}{11}\selectfont] at (3.43529,0.87000) {32};
\draw[traxis,line width=0.4pt] (5.20000,1.10000) -- (5.20000,1.04500);
\node[anchor=center,inner sep=1pt,font=\rmfamily\fontsize{9}{11}\selectfont] at (5.20000,0.87000) {52};
\node[anchor=center,inner sep=1pt,font=\rmfamily\fontsize{10}{11}\selectfont] at (8.45000,3.30000) {(e) Multilabel};
\node[anchor=center,inner sep=1pt,font=\rmfamily\fontsize{9}{11}\selectfont] at (8.45000,2.93000) {68.95\% $\to$ 67.39\%};
\fill[trwindow] (6.20000,1.10000) -- (6.55294,1.10000) -- (6.55294,2.65000) -- (6.20000,2.65000) -- cycle;
\fill[trwindow] (10.34706,1.10000) -- (10.70000,1.10000) -- (10.70000,2.65000) -- (10.34706,2.65000) -- cycle;
\draw[trgrid,line width=0.35pt] (6.20000,1.10000) -- (10.70000,1.10000);
\draw[trgrid,line width=0.35pt] (6.20000,1.78889) -- (10.70000,1.78889);
\draw[trgrid,line width=0.35pt] (6.20000,2.65000) -- (10.70000,2.65000);
\draw[trM,line width=0.7pt] (6.20000,1.99324) -- (6.28824,1.86589) -- (6.37647,1.94269) -- (6.46471,1.76031) -- (6.55294,2.19734) -- (6.64118,2.04848) -- (6.72941,2.37384) -- (6.81765,2.31761) -- (6.90588,2.39943) -- (6.99412,2.14647) -- (7.08235,1.70695) -- (7.17059,2.19703) -- (7.25882,1.83751) -- (7.34706,1.82874) -- (7.43529,1.86561) -- (7.52353,2.02613) -- (7.61176,1.99974) -- (7.70000,1.89421) -- (7.78824,1.43949) -- (7.87647,1.71484) -- (7.96471,2.06714) -- (8.05294,1.60485) -- (8.14118,1.94509) -- (8.22941,1.94944) -- (8.31765,1.92044) -- (8.40588,1.58042) -- (8.49412,2.16825) -- (8.58235,1.49555) -- (8.67059,2.02593) -- (8.75882,1.59276) -- (8.84706,1.99195) -- (8.93529,1.50328) -- (9.02353,1.52036) -- (9.11176,1.60669) -- (9.20000,1.94877) -- (9.28824,1.92691) -- (9.37647,2.01669) -- (9.46471,1.99338) -- (9.55294,2.13595) -- (9.64118,1.98686) -- (9.72941,1.46188) -- (9.81765,1.83247) -- (9.90588,1.75724) -- (9.99412,2.11229) -- (10.08235,1.71053) -- (10.17059,1.66809) -- (10.25882,1.88492) -- (10.34706,1.80809) -- (10.43529,1.79716) -- (10.52353,1.63248) -- (10.61176,1.38630) -- (10.70000,1.79795);
\node[anchor=east,inner sep=1pt,font=\rmfamily\fontsize{9}{11}\selectfont] at (6.08000,1.10000) {64};
\node[anchor=east,inner sep=1pt,font=\rmfamily\fontsize{9}{11}\selectfont] at (6.08000,1.78889) {68};
\node[anchor=east,inner sep=1pt,font=\rmfamily\fontsize{9}{11}\selectfont] at (6.08000,2.65000) {73};
\draw[traxis,line width=0.45pt] (6.20000,2.65000) -- (6.20000,1.10000) -- (10.70000,1.10000);
\draw[traxis,line width=0.4pt] (6.20000,1.10000) -- (6.20000,1.04500);
\node[anchor=center,inner sep=1pt,font=\rmfamily\fontsize{9}{11}\selectfont] at (6.20000,0.87000) {1};
\draw[traxis,line width=0.4pt] (7.52353,1.10000) -- (7.52353,1.04500);
\node[anchor=center,inner sep=1pt,font=\rmfamily\fontsize{9}{11}\selectfont] at (7.52353,0.87000) {16};
\draw[traxis,line width=0.4pt] (8.93529,1.10000) -- (8.93529,1.04500);
\node[anchor=center,inner sep=1pt,font=\rmfamily\fontsize{9}{11}\selectfont] at (8.93529,0.87000) {32};
\draw[traxis,line width=0.4pt] (10.70000,1.10000) -- (10.70000,1.04500);
\node[anchor=center,inner sep=1pt,font=\rmfamily\fontsize{9}{11}\selectfont] at (10.70000,0.87000) {52};
\node[anchor=center,inner sep=1pt,font=\rmfamily\fontsize{10}{11}\selectfont] at (13.95000,3.30000) {(f) Other / protected};
\node[anchor=center,inner sep=1pt,font=\rmfamily\fontsize{9}{11}\selectfont] at (13.95000,2.93000) {6.86\% $\to$ 7.58\%};
\fill[trwindow] (11.70000,1.10000) -- (12.05294,1.10000) -- (12.05294,2.65000) -- (11.70000,2.65000) -- cycle;
\fill[trwindow] (15.84706,1.10000) -- (16.20000,1.10000) -- (16.20000,2.65000) -- (15.84706,2.65000) -- cycle;
\draw[trgrid,line width=0.35pt] (11.70000,1.10000) -- (16.20000,1.10000);
\draw[trgrid,line width=0.35pt] (11.70000,1.87500) -- (16.20000,1.87500);
\draw[trgrid,line width=0.35pt] (11.70000,2.65000) -- (16.20000,2.65000);
\draw[trU,line width=0.7pt] (11.70000,1.73984) -- (11.78824,1.61108) -- (11.87647,1.52622) -- (11.96471,1.63610) -- (12.05294,1.86291) -- (12.14118,1.97843) -- (12.22941,1.62063) -- (12.31765,1.24851) -- (12.40588,1.52215) -- (12.49412,1.49196) -- (12.58235,1.64148) -- (12.67059,1.96983) -- (12.75882,1.27705) -- (12.84706,1.68697) -- (12.93529,1.48540) -- (13.02353,1.75266) -- (13.11176,1.40888) -- (13.20000,1.51529) -- (13.28824,1.92592) -- (13.37647,1.71707) -- (13.46471,1.91769) -- (13.55294,2.02626) -- (13.64118,1.82777) -- (13.72941,1.88292) -- (13.81765,1.71718) -- (13.90588,1.84560) -- (13.99412,1.74464) -- (14.08235,1.70044) -- (14.17059,1.55626) -- (14.25882,1.70707) -- (14.34706,1.70843) -- (14.43529,1.58290) -- (14.52353,2.09673) -- (14.61176,2.05141) -- (14.70000,1.65898) -- (14.78824,1.93481) -- (14.87647,1.66456) -- (14.96471,1.66335) -- (15.05294,1.71787) -- (15.14118,2.24852) -- (15.22941,1.90401) -- (15.31765,1.81000) -- (15.40588,2.23352) -- (15.49412,1.91389) -- (15.58235,1.99652) -- (15.67059,1.56662) -- (15.75882,1.92250) -- (15.84706,1.88439) -- (15.93529,1.69146) -- (16.02353,1.99731) -- (16.11176,2.13625) -- (16.20000,1.79597);
\node[anchor=east,inner sep=1pt,font=\rmfamily\fontsize{9}{11}\selectfont] at (11.58000,1.10000) {5};
\node[anchor=east,inner sep=1pt,font=\rmfamily\fontsize{9}{11}\selectfont] at (11.58000,1.87500) {7.5};
\node[anchor=east,inner sep=1pt,font=\rmfamily\fontsize{9}{11}\selectfont] at (11.58000,2.65000) {10};
\draw[traxis,line width=0.45pt] (11.70000,2.65000) -- (11.70000,1.10000) -- (16.20000,1.10000);
\draw[traxis,line width=0.4pt] (11.70000,1.10000) -- (11.70000,1.04500);
\node[anchor=center,inner sep=1pt,font=\rmfamily\fontsize{9}{11}\selectfont] at (11.70000,0.87000) {1};
\draw[traxis,line width=0.4pt] (13.02353,1.10000) -- (13.02353,1.04500);
\node[anchor=center,inner sep=1pt,font=\rmfamily\fontsize{9}{11}\selectfont] at (13.02353,0.87000) {16};
\draw[traxis,line width=0.4pt] (14.43529,1.10000) -- (14.43529,1.04500);
\node[anchor=center,inner sep=1pt,font=\rmfamily\fontsize{9}{11}\selectfont] at (14.43529,0.87000) {32};
\draw[traxis,line width=0.4pt] (16.20000,1.10000) -- (16.20000,1.04500);
\node[anchor=center,inner sep=1pt,font=\rmfamily\fontsize{9}{11}\selectfont] at (16.20000,0.87000) {52};
\node[anchor=center,inner sep=1pt,font=\rmfamily\fontsize{9}{11}\selectfont] at (8.45000,0.35000) {Training update (all 52 observations)};
\end{tikzpicture}}
\caption{Token phase-label shares during default PCRS-T training. (a) Full composition; (b--f) individual shares with separate, explicitly labeled y-axis ranges (all in percent). Every curve retains all 52 unsmoothed observations. Annotations report the early (updates 1--5) and late (48--52) five-update means; gray windows mark the corresponding updates. R/E/T denote single-label buckets; M pools RE/RT/ET/RET; other/protected includes unclassified and protected tokens. Shares are computed over all valid response tokens before reward-eligibility filtering on changing batches. Exact values appear in Appendix~\ref{app:training_role_history}.}
\label{fig:trainingroles}
\end{minipage}
\par\vspace{14pt}
\begin{minipage}{\textwidth}
\captionsetup{type=figure}

\centering\resizebox{\textwidth}{!}{%
\begin{tikzpicture}[x=1cm,y=1cm,font=\rmfamily\fontsize{9}{11}\selectfont,line cap=round,line join=round]
\definecolor{Bcol}{HTML}{737C87}
\definecolor{Tcol}{HTML}{AA6187}
\definecolor{Scol}{HTML}{397DB5}
\definecolor{Ecol}{HTML}{B78429}
\definecolor{Rcol}{HTML}{397DB5}
\definecolor{Mcol}{HTML}{788D3F}
\definecolor{Ucol}{HTML}{737C87}
\node[anchor=west,font=\rmfamily\bfseries\fontsize{9}{11}\selectfont] at (0.15000,4.08000) {Reward setting};
\node[font=\rmfamily\bfseries\fontsize{9}{11}\selectfont] at (5.37500,4.08000) {R};
\node[font=\rmfamily\bfseries\fontsize{9}{11}\selectfont] at (7.52500,4.08000) {E};
\node[font=\rmfamily\bfseries\fontsize{9}{11}\selectfont] at (9.67500,4.08000) {T};
\node[font=\rmfamily\bfseries\fontsize{9}{11}\selectfont] at (11.82500,4.08000) {M};
\node[font=\rmfamily\bfseries\fontsize{9}{11}\selectfont] at (13.97500,4.08000) {U};
\node[anchor=west] at (0.15000,3.54000) {Default PCRS-T};
\fill[Scol!21.17] (4.31500,3.20000) -- (6.43500,3.20000) -- (6.43500,3.88000) -- (4.31500,3.88000) -- cycle;
\node[font=\rmfamily\fontsize{9}{11}\selectfont] at (5.37500,3.54000) {+0.814};
\fill[Scol!0.29] (6.46500,3.20000) -- (8.58500,3.20000) -- (8.58500,3.88000) -- (6.46500,3.88000) -- cycle;
\node[font=\rmfamily\fontsize{9}{11}\selectfont] at (7.52500,3.54000) {+0.011};
\fill[Ecol!0.02] (8.61500,3.20000) -- (10.73500,3.20000) -- (10.73500,3.88000) -- (8.61500,3.88000) -- cycle;
\node[font=\rmfamily\fontsize{9}{11}\selectfont] at (9.67500,3.54000) {-0.000785};
\fill[Ecol!40.38] (10.76500,3.20000) -- (12.88500,3.20000) -- (12.88500,3.88000) -- (10.76500,3.88000) -- cycle;
\node[font=\rmfamily\fontsize{9}{11}\selectfont] at (11.82500,3.54000) {-1.553};
\fill[Scol!18.94] (12.91500,3.20000) -- (15.03500,3.20000) -- (15.03500,3.88000) -- (12.91500,3.88000) -- cycle;
\node[font=\rmfamily\fontsize{9}{11}\selectfont] at (13.97500,3.54000) {+0.729};
\node[anchor=west] at (0.15000,2.69000) {E-priority};
\fill[Scol!9.05] (4.31500,2.35000) -- (6.43500,2.35000) -- (6.43500,3.03000) -- (4.31500,3.03000) -- cycle;
\node[font=\rmfamily\fontsize{9}{11}\selectfont] at (5.37500,2.69000) {+0.348};
\fill[Ecol!31.37] (6.46500,2.35000) -- (8.58500,2.35000) -- (8.58500,3.03000) -- (6.46500,3.03000) -- cycle;
\node[font=\rmfamily\fontsize{9}{11}\selectfont] at (7.52500,2.69000) {-1.206};
\fill[Scol!0.1] (8.61500,2.35000) -- (10.73500,2.35000) -- (10.73500,3.03000) -- (8.61500,3.03000) -- cycle;
\node[font=\rmfamily\fontsize{9}{11}\selectfont] at (9.67500,2.69000) {+0.003930};
\fill[Scol!20.96] (10.76500,2.35000) -- (12.88500,2.35000) -- (12.88500,3.03000) -- (10.76500,3.03000) -- cycle;
\node[font=\rmfamily\fontsize{9}{11}\selectfont] at (11.82500,2.69000) {+0.806};
\fill[Scol!1.26] (12.91500,2.35000) -- (15.03500,2.35000) -- (15.03500,3.03000) -- (12.91500,3.03000) -- cycle;
\node[font=\rmfamily\fontsize{9}{11}\selectfont] at (13.97500,2.69000) {+0.049};
\node[anchor=west] at (0.15000,1.84000) {T-priority};
\fill[Scol!13.25] (4.31500,1.50000) -- (6.43500,1.50000) -- (6.43500,2.18000) -- (4.31500,2.18000) -- cycle;
\node[font=\rmfamily\fontsize{9}{11}\selectfont] at (5.37500,1.84000) {+0.510};
\fill[Ecol!20.23] (6.46500,1.50000) -- (8.58500,1.50000) -- (8.58500,2.18000) -- (6.46500,2.18000) -- cycle;
\node[font=\rmfamily\fontsize{9}{11}\selectfont] at (7.52500,1.84000) {-0.778};
\fill[Ecol!0.04] (8.61500,1.50000) -- (10.73500,1.50000) -- (10.73500,2.18000) -- (8.61500,2.18000) -- cycle;
\node[font=\rmfamily\fontsize{9}{11}\selectfont] at (9.67500,1.84000) {-0.001354};
\fill[Ecol!4.42] (10.76500,1.50000) -- (12.88500,1.50000) -- (12.88500,2.18000) -- (10.76500,2.18000) -- cycle;
\node[font=\rmfamily\fontsize{9}{11}\selectfont] at (11.82500,1.84000) {-0.170};
\fill[Scol!11.43] (12.91500,1.50000) -- (15.03500,1.50000) -- (15.03500,2.18000) -- (12.91500,2.18000) -- cycle;
\node[font=\rmfamily\fontsize{9}{11}\selectfont] at (13.97500,1.84000) {+0.440};
\node[anchor=west] at (0.15000,0.99000) {R-milder};
\fill[Scol!26.77] (4.31500,0.65000) -- (6.43500,0.65000) -- (6.43500,1.33000) -- (4.31500,1.33000) -- cycle;
\node[font=\rmfamily\fontsize{9}{11}\selectfont] at (5.37500,0.99000) {+1.030};
\fill[Ecol!57.77] (6.46500,0.65000) -- (8.58500,0.65000) -- (8.58500,1.33000) -- (6.46500,1.33000) -- cycle;
\node[font=\rmfamily\fontsize{9}{11}\selectfont] at (7.52500,0.99000) {-2.222};
\fill[Scol!0.04] (8.61500,0.65000) -- (10.73500,0.65000) -- (10.73500,1.33000) -- (8.61500,1.33000) -- cycle;
\node[font=\rmfamily\fontsize{9}{11}\selectfont] at (9.67500,0.99000) {+0.001441};
\fill[Scol!13.62] (10.76500,0.65000) -- (12.88500,0.65000) -- (12.88500,1.33000) -- (10.76500,1.33000) -- cycle;
\node[font=\rmfamily\fontsize{9}{11}\selectfont] at (11.82500,0.99000) {+0.524};
\fill[Scol!17.34] (12.91500,0.65000) -- (15.03500,0.65000) -- (15.03500,1.33000) -- (12.91500,1.33000) -- cycle;
\node[font=\rmfamily\fontsize{9}{11}\selectfont] at (13.97500,0.99000) {+0.667};
\node[anchor=west,font=\rmfamily\fontsize{9}{11}\selectfont] at (0.15000,0.25000) {Late $-$ early (percentage points)};
\fill[Ecol!64.35] (6.25000,0.13000) -- (6.31600,0.13000) -- (6.31600,0.35000) -- (6.25000,0.35000) -- cycle;
\fill[Ecol!63.05] (6.31500,0.13000) -- (6.38100,0.13000) -- (6.38100,0.35000) -- (6.31500,0.35000) -- cycle;
\fill[Ecol!61.75] (6.38000,0.13000) -- (6.44600,0.13000) -- (6.44600,0.35000) -- (6.38000,0.35000) -- cycle;
\fill[Ecol!60.45] (6.44500,0.13000) -- (6.51100,0.13000) -- (6.51100,0.35000) -- (6.44500,0.35000) -- cycle;
\fill[Ecol!59.14999999999999] (6.51000,0.13000) -- (6.57600,0.13000) -- (6.57600,0.35000) -- (6.51000,0.35000) -- cycle;
\fill[Ecol!57.85] (6.57500,0.13000) -- (6.64100,0.13000) -- (6.64100,0.35000) -- (6.57500,0.35000) -- cycle;
\fill[Ecol!56.54999999999999] (6.64000,0.13000) -- (6.70600,0.13000) -- (6.70600,0.35000) -- (6.64000,0.35000) -- cycle;
\fill[Ecol!55.25] (6.70500,0.13000) -- (6.77100,0.13000) -- (6.77100,0.35000) -- (6.70500,0.35000) -- cycle;
\fill[Ecol!53.95] (6.77000,0.13000) -- (6.83600,0.13000) -- (6.83600,0.35000) -- (6.77000,0.35000) -- cycle;
\fill[Ecol!52.65] (6.83500,0.13000) -- (6.90100,0.13000) -- (6.90100,0.35000) -- (6.83500,0.35000) -- cycle;
\fill[Ecol!51.35] (6.90000,0.13000) -- (6.96600,0.13000) -- (6.96600,0.35000) -- (6.90000,0.35000) -- cycle;
\fill[Ecol!50.050000000000004] (6.96500,0.13000) -- (7.03100,0.13000) -- (7.03100,0.35000) -- (6.96500,0.35000) -- cycle;
\fill[Ecol!48.75] (7.03000,0.13000) -- (7.09600,0.13000) -- (7.09600,0.35000) -- (7.03000,0.35000) -- cycle;
\fill[Ecol!47.449999999999996] (7.09500,0.13000) -- (7.16100,0.13000) -- (7.16100,0.35000) -- (7.09500,0.35000) -- cycle;
\fill[Ecol!46.15] (7.16000,0.13000) -- (7.22600,0.13000) -- (7.22600,0.35000) -- (7.16000,0.35000) -- cycle;
\fill[Ecol!44.85] (7.22500,0.13000) -- (7.29100,0.13000) -- (7.29100,0.35000) -- (7.22500,0.35000) -- cycle;
\fill[Ecol!43.550000000000004] (7.29000,0.13000) -- (7.35600,0.13000) -- (7.35600,0.35000) -- (7.29000,0.35000) -- cycle;
\fill[Ecol!42.25] (7.35500,0.13000) -- (7.42100,0.13000) -- (7.42100,0.35000) -- (7.35500,0.35000) -- cycle;
\fill[Ecol!40.95] (7.42000,0.13000) -- (7.48600,0.13000) -- (7.48600,0.35000) -- (7.42000,0.35000) -- cycle;
\fill[Ecol!39.65] (7.48500,0.13000) -- (7.55100,0.13000) -- (7.55100,0.35000) -- (7.48500,0.35000) -- cycle;
\fill[Ecol!38.35000000000001] (7.55000,0.13000) -- (7.61600,0.13000) -- (7.61600,0.35000) -- (7.55000,0.35000) -- cycle;
\fill[Ecol!37.050000000000004] (7.61500,0.13000) -- (7.68100,0.13000) -- (7.68100,0.35000) -- (7.61500,0.35000) -- cycle;
\fill[Ecol!35.75] (7.68000,0.13000) -- (7.74600,0.13000) -- (7.74600,0.35000) -- (7.68000,0.35000) -- cycle;
\fill[Ecol!34.45] (7.74500,0.13000) -- (7.81100,0.13000) -- (7.81100,0.35000) -- (7.74500,0.35000) -- cycle;
\fill[Ecol!33.15] (7.81000,0.13000) -- (7.87600,0.13000) -- (7.87600,0.35000) -- (7.81000,0.35000) -- cycle;
\fill[Ecol!31.85] (7.87500,0.13000) -- (7.94100,0.13000) -- (7.94100,0.35000) -- (7.87500,0.35000) -- cycle;
\fill[Ecol!30.55] (7.94000,0.13000) -- (8.00600,0.13000) -- (8.00600,0.35000) -- (7.94000,0.35000) -- cycle;
\fill[Ecol!29.25] (8.00500,0.13000) -- (8.07100,0.13000) -- (8.07100,0.35000) -- (8.00500,0.35000) -- cycle;
\fill[Ecol!27.95] (8.07000,0.13000) -- (8.13600,0.13000) -- (8.13600,0.35000) -- (8.07000,0.35000) -- cycle;
\fill[Ecol!26.65] (8.13500,0.13000) -- (8.20100,0.13000) -- (8.20100,0.35000) -- (8.13500,0.35000) -- cycle;
\fill[Ecol!25.35] (8.20000,0.13000) -- (8.26600,0.13000) -- (8.26600,0.35000) -- (8.20000,0.35000) -- cycle;
\fill[Ecol!24.05] (8.26500,0.13000) -- (8.33100,0.13000) -- (8.33100,0.35000) -- (8.26500,0.35000) -- cycle;
\fill[Ecol!22.75] (8.33000,0.13000) -- (8.39600,0.13000) -- (8.39600,0.35000) -- (8.33000,0.35000) -- cycle;
\fill[Ecol!21.449999999999996] (8.39500,0.13000) -- (8.46100,0.13000) -- (8.46100,0.35000) -- (8.39500,0.35000) -- cycle;
\fill[Ecol!20.149999999999995] (8.46000,0.13000) -- (8.52600,0.13000) -- (8.52600,0.35000) -- (8.46000,0.35000) -- cycle;
\fill[Ecol!18.85] (8.52500,0.13000) -- (8.59100,0.13000) -- (8.59100,0.35000) -- (8.52500,0.35000) -- cycle;
\fill[Ecol!17.55] (8.59000,0.13000) -- (8.65600,0.13000) -- (8.65600,0.35000) -- (8.59000,0.35000) -- cycle;
\fill[Ecol!16.25] (8.65500,0.13000) -- (8.72100,0.13000) -- (8.72100,0.35000) -- (8.65500,0.35000) -- cycle;
\fill[Ecol!14.95] (8.72000,0.13000) -- (8.78600,0.13000) -- (8.78600,0.35000) -- (8.72000,0.35000) -- cycle;
\fill[Ecol!13.649999999999999] (8.78500,0.13000) -- (8.85100,0.13000) -- (8.85100,0.35000) -- (8.78500,0.35000) -- cycle;
\fill[Ecol!12.350000000000001] (8.85000,0.13000) -- (8.91600,0.13000) -- (8.91600,0.35000) -- (8.85000,0.35000) -- cycle;
\fill[Ecol!11.049999999999995] (8.91500,0.13000) -- (8.98100,0.13000) -- (8.98100,0.35000) -- (8.91500,0.35000) -- cycle;
\fill[Ecol!9.75] (8.98000,0.13000) -- (9.04600,0.13000) -- (9.04600,0.35000) -- (8.98000,0.35000) -- cycle;
\fill[Ecol!8.450000000000005] (9.04500,0.13000) -- (9.11100,0.13000) -- (9.11100,0.35000) -- (9.04500,0.35000) -- cycle;
\fill[Ecol!7.149999999999998] (9.11000,0.13000) -- (9.17600,0.13000) -- (9.17600,0.35000) -- (9.11000,0.35000) -- cycle;
\fill[Ecol!5.850000000000002] (9.17500,0.13000) -- (9.24100,0.13000) -- (9.24100,0.35000) -- (9.17500,0.35000) -- cycle;
\fill[Ecol!4.549999999999995] (9.24000,0.13000) -- (9.30600,0.13000) -- (9.30600,0.35000) -- (9.24000,0.35000) -- cycle;
\fill[Ecol!3.25] (9.30500,0.13000) -- (9.37100,0.13000) -- (9.37100,0.35000) -- (9.30500,0.35000) -- cycle;
\fill[Ecol!1.9500000000000046] (9.37000,0.13000) -- (9.43600,0.13000) -- (9.43600,0.35000) -- (9.37000,0.35000) -- cycle;
\fill[Ecol!0.6499999999999977] (9.43500,0.13000) -- (9.50100,0.13000) -- (9.50100,0.35000) -- (9.43500,0.35000) -- cycle;
\fill[Scol!0.6499999999999977] (9.50000,0.13000) -- (9.56600,0.13000) -- (9.56600,0.35000) -- (9.50000,0.35000) -- cycle;
\fill[Scol!1.9500000000000046] (9.56500,0.13000) -- (9.63100,0.13000) -- (9.63100,0.35000) -- (9.56500,0.35000) -- cycle;
\fill[Scol!3.25] (9.63000,0.13000) -- (9.69600,0.13000) -- (9.69600,0.35000) -- (9.63000,0.35000) -- cycle;
\fill[Scol!4.549999999999995] (9.69500,0.13000) -- (9.76100,0.13000) -- (9.76100,0.35000) -- (9.69500,0.35000) -- cycle;
\fill[Scol!5.850000000000002] (9.76000,0.13000) -- (9.82600,0.13000) -- (9.82600,0.35000) -- (9.76000,0.35000) -- cycle;
\fill[Scol!7.149999999999998] (9.82500,0.13000) -- (9.89100,0.13000) -- (9.89100,0.35000) -- (9.82500,0.35000) -- cycle;
\fill[Scol!8.450000000000005] (9.89000,0.13000) -- (9.95600,0.13000) -- (9.95600,0.35000) -- (9.89000,0.35000) -- cycle;
\fill[Scol!9.75] (9.95500,0.13000) -- (10.02100,0.13000) -- (10.02100,0.35000) -- (9.95500,0.35000) -- cycle;
\fill[Scol!11.049999999999995] (10.02000,0.13000) -- (10.08600,0.13000) -- (10.08600,0.35000) -- (10.02000,0.35000) -- cycle;
\fill[Scol!12.350000000000001] (10.08500,0.13000) -- (10.15100,0.13000) -- (10.15100,0.35000) -- (10.08500,0.35000) -- cycle;
\fill[Scol!13.649999999999999] (10.15000,0.13000) -- (10.21600,0.13000) -- (10.21600,0.35000) -- (10.15000,0.35000) -- cycle;
\fill[Scol!14.950000000000005] (10.21500,0.13000) -- (10.28100,0.13000) -- (10.28100,0.35000) -- (10.21500,0.35000) -- cycle;
\fill[Scol!16.25] (10.28000,0.13000) -- (10.34600,0.13000) -- (10.34600,0.35000) -- (10.28000,0.35000) -- cycle;
\fill[Scol!17.549999999999994] (10.34500,0.13000) -- (10.41100,0.13000) -- (10.41100,0.35000) -- (10.34500,0.35000) -- cycle;
\fill[Scol!18.85] (10.41000,0.13000) -- (10.47600,0.13000) -- (10.47600,0.35000) -- (10.41000,0.35000) -- cycle;
\fill[Scol!20.149999999999995] (10.47500,0.13000) -- (10.54100,0.13000) -- (10.54100,0.35000) -- (10.47500,0.35000) -- cycle;
\fill[Scol!21.450000000000003] (10.54000,0.13000) -- (10.60600,0.13000) -- (10.60600,0.35000) -- (10.54000,0.35000) -- cycle;
\fill[Scol!22.75] (10.60500,0.13000) -- (10.67100,0.13000) -- (10.67100,0.35000) -- (10.60500,0.35000) -- cycle;
\fill[Scol!24.049999999999997] (10.67000,0.13000) -- (10.73600,0.13000) -- (10.73600,0.35000) -- (10.67000,0.35000) -- cycle;
\fill[Scol!25.35] (10.73500,0.13000) -- (10.80100,0.13000) -- (10.80100,0.35000) -- (10.73500,0.35000) -- cycle;
\fill[Scol!26.65] (10.80000,0.13000) -- (10.86600,0.13000) -- (10.86600,0.35000) -- (10.80000,0.35000) -- cycle;
\fill[Scol!27.950000000000003] (10.86500,0.13000) -- (10.93100,0.13000) -- (10.93100,0.35000) -- (10.86500,0.35000) -- cycle;
\fill[Scol!29.25] (10.93000,0.13000) -- (10.99600,0.13000) -- (10.99600,0.35000) -- (10.93000,0.35000) -- cycle;
\fill[Scol!30.549999999999994] (10.99500,0.13000) -- (11.06100,0.13000) -- (11.06100,0.35000) -- (10.99500,0.35000) -- cycle;
\fill[Scol!31.85] (11.06000,0.13000) -- (11.12600,0.13000) -- (11.12600,0.35000) -- (11.06000,0.35000) -- cycle;
\fill[Scol!33.15] (11.12500,0.13000) -- (11.19100,0.13000) -- (11.19100,0.35000) -- (11.12500,0.35000) -- cycle;
\fill[Scol!34.45] (11.19000,0.13000) -- (11.25600,0.13000) -- (11.25600,0.35000) -- (11.19000,0.35000) -- cycle;
\fill[Scol!35.75] (11.25500,0.13000) -- (11.32100,0.13000) -- (11.32100,0.35000) -- (11.25500,0.35000) -- cycle;
\fill[Scol!37.05] (11.32000,0.13000) -- (11.38600,0.13000) -- (11.38600,0.35000) -- (11.32000,0.35000) -- cycle;
\fill[Scol!38.35000000000001] (11.38500,0.13000) -- (11.45100,0.13000) -- (11.45100,0.35000) -- (11.38500,0.35000) -- cycle;
\fill[Scol!39.650000000000006] (11.45000,0.13000) -- (11.51600,0.13000) -- (11.51600,0.35000) -- (11.45000,0.35000) -- cycle;
\fill[Scol!40.95000000000001] (11.51500,0.13000) -- (11.58100,0.13000) -- (11.58100,0.35000) -- (11.51500,0.35000) -- cycle;
\fill[Scol!42.25] (11.58000,0.13000) -- (11.64600,0.13000) -- (11.64600,0.35000) -- (11.58000,0.35000) -- cycle;
\fill[Scol!43.55] (11.64500,0.13000) -- (11.71100,0.13000) -- (11.71100,0.35000) -- (11.64500,0.35000) -- cycle;
\fill[Scol!44.84999999999999] (11.71000,0.13000) -- (11.77600,0.13000) -- (11.77600,0.35000) -- (11.71000,0.35000) -- cycle;
\fill[Scol!46.15000000000001] (11.77500,0.13000) -- (11.84100,0.13000) -- (11.84100,0.35000) -- (11.77500,0.35000) -- cycle;
\fill[Scol!47.45] (11.84000,0.13000) -- (11.90600,0.13000) -- (11.90600,0.35000) -- (11.84000,0.35000) -- cycle;
\fill[Scol!48.75] (11.90500,0.13000) -- (11.97100,0.13000) -- (11.97100,0.35000) -- (11.90500,0.35000) -- cycle;
\fill[Scol!50.05] (11.97000,0.13000) -- (12.03600,0.13000) -- (12.03600,0.35000) -- (11.97000,0.35000) -- cycle;
\fill[Scol!51.34999999999999] (12.03500,0.13000) -- (12.10100,0.13000) -- (12.10100,0.35000) -- (12.03500,0.35000) -- cycle;
\fill[Scol!52.65000000000001] (12.10000,0.13000) -- (12.16600,0.13000) -- (12.16600,0.35000) -- (12.10000,0.35000) -- cycle;
\fill[Scol!53.95] (12.16500,0.13000) -- (12.23100,0.13000) -- (12.23100,0.35000) -- (12.16500,0.35000) -- cycle;
\fill[Scol!55.25] (12.23000,0.13000) -- (12.29600,0.13000) -- (12.29600,0.35000) -- (12.23000,0.35000) -- cycle;
\fill[Scol!56.54999999999999] (12.29500,0.13000) -- (12.36100,0.13000) -- (12.36100,0.35000) -- (12.29500,0.35000) -- cycle;
\fill[Scol!57.849999999999994] (12.36000,0.13000) -- (12.42600,0.13000) -- (12.42600,0.35000) -- (12.36000,0.35000) -- cycle;
\fill[Scol!59.150000000000006] (12.42500,0.13000) -- (12.49100,0.13000) -- (12.49100,0.35000) -- (12.42500,0.35000) -- cycle;
\fill[Scol!60.45] (12.49000,0.13000) -- (12.55600,0.13000) -- (12.55600,0.35000) -- (12.49000,0.35000) -- cycle;
\fill[Scol!61.75] (12.55500,0.13000) -- (12.62100,0.13000) -- (12.62100,0.35000) -- (12.55500,0.35000) -- cycle;
\fill[Scol!63.05] (12.62000,0.13000) -- (12.68600,0.13000) -- (12.68600,0.35000) -- (12.62000,0.35000) -- cycle;
\fill[Scol!64.35] (12.68500,0.13000) -- (12.75100,0.13000) -- (12.75100,0.35000) -- (12.68500,0.35000) -- cycle;
\node[font=\rmfamily\fontsize{9}{11}\selectfont] at (6.25000,-0.10000) {-2.5};
\node[font=\rmfamily\fontsize{9}{11}\selectfont] at (9.50000,-0.10000) {+0};
\node[font=\rmfamily\fontsize{9}{11}\selectfont] at (12.75000,-0.10000) {+2.5};
\end{tikzpicture}
}
\caption{Training composition changes: mean of updates 48--52 minus mean of updates 1--5, with each update weighted equally. A single symmetric color scale applies to every cell. Printed T values retain six decimals. Cells show early-to-late changes in exclusive token-bucket shares. M pools RE/RT/ET/RET; U \mbox{is unclassified or protected.}}
\label{fig:rolechanges}
\end{minipage}
\par\vspace{14pt}
}]
\begin{table*}[tp]\centering\small
\begin{tabular}{llrrrrr}\toprule
Case & Checkpoint & Index & Selected tokens & Selected answer & Correct / 16 & Mean tokens\\\midrule
TR005 & T-priority & 6 & 5,455 & 117 (correct) & 11/16 & 8,370.50\\
 & R-milder & 6 & 5,576 & 167 (incorrect) & 12/16 & 6,738.69\\
TR066 & T-priority & 10 & 10,322 & 8 (correct) & 16/16 & 8,016.81\\
 & R-milder & 10 & 7,225 & 2 (incorrect) & 14/16 & 8,307.38\\
TR071 & T-priority & 1 & 16,384 & Truncated & 2/16 & 11,259.56\\
 & R-milder & 1 & 4,902 & 237 (correct) & 2/16 & 11,777.94\\
TR366 & T-priority & 6 & 1,071 & 40 (correct) & 16/16 & 1,833.56\\
 & R-milder & 6 & 1,422 & 40 (correct) & 16/16 & 1,884.31\\\bottomrule
\end{tabular}
\caption{Selected-response outcomes and all-16 statistics for the supplementary T-priority and R-milder cases. TR005/TR071 are from AIME 2025, TR066 from AMC23, and TR366 from OlympiadBench. In table order, DECS has 13/16, 13/16, 0/16, and 16/16 correct samples. Means are rounded to two decimals; source means retain full precision. TR identifiers are archive sequence IDs, not benchmark question numbers.}
\label{tab:trcases}\end{table*}

\section{Case Studies and Trace Comparisons}
\label{app:caseinterpretation}
\label{app:cases}
Section~\ref{sec:cases} compares DECS and default PCRS-T on Q211. This appendix follows the four selected responses from their initial formulation through checking to the final answer, then compares T-priority and R-milder responses.

\subsection{Selection and source protocol}
\label{app:caseprotocol}
For each question, the displayed response is the first member of the largest original answer group, with first-occurrence tie breaking. Cases pair responses by question, reporting selected-response outcomes and all-16 statistics.

Case analysis covers 227 questions across three models (681 representative responses) and an expanded four-checkpoint set of 600 questions (2,400 representative responses). The displayed cases illustrate specific mathematical decisions; selection rules and all-16 response statistics accompany the excerpts.

\paragraph{Reading the gray panels.}
Panels present chronological excerpts with [...] for omissions. Mathematical Unicode is typeset without changing the quotations. Authors supply question condensations, headings, summaries, and annotations: blue marks valid steps or repairs, orange errors, and purple repeated checking. The prose identifies these distinctions for monochrome reading. Selected-response lengths and all-16 means are reported separately.

\subsection{Four-configuration comparisons}
\label{app:maincases}
Figures~\ref{fig:atlasQ211} and~\ref{fig:atlasQ057} retain the chronological excerpts and all-16 statistics for Q211 and the supplementary Q057 example.

\subsection{T-priority and R-milder Cases}
\label{app:trchecks}
The supplementary cases compare step-52 T-priority and R-milder checkpoints of PCRS-T.

\paragraph{Completed reading coverage.}
The archive covers 373 paired questions (746 responses): 105 each from MATH500, OlympiadBench, and GPQA-Diamond; 18 from AIME 2024, 17 from AIME 2025, and 23 from AMC23. It combines 295 outcome-independent selections with 78 diagnostic selections. Annotation/scorer discrepancies and unreconciled tags are excluded from quantitative mechanism conclusions.

\paragraph{Selection and evidence.}
Each pair uses the same sample index, chosen deterministically before reading and independently of correctness or length. These indices identify cache draws, not matched random trajectories. The four cases provide model-assisted qualitative analysis rather than blind human coding, mechanism-frequency estimates, or causal tests. Integer enumeration, polygon area, subset enumeration, and substitution independently confirm the four answers.

\paragraph{Constraint retention can favor either variant.}
TR005 (Figure~\ref{fig:tr005}) contrasts a check of both coordinate bounds with checks that leave one bound untested. TR066 (Figure~\ref{fig:tr066}) contrasts a correct final answer containing faulty intermediate checks with a correct local calculation applied to an incomplete region. The reverse selected-response outcome occurs in TR071, which asks for the probability that a uniformly chosen subset of the divisors of 2025 is nonempty and has least common multiple 2025, written as $m/n$, and requests $m+n$. The 15 divisors have form $3^a5^b$. Excluding all three divisors with $a=4$, all five with $b=2$, and accounting for their one-element overlap gives $2^{15}-2^{12}-2^{10}+2^8=27{,}904$ valid subsets. Hence the probability is $109/128$ and the answer is 237. R-milder retains this complement count (``32768 - 4096 -1024 +256=27904.'') and repairs a proposed independence error. T-priority incorrectly allows an excluded divisor to remain optional (``if it has a=4, it can be included or not, which is 2 choices.''). It repeatedly miscounts these groups and reaches the token cap without a final answer. Yet both checkpoints have only 2/16 correct samples; R-milder has a higher all-16 mean length and more capped responses (5 versus 4). These cases examine whether checks preserve \mbox{the original constraints.}

\paragraph{A stable control.}
TR366 asks how many gold marbles to add to 100 marbles to change the black:gold ratio from $1:4$ to $1:6$. Both responses retain 20 black and initially 80 gold marbles, solve $20/(80+x)=1/6$, and obtain $x=40$. R-milder additionally checks the total through ``20 = (1/7)(100 + x)'', also yielding 40. Its selected response is longer (1,422 versus 1,071 tokens), and both checkpoints achieve 16/16 correct samples. Thus additional checking can remain valid; its presence or length alone does \mbox{not identify a failure.}

\begin{figure*}[tp]\centering
\setlength{\smallskipamount}{2pt}\setlength{\medskipamount}{4pt}
\begin{lrbox}{\CaseLayoutBox}\begin{minipage}{\textwidth}\centering
\CaseQuestion{Q211 | Correct answers with different checking failures}{\textbf{Question (condensed).} Two-digit primes $p,q$ satisfy $p^2-q^2=2p+6q+8$. Maximize $p+q$. \textbf{Reference:} $162$}
\CaseCard{.485}{DECS}{0}{9,559}{correct}{
\textbf{Derive and enumerate.} [\ldots]\par ``So, only case 1 is possible, meaning p = q + 4.'' [\ldots] ``(q, p) = (13, 17), (19, 23), (37, 41), (43, 47), (67, 71), (79, 83).''\par\smallskip
\textbf{Swap the assignment.} [\ldots]\par ``Compute p\textsuperscript{2} - q\textsuperscript{2}: \textbf{\textcolor{orange!70!black}{79\textsuperscript{2} - 83\textsuperscript{2}}}. Let me compute 79\textsuperscript{2}: 6241, and 83\textsuperscript{2}: 6889. So, 6241 - 6889 = -648.''\par\smallskip
\textbf{False contradiction.} [\ldots]\par ``Wait, that's not equal to -648. There's a problem here. \textbf{\textcolor{orange!70!black}{Maybe I made a mistake in my initial assumption}}.''\par\smallskip
\textbf{Persistent confusion.} [\ldots]\par ``So, my algebra is correct, but when I plug in, \textbf{\textcolor{violet!80!black}{the negative comes in}}.''\par\smallskip
\textbf{Recover the assignment.} [\ldots]\par ``p = 83, q = 79.'' [\ldots] ``Wait, so 648 = 648. So, why did I get a negative before?''\par\smallskip
[\ldots]\ \textbf{Final answer:} $162$ (correct).\par\smallskip
}\hfill
\CaseCard{.485}{PCRS-H}{0}{3,393}{correct}{
\textbf{Derive the constraint.} [\ldots]\par ``Therefore, only Case 1 is possible: p = q + 4.'' [\ldots] ``So, the sums are 30, 42, 78, 90, 138, 162.''\par\smallskip
\textbf{Keep the assignment.} [\ldots]\par ``Let me compute p\textsuperscript{2} - q\textsuperscript{2} where \textbf{\textcolor{blue!65!black}{p=83, q=79}}.''\par\smallskip
\textbf{Verify.} [\ldots]\par ``Yes, that checks out. So, \textbf{\textcolor{blue!65!black}{648 = 648}}. Perfect.''\par\smallskip
\textbf{Overstate uniqueness.} [\ldots]\par ``Hence, (79,83) is \textbf{\textcolor{orange!70!black}{the only pair}} that satisfies the equation, and p + q is 162.''\par\smallskip
[\ldots]\ \textbf{Final answer:} $162$ (correct).\par\smallskip
}\par\medskip
\CaseCard{.485}{PCRS-T (Default)}{0}{2,754}{correct}{
\textbf{Derive and enumerate.} [\ldots]\par ``Simplifying, p = q + 4.'' [\ldots] ``(17,13), (23,19), (41,37), (47,43), (71,67), (83,79).''\par\smallskip
\textbf{Keep the assignment.} [\ldots]\par ``\textbf{\textcolor{blue!65!black}{83,79}}: 6889 - 6241=648\par 2p +6q +8=166 +474 +8=648. Correct.''\par\smallskip
\textbf{Retain other solutions.} [\ldots]\par ``So \textbf{\textcolor{blue!65!black}{all these pairs}} satisfy the equation.''\par\smallskip
\textbf{Check the maximum.} [\ldots]\par ``So, only those \textbf{\textcolor{blue!65!black}{six pairs}}. Therefore, 162 is indeed the largest.''\par\smallskip
[\ldots]\ \textbf{Final answer:} $162$ (correct).\par\smallskip
}\hfill
\CaseCard{.485}{PCRS-T (E-priority)}{0}{13,671}{correct}{
\textbf{Derive the constraint.} [\ldots]\par ``(p - q - 4)(p + q + 2) = 0.'' [\ldots] ``Therefore, only Case 1 is valid: p = q + 4.''\par\smallskip
\textbf{Swap the assignment.} [\ldots]\par ``\textbf{\textcolor{orange!70!black}{79\textsuperscript{2} - 83\textsuperscript{2}}}. Let's calculate 79\textsuperscript{2}: 6241. 83\textsuperscript{2}: 6889. So, 6241 - 6889 = -648.''\par\smallskip
\textbf{Repeat the contradiction.} [\ldots]\par ``But when I plug in \textbf{\textcolor{orange!70!black}{p=79 and q=83, p=q+4}}, but the equation doesn't hold.''\par\smallskip
\textbf{Late repair.} [\ldots]\par ``Therefore, in the case of p=79, q=83, \textbf{\textcolor{blue!65!black}{p \textless{} q, which is not allowed}} in this case.''\par\smallskip
[\ldots]\ \textbf{Final answer:} $162$ (correct).\par\smallskip
}\par\medskip
{\fontsize{9}{11}\selectfont\setlength{\tabcolsep}{8pt}
\begin{tabular}{lrrrr}\toprule
All 16 samples & DECS & PCRS-H & PCRS-T & E-priority\\\midrule
Correct / 16 & 13/16 & 14/16 & 16/16 & 13/16\\
Mean tokens & 7,685.25 & 6,543.19 & 6,691.50 & 5,805.94\\\bottomrule
\end{tabular}}\par\medskip
{\fontsize{9}{11}\selectfont\raggedright
\textbf{What the comparison shows.} Factoring gives $(p-q-4)(p+q+2)=0$, hence $p=q+4$. There are six valid pairs, with maximum at $(p,q)=(83,79)$. All four selected answers are correct; their intermediate reasoning is not equally reliable. PCRS-T keeps the assignment while still rechecking the enumeration; PCRS-H makes a uniqueness error despite listing other valid pairs in its final explanation; DECS/E-priority generate false contradictions, with E-priority repairing its swap late. E-priority's selected response uses 13,671 tokens; its all-16 mean is 5,805.94.\par}
\CaseLegend
\caption{Q211 (OlympiadBench): four-method chronological trace excerpts. Excerpts compare selected responses to the same question.}\label{fig:atlasQ211}\end{minipage}\end{lrbox}
\ifdim\dimexpr\ht\CaseLayoutBox+\dp\CaseLayoutBox\relax>\textheight
\PackageError{case-layout}{Case Q211 height \the\dimexpr\ht\CaseLayoutBox+\dp\CaseLayoutBox\relax\space exceeds \the\textheight}{Shorten the author commentary or adjust the layout.}
\fi
\usebox{\CaseLayoutBox}\end{figure*}

\begin{figure*}[tp]\centering
\setlength{\smallskipamount}{2pt}\setlength{\medskipamount}{4pt}
\begin{lrbox}{\CaseLayoutBox}\begin{minipage}{\textwidth}\centering
\CaseQuestion{Q057 | A repair found, retained, or discarded}{\textbf{Question (condensed).} A tube (radius 10\,mm, height 100\,mm) contains water of depth $h$. A rod (radius 2.5\,mm, height 150\,mm) rests on the bottom after insertion; the water depth becomes 64\,mm. Find $h$. \textbf{Reference:} $60$\,mm}
\CaseCard{.485}{DECS}{2}{1,772}{incorrect}{
\textbf{Initial water volume.} [\ldots]\par ``So, the initial volume of water is \ensuremath{\pi}*(10)\textasciicircum{}2*h = 100\ensuremath{\pi}h mm\textsuperscript{3}.''\par\smallskip
\textbf{Submerged height.} [\ldots]\par ``So, the rod is 150 mm tall, but \textbf{\textcolor{orange!70!black}{only 100 mm is submerged}}. So, the submerged height is 100 mm.''\par\smallskip
\textbf{Conservation equation.} [\ldots]\par ``\textbf{\textcolor{orange!70!black}{100\ensuremath{\pi}h - 625\ensuremath{\pi}}} = 100\ensuremath{\pi}*64.''\par\smallskip
\textbf{Solve and recheck.} [\ldots]\par ``100h = 6400 + 625 = 7025.'' [\ldots] ``So, the equation is correct.''\par\smallskip
[\ldots]\ \textbf{Final answer:} $70.25$ (incorrect).\par\smallskip
}\hfill
\CaseCard{.485}{PCRS-H}{1}{6,407}{incorrect}{
\textbf{Start with the full rod.} [\ldots]\par ``6400\ensuremath{\pi} = 100\ensuremath{\pi}h + 937.5\ensuremath{\pi}''\par\smallskip
\textbf{Correct intermediate step.} [\ldots]\par ``So, the submerged length of the rod is \textbf{\textcolor{blue!65!black}{64 mm}}.''\par\smallskip
\textbf{Discard the repair.} [\ldots]\par ``So, I think the \textbf{\textcolor{orange!70!black}{initial mistake was considering the submerged length as 64 mm}}, but actually, the submerged length is determined by the displaced volume, which is equal to the volume of the rod, so the increase in water level is (V\_rod / (\ensuremath{\pi}r\_tube\textsuperscript{2})), which is (937.5\ensuremath{\pi} / (\ensuremath{\pi}*10\textsuperscript{2})) = 937.5 / 100 = 9.375 mm.''\par\smallskip
[\ldots]\ \textbf{Final answer:} $54.625$ (incorrect).\par\smallskip
}\par\medskip
\CaseCard{.485}{PCRS-T (Default)}{5}{2,301}{correct}{
\textbf{Initial water volume.} [\ldots]\par ``So, initial volume of water is V1 = A * h = 100\ensuremath{\pi} h.''\par\smallskip
\textbf{Submerged volume.} [\ldots]\par ``V\_rod = \ensuremath{\pi} * (2.5)\textasciicircum{}2 \textbf{\textcolor{blue!65!black}{* 64}} = \ensuremath{\pi} * 6.25 * 64 = \ensuremath{\pi} * 400 = 400\ensuremath{\pi} mm\textsuperscript{3}.''\par\smallskip
\textbf{Conservation equation.} [\ldots]\par ``So, \textbf{\textcolor{blue!65!black}{100\ensuremath{\pi} h + 400\ensuremath{\pi} = 6400\ensuremath{\pi}}}''\par\smallskip
\textbf{Solve.} [\ldots]\par ``So, h = 6000 / 100 = \textbf{\textcolor{blue!65!black}{60 mm}}.''\par\smallskip
\textbf{Check the rise.} [\ldots]\par ``So, the increase in water height is (400\ensuremath{\pi}) / (100\ensuremath{\pi}) = 4 mm.'' [\ldots] ``Wait, so 64 - h = 4, so h = 60 mm. So, same result.''\par\smallskip
[\ldots]\ \textbf{Final answer:} $60$ (correct).\par\smallskip
}\hfill
\CaseCard{.485}{PCRS-T (E-priority)}{6}{4,597}{incorrect}{
\textbf{Correct intermediate step.} [\ldots]\par ``But how much is submerged? Since the water is at 64 mm, the \textbf{\textcolor{blue!65!black}{submerged portion of the rod is 64 mm}}.''\par\smallskip
\textbf{Wrong conservation sign.} [\ldots]\par ``So, \ensuremath{\pi} * R\textasciicircum{}2 * h = \ensuremath{\pi} * R\textasciicircum{}2 * 64 \textbf{\textcolor{orange!70!black}{+ \ensuremath{\pi} * r\_rod\textasciicircum{}2 * 64}}.''\par\smallskip
\textbf{Evaluate.} [\ldots]\par ``h = 6800 / 100 = 68 mm''\par\smallskip
\textbf{Repeated confirmation.} [\ldots]\par ``So, \textbf{\textcolor{violet!80!black}{both methods}} give h = 68 mm.''\par\smallskip
[\ldots]\ \textbf{Final answer:} $68$ (incorrect).\par\smallskip
}\par\medskip
{\fontsize{9}{11}\selectfont\setlength{\tabcolsep}{8pt}
\begin{tabular}{lrrrr}\toprule
All 16 samples & DECS & PCRS-H & PCRS-T & E-priority\\\midrule
Correct / 16 & 1/16 & 3/16 & 4/16 & 1/16\\
Mean tokens & 5,872.94 & 6,333.00 & 4,980.50 & 5,911.56\\\bottomrule
\end{tabular}}\par\medskip
{\fontsize{9}{11}\selectfont\raggedright
\textbf{What the comparison shows.} Conserved water volume is $(100-6.25)\pi\times64=6000\pi$, so $100\pi h=6000\pi$ and $h=60$. DECS uses 100\,mm and a wrong conservation sign; PCRS-H returns to the full 150\,mm rod after finding 64\,mm; E-priority uses 64\,mm but the wrong sign. E-priority's two checks reuse the same wrong relation. PCRS-T combines the 64\,mm submerged height with the correct conservation equation and obtains 60\,mm. Across all 16 samples, PCRS-T records 4 correct responses versus 1 for E-priority, with mean lengths of 4,980.50 and 5,911.56 tokens, respectively.\par}
\CaseLegend
\caption{Q057 (OlympiadBench): four-method chronological trace excerpts. Excerpts compare selected responses to the same question.}\label{fig:atlasQ057}\end{minipage}\end{lrbox}
\ifdim\dimexpr\ht\CaseLayoutBox+\dp\CaseLayoutBox\relax>\textheight
\PackageError{case-layout}{Case Q057 exceeds the page height}{Shorten the author commentary or adjust the layout.}
\fi
\usebox{\CaseLayoutBox}\end{figure*}

\begin{figure*}[tp]\centering
\setlength{\smallskipamount}{2pt}\setlength{\medskipamount}{3pt}
\begin{lrbox}{\CaseLayoutBox}\begin{minipage}{\textwidth}\centering
\CaseQuestion{TR005 | Joint coordinate bounds}{\textbf{Question (condensed).} Count integer solutions in $[-100,100]^2$ to $12x^2-xy-6y^2=0$. \textbf{Reference:} 117.}
\CaseCard{.485}{PCRS-T (T-priority)}{6}{5,455}{correct}{
\textbf{Apply both bounds.} [\ldots]\par
``\textcolor{blue!65!black}{But since k must satisfy both}''\par\smallskip
\textbf{Test an excluded point.} [\ldots]\par
``\textcolor{blue!65!black}{k=26: x=78, y=104. But y=104 is beyond 100}''\par\smallskip
\textbf{Combine the counts.} [\ldots]\par
``\textcolor{blue!65!black}{total number of solutions is 51 + 67 -1 =117.}''\par\smallskip
\textbf{Final answer:} $117$. [\ldots]
}\hfill
\CaseCard{.485}{PCRS-T (R-milder)}{6}{5,576}{incorrect}{
\textbf{Count from the x-bound.} [\ldots]\par
``\textcolor{orange!70!black}{So, 67 ordered pairs from this case.}''\par\smallskip
\textbf{Assert both bounds.} [\ldots]\par
``\textcolor{orange!70!black}{So, all these x and y are within the required range.}''\par\smallskip
\textbf{Combine the counts.} [\ldots]\par
``\textcolor{orange!70!black}{67 + 101 - 1 = 167}''\par\smallskip
\textbf{Final answer:} $167$. [\ldots]
}\par\medskip
{\fontsize{9}{11}\selectfont\raggedright\textbf{What the comparison shows.} Factoring gives $(4x-3y)(3x+2y)=0$: $(3k,4k)$ permits $-25\le k\le25$ and $(2k,-3k)$ permits $-33\le k\le33$. Removing the shared origin gives $51+67-1=117$. T-priority checks both bounds; R-milder admits $(99,132)$ and $(100,-150)$ outside the $y$-bound. T-priority still rechecks redundantly and repairs an algebraic error; its selected success does not establish better all-16 performance.\par}
\caption{TR005: paired chronological excerpts; statistics in Table~\ref{tab:trcases}.}\label{fig:tr005}
\end{minipage}\end{lrbox}
\ifdim\dimexpr\ht\CaseLayoutBox+\dp\CaseLayoutBox\relax>\textheight
\PackageError{case-layout}{Supplementary case exceeds page height}{Adjust the layout.}
\fi
\usebox{\CaseLayoutBox}\end{figure*}

\begin{figure*}[tp]\centering
\setlength{\smallskipamount}{2pt}\setlength{\medskipamount}{3pt}
\begin{lrbox}{\CaseLayoutBox}\begin{minipage}{\textwidth}\centering
\CaseQuestion{TR066 | A local check loses the global region}{\textbf{Question (condensed).} Find the area defined by $\bigl||x|-1\bigr|+\bigl||y|-1\bigr|\le1$. \textbf{Reference:} 8.}
\CaseCard{.485}{PCRS-T (T-priority)}{10}{10,322}{correct}{
\textbf{Faulty case split.} [\ldots]\par
``\textcolor{orange!70!black}{Case 2: |x| > 1 or |y| > 1.}''\par\smallskip
\textbf{Misinterpret the mapping.} [\ldots]\par
``\textcolor{orange!70!black}{the Jacobian determinant is 4}''\par\smallskip
\textbf{Return to the correct answer.} [\ldots]\par
``\textcolor{blue!65!black}{Therefore, the correct area is 8.}''\par\smallskip
\textbf{Final answer:} $8$. [\ldots]
}\hfill
\CaseCard{.485}{PCRS-T (R-milder)}{10}{7,225}{incorrect}{
\textbf{Check one diamond.} [\ldots]\par
``\textcolor{blue!65!black}{Area = 0.5 * |Sum1 - Sum2| = 0.5 * |6 - 2| = 0.5 * 4 = 2.}''\par\smallskip
\textbf{Restrict the whole region.} [\ldots]\par
``\textcolor{orange!70!black}{However, upon closer inspection, we realize that the region is entirely within the first quadrant and does not extend into other quadrants.}''\par\smallskip
\textbf{Final answer:} $2$. [\ldots]
}\par\medskip
{\fontsize{9}{11}\selectfont\raggedright\textbf{What the comparison shows.} The diamond with vertices $(0,1),(1,0),(2,1),(1,2)$ has area 2; four reflected copies give 8. R-milder computes one diamond correctly but excludes the other quadrants, including valid $(-1,1)$. T-priority returns 8 despite a faulty OR split and a false Jacobian claim: four counts reflected copies. Selected-response and all-16 length \mbox{rankings differ (Table~\ref{tab:trcases}).}\par}
\caption{TR066: paired chronological excerpts; statistics in Table~\ref{tab:trcases}.}\label{fig:tr066}
\end{minipage}\end{lrbox}
\ifdim\dimexpr\ht\CaseLayoutBox+\dp\CaseLayoutBox\relax>\textheight
\PackageError{case-layout}{Supplementary case exceeds page height}{Adjust the layout.}
\fi
\usebox{\CaseLayoutBox}\end{figure*}

\clearpage
\end{document}